\documentclass[twocolumn,runningheads]{svjour3}

\usepackage[latin1]{inputenc}
\usepackage{times}

\usepackage{psfrag,amsmath, amssymb, amscd, amsfonts, latexsym, epsf, graphicx,amscd}

\graphicspath{{../pdf/}{../jpeg/}}
\DeclareGraphicsExtensions{.pdf,.jpeg,.png}
\usepackage[lofdepth,lotdepth]{subfig}

\usepackage{textcomp}
\usepackage{enumerate}

\def\bbbr{{\mathbb R}} 
\def\bbbz{{\mathbb Z}}
\def\calD{\mathcal{D}}
\def\calT{\mathcal{T}}
\def\calS{\mathcal{S}}
\usepackage{amsmath}
\usepackage{amsfonts}
\usepackage{amssymb}

\newcommand{\local}{\operatorname{local}}

\newcommand{\avg}{\operatorname{avg}}
\newcommand{\conc}{\operatorname{conc}}

\newcommand{\diag}{\operatorname{diag}}

\journalname{Journal of Mathematical Imaging and Vision}
\journalname{arXiv preprint}

\begin{document}

\titlerunning{Scale-invariant scale-channel networks}
\title{\bf Scale-invariant scale-channel networks: Deep networks that
  generalise to previously unseen scales%
\thanks{The support from the Swedish Research Council 
              (contract 2018-03586) is gratefully acknowledged. }}
\author{Ylva Jansson and Tony Lindeberg }

\institute{Computational Brain Science Lab,
        Division of Computational Science and Technology,
        KTH Royal Institute of Technology,
        SE-100 44 Stockholm, Sweden.
       \email{yjansson@kth.se \and tony@kth.se}}

\date{}

\maketitle

\begin{abstract}
\noindent
The ability to handle large scale variations is crucial for many real world visual tasks. 
A straightforward approach for handling scale in a deep network is to
process an image at several scales simultaneously in a set of 
{\em scale channels}. 
Scale invariance can then, in principle, be achieved by using weight
sharing between the scale channels together with max or average
pooling over the outputs from the scale channels.
The ability of such {\em scale-channel networks\/} to generalise 
to scales not present in the training set over significant scale
ranges has, however, not previously been explored. 

In this paper, we present a systematic study of this methodology by
implementing different types of scale-channel networks and evaluating
their ability to generalise to previously unseen scales. 
We develop a formalism for analysing the covariance and invariance
properties of scale-channel networks, including exploring their relations to scale-space theory, and exploring how different design choices, 
unique to scaling transformations, affect the overall performance of
scale-channel networks.
We first show that two previously proposed scale-channel network designs, in one case, generalise \emph{no better than a standard CNN} to scales not present in the training set, and in the second case, have \emph{limited scale generalisation ability}.
We explain theoretically and demonstrate experimentally 
why generalisation fails or is limited in these cases. 
We then propose a new type of {\em foveated scale-channel architecture}, 
where the scale channels process increasingly larger parts of the image with decreasing resolution. 
This new type of scale-channel network is shown to generalise
extremely well, provided sufficient image resolution and the absence
of boundary effects. 
Our proposed FovMax and FovAvg networks perform almost 
identically over a scale range of 8, also when training on 
{\em single-scale training data}, and do also give improved
performance  when learning from datasets with large scale variations
in the small sample regime.


\end{abstract}

\section{Introduction}
\label{sec-intro}

Scaling transformations are as pervasive in natural image data as
translations. In any natural scene, the size of the projection of an object on the
retina or a digital sensor varies continuously with the distance
between the object and the observer. 
Compared to translations, scale variability is in some sense harder to
handle for a biological or artificial agent. It is possible to fixate
an object, thus centering it on the retina. The equivalence for
scaling, which would be to ensure a constant distance to objects
before further processing, is not a viable solution. A human observer
can nonetheless recognise an object at a range of scales, from a
single observation, and there is, indeed, experimental evidence
demonstrating scale-invariant processing in the primate visual cortex
\cite{BieCoo92-ExpPhys,LogPauPog95-CurrBiol,ItoTamFujTan95-JNeuroPhys,FurEng00-VisRes,HunKrePogDiC05-Science,IsiMeyLeiPog13-JNPhys}.
Convolutional neural networks (CNNs) already encode structural
assumptions about translation invariance and locality, which by the
successful application of CNNs for computer vision tasks has been
demonstrated to constitute useful priors for processing visual data.
We propose that structural assumptions about scale could, similarly 
to translation covariance, be a useful prior in convolutional neural
networks. 

Encoding structural priors about a larger group of visual
transformations, including scaling transformations and affine
transformations, is an integrated part of a range of successful 
classical computer vision approaches
\cite{Lin97-IJCV,Lin98-IJCV,LG96-IVC,BL97-CVIU,ChoVerHalCro00-ECCV,Bau00-CVPR,MikSch04-IJCV,Low04-IJCV,BayEssTuyGoo08-CVIU,TuyMik08-Book,MorYu09-SIAM,Lin15-JMIV}
and in a theory for explaining the computational function of early visual
receptive fields \cite{Lin13-BICY,Lin21-Heliyon}.
There is a growing body of work on invariant CNNs, especially
concerning invariance to 2D/3D rotations and flips 
\cite{BruMal13-PAMI,WuHuKon15-arXiv,MarVolTui16-ICPR,CohWel16-ICML,DieFauKav16-ICML,LapSavBuhPol16-CVPR,WorGarTurBro17-CVPR,ZhoYeQiuJia17-CVPR,MarVolKomTui17-ICCV,CohWel17-ICLR,WeiGeiWelBooCoh18-NIPS,WeiHamSto18-CVPR,WorBro18-ECCV,CheHanZhoXu18-IP,CohGeiKoeWel18-ICLR,ThoSmiKeaYanLiKohRil18-arXiv}. 
There has been some recent work on scale-covariant and scale-invariant recognition in CNNs,
where recent approaches
\cite{XuXiaZhaYanZha14-arXiv,KanShaJac14-arXiv,MarKelLobTui18-arXiv,GhoGup19-arXiv,WorWel19-NeuroIPS}
have shown improvements compared to standard CNNs for scale
variability present both in the training and the testing sets. 
These approaches have, however, either not been evaluated for the 
task of generalisation to scales {\em not present in the training set\/} 
\cite{KanShaJac14-arXiv,EstAllZhoDan18-ICLR,MarKelLobTui18-arXiv,WorWel19-NeuroIPS} 
or only across a very limited scale range \cite{XuXiaZhaYanZha14-arXiv,GhoGup19-arXiv}. 
Thus, the possibilities for CNNs to generalise to previously unseen
scales have so far not been well explored. 

The structure of a standard CNN implies a preferred scale as decided by
the fixed size of the filters (often $3 \times 3$ or $5 \times 5$
kernels) together with the depth and max pooling strategy
applied. This determines the resolution at which the image is
processed and the size of the receptive fields of individual units at
different depths.
A vanilla CNN is, therefore, not designed for multi-scale
processing. Because of this, state-of-the-art object detection
approaches that are exposed to larger scale variability employ different mechanisms, such as
branching off classifiers at different depths
\cite{SerLeC11-IJCNN,CaiFanFerVas16-ECCV}, learning to transform the
input or the filters \cite{JadSimZisKav15-NIPS,LinLuc17-CVPR,HenVed17-ICML}, or
by combining the deep network with different types of image pyramids
\cite{SerEigZhaMatFerLeC13-arXiv,Gir15-ICCV,LinDolGirHeHarBel17-CVPR,LinGoyGirHeDol17-ICCV,HeKiDolGir17-ICCV,HuRam17-CVPR}. 

The goal of these approaches has, however, not been to {\em generalise between scales\/} 
and even though they enable multi-scale processing, they lack the type
of structure necessary for true scale invariance.
Thus, it is not possible to predict how they will react to objects
appearing at new scales in the testing set or a to real world
scenario. This can lead to undesirable effects, as shown in the rich
literature on adversarial examples, where it has been demonstrated
that CNNs suffer from unintuitive failure modes when presented with
data outside the training distribution
\cite{SzeZarSutBruErhGooFer13-arXiv,NguYoClu15-CVPR,MooFawFro16-CVPR,TanGri16-arXiv,SuVarKou17-arXiv,MooFawFawFro17-CVPR,BakLuErlKel18-CompBiol}.
This includes adversarial examples constructed by means of small
translations, rotations and scalings
\cite{EngTraTsiSchMad17-arXiv,FawFro15-BMVC}, that is transformations
that are partially represented in a training set of natural
images. {\em Scale-invariant CNNs\/} could enable both multi-scale
processing and predictable behaviour when encountering objects at
novel scales, without the need to fully span all possible scales in
the training set. 

Most likely, a set of different strategies will be needed to handle
the full scale variability in the natural world. {\em Full
  invariance\/} over scale factors of 100 or more, as present in natural
images, might not be viable in a network with similar type of
processing at fine and coarse 
scales\footnote{When analysing image data with very large scale
  variations, the finite receptive field of any detector and the
  difference in image resolution between objects observed at different
  scales will imply a large difference in appearance between very
  small and very large objects. This implies that  fully invariant
  processing over such wide scale ranges might not be an applicable
  strategy. Instead different strategies will likely be needed to
  recognise objects at very low resolution from those needed to
  recognise objects at very high resolution.}. 
We argue, however, that a deep learning based approach that is
invariant over a significant scale range could be an important part 
of the solution to handling also such large scale variations. 
Note that the term {\em scale invariance\/} has sometimes, in the
computer vision literature, been used in a weaker sense of ``the
ability to process objects of varying sizes" or ``learn in the presence
of scale variability". We will here use the term in a stricter classical
sense of a classifier/feature extractor whose output does not change 
when the input is transformed.

One of the simplest CNN architectures used for covariant and invariant
image processing is a channel network (also referred to as siamese
network)
\cite{CirMeiSch12-CVPR,LapSavBuhPol16-CVPR,DieWilDam15-RoyAstro}. 
In such an architecture, transformed copies of the input image are
processed in parallel by different ``channels" (subnetworks)
corresponding to a set of image transformations. This approach can
together with weight sharing and max or average pooling over the
output from the channels enable invariant recognition for finite
transformation groups, such as  90 degree rotations and flips. An
{\em invariant scale-channel network\/} is a natural extension of
invariant channel networks as previously explored for rotations in
\cite{LapSavBuhPol16-CVPR}.  It can equivalently be seen as a way of
extending ideas underlying the classical scale-space methodology to
deep learning
\cite{Iij62,Wit83,Koe84,KoeDoo92-PAMI,Lin93-Dis,Lin94-SI,Flo97-book,WeiIshImi99-JMIV,Haa04-book,DuiFloGraRom04-JMIV,Lin10-JMIV},
in the sense that the in the absense of further information, the image
data is processed at all scales simultaneously, and that the outputs from the scale channels will constitute
a non-linear scale-covariant multi-scale representation of the input
image.

\subsection{Contribution and novelty}

The subject of this paper is to investigate the possibility to
construct a scale-invariant CNN based on a scale-channel
architecture.
The key contributions of our work are to implement
different possible types of scale-channel networks and to evaluate the
ability of these networks to generalise to previously unseen scales,
so that we can train a network at some scale(s) and
test it at other scales, without complementary use of data
augmentation. 
It should be noted that previous scale-channel networks exist, but
those are explicitly designed for multi-scale processing
\cite{FarCouNajLeC13-PAMI,NooPos16-PR} rather than scale invariance or
have not been evaluated with regard to their ability to generalise to
unseen scales over any significant scale range
\cite{XuXiaZhaYanZha14-arXiv}. We here implement and evaluate networks
based on principles similar to these previous approaches, but also a
new type of foveated scale-channel network, where the individual scale
channels process increasingly larger parts of the image with
decreasing resolution. 

To enable testing each approach over a large range of scales, we create 
a new variation of the MNIST dataset, referred to as the MNIST Large Scale dataset, 
with scale variations up to a factor of 8. This represents a dataset
with sufficient resolution and image size to enable invariant recognition 
over a wide range of scale factors. We also rescale the CIFAR-10
dataset over a scale factor of 4, which is a wider scale range than 
has previously been evaluated for this dataset. This rescaled CIFAR-10 
dataset is used to test if scale-invariant networks can still give 
significant improvements in generalisation to new scales, 
in the presence of limited image resolution and for small image sizes. 
We evaluate the ability to generalise to previously unseen scales for
the different types of channel networks, by first training on a single
scale or a limited range of scales and then testing recognition for
scales not present in the training set. The results are compared to a
vanilla CNN baseline.

Our experiments on the MNIST Large Scale dataset show that two
previously used scale-channel network designs or
methodologies, in one case, do not generalise any better than a standard CNN to scales not present in the training set or, in the second case, have limited generalisation ability.
The first type of method is based on
{\em concatenating the outputs from the scale channels\/} and using
this as input to a fully connected layer (as opposed to applying max
or average pooling over the scale-dimension). We show that such a
network does not learn to combine the output from the scale channels
in a correct way so as to enable generalisation to previously unseen
scales. The reason for this is the absence of a structure to enforce
scale invariance. The second type of method is to handle the difference in
image size between the rescaled images in the scale channels, by applying the subnetwork
corresponding to each channel in {\em a sliding window manner}. This
methodology, however, implies that the rescaled copies of an image are
not processed {\em in the same way\/}, since for an object processed in
scale channels corresponding to an upscaled image, a wide range of
different, ({\em e.g.\/}, non-centered) object views, will be processed, compared
to only processing the central view for an object in a downscaled
image. This implies that full invariance cannot be achieved, since max (or
average) pooling will be performed over {\em different views of the objects
  for different scales\/}, which implies that the max (or average) over
the scale dimension is not guaranteed to be stable when the input
is transformed.

We do, instead, propose a new type of foveated scale-channel architecture, 
where the scale channels process increasingly larger parts of the image 
with decreasing resolution. Together with max or average pooling,
this leads to our FovMax and FovAvg networks. 
We show that this approach enables extremely good generalisation,
when the image resolution is sufficient and there is an absence 
of boundary effects. Notably, for rescalings of MNIST, 
almost identical performance over a scale range of 8 is achieved, 
when training on {\em single size\/} training data. 
We further show that, also on the CIFAR-10 dataset, in the presence 
of severe limitations regarding image resolution and image size, 
the foveated scale-channel networks still provide considerably 
better generalisation ability compared to both a standard CNN 
and an alternative scale-channel approach.
We also demonstrate that the FovMax and FovAvg networks give improved
performance for datasets with large scale variations in both the
training and testing data, in the small sample regime. 

We propose that the presented foveated scale-channel networks
will prove useful in situations where a simple approach that can 
generalise to unseen scales or learning from small datasets with 
large scale variations is needed. Our study also highlights 
possibilities and limitations for scale-invariant CNNs and 
provides a simple baseline to evaluate other approaches against. 
Finally, we see possibilities to integrate the foveated scale-channel 
network, or similar types of foveated scale-invariant processing, 
as subparts in more complex frameworks dealing with large scale variations.
 
\subsection{Relations to previous contribution}

This paper constitutes a substantially extended version of a
conference paper presented at the ICPR 2020 conference \cite{JanLin21-ICPR} 
and with substantial additions concerning:
\begin{itemize}
\item
  the motivations underlying this work and the importance of a scale
  generalisation ability for deep networks (Section~\ref{sec-intro}),
\item 
  a wider overview of related work (Section~\ref{sec-intro} and Section~\ref{sec-related-work}),
\item 
  theoretical relationships between the presented scale-channel
  networks and the notion of scale-space representation, including theoretical relationships between the presented scale-channel
  networks and scale-normalised derivatives with associated methods
  for scale selection
  (Section~\ref{sec-relations-scsp-theory}),
\item
  more extensive experimental results on the MNIST Large Scale dataset,
  specifically new experiments that investigate (i)~the dependency on the
  scale range spanned by the scale channels, (ii)~the dependency on the sampling density of
  the scale levels in the scale channels, (iii)~the influence of multi-scale
  learning over different scale intervals, and (iv)~an analysis of the scale selection
  properties over the multiple scale channels for the different types
  of scale-channel networks (Section~\ref{sec-exp-MNISTLargeScale}),
\item
  experimental results for the CIFAR-10 dataset subject to scaling
  transformations of the testing data
  (Section~\ref{sec-exp-rescaledCIFAR-10}),
\item
  details about the dataset creation for the MNIST Large Scale dataset
  (Appendix~\ref{app-mnist-large-scale}).
\end{itemize}
In relation to the ICPR 2020 paper, this paper therefore (i)~gives a more
general motivation for scale-channel networks in relation to the topic 
of scale generalisation, (ii)~presents more experimental results for further use
cases and an additional dataset, (iii)~gives deeper theoretical relationships
between scale-channel networks and scale-space theory and (iv)~gives 
overall better descriptions of several of the subjects 
treated in the paper, including (v)~more extensive references to related
literature.

\section{Relations to previous work}
\label{sec-related-work}

In the area of scale-space theory 
\cite{Iij62,Wit83,Koe84,KoeDoo92-PAMI,Lin93-Dis,Lin94-SI,Flo97-book,WeiIshImi99-JMIV,Haa04-book,DuiFloGraRom04-JMIV,Lin10-JMIV},
a multi-scale representation of an input image is created by
convolving the image with a set of rescaled Gaussian kernels and
Gaussian derivative filters, which are then often combined in
non-linear ways. 
In this way, a powerful methodology has been developed to handle
scaling transformations in classical computer vision
\cite{Lin97-IJCV,Lin98-IJCV,LG96-IVC,BL97-CVIU,ChoVerHalCro00-ECCV,MikSch04-IJCV,Low04-IJCV,BayEssTuyGoo08-CVIU,TuyMik08-Book,Lin15-JMIV}.
The scale-channel networks described in this paper can be seen as an extension of this philosophy of 
processing an image {\em at all scales simultaneously\/}, as a means of
achieving scale invariance, but instead using deep non-linear feature
extractors learned from data, as opposed to hand-crafted image features
or image descriptors.

CNNs can give impressive performance, but they are sensitive to scale
variations. Provided that the architecture of the deep network is
sufficiently flexible, moderate increase in the robustness to scaling
transformations can be obtained by augmenting the training images with
multiple rescaled copies of each training image (scale jittering)
\cite{BarCas91-TNN,SimZis15-ICLR}.
The performance does, however, degrade for scales not present in the training set
\cite{EngTraTsiSchMad19-ICML,FawFro15-BMVC,SinDav18-CVPR}, and
different network structure may be optimal for small {\em vs.\/}\ large images
\cite{SinDav18-CVPR}. It is furthermore possible to construct adversarial
examples by means of small translations, rotations and scalings
\cite{EngTraTsiSchMad17-arXiv,FawFro15-BMVC}.

State-of-the-art CNN based object detection approaches 
all employ different mechanisms to deal with scale variability, 
{\em e.g.\/}, branching off classifiers at different depths
\cite{CaiFanFerVas16-ECCV}, 
learning to transform the input or the filters 
\cite{JadSimZisKav15-NIPS,LinLuc17-CVPR,HenVed17-ICML}, 
using different types of image pyramids 
\cite{SerEigZhaMatFerLeC13-arXiv,Gir15-ICCV,LinDolGirHeHarBel17-CVPR,LinGoyGirHeDol17-ICCV,HeKiDolGir17-ICCV,HuRam17-CVPR},
or other approaches, where the image is rescaled to different
resolutions, possibly combined with
interactions or pooling between the layers 
\cite{RenHeGirZhaSun16-PAMI,NahKimLee17-CVPR,ChePapKokMurYui17-PAMI,SinDav18-CVPR}.
There are also deep networks that somehow handle the notion of scale
by approaches such as dilated convolutions
\cite{YuKol16-ICLR,YuKolFun17-CVPR,MehRasCasShaHaj18-ECCV},
scale-dependent pooling \cite{YanChoLin16-CVPR},
scale-adaptive convolutions \cite{ZhaTanZhaLiYan17-ICCV},
by spatially warping the image data by a log-polar transformation
prior to image filtering \cite{HenVed17-ICML,EstAllZhoDan18-ICLR},
 or
adding additional branches of down-samp\-lings and/or up-samplings in each
layer of the network \cite{WanKemFarYuiRas19-CVPR,CheFanXuYanKalRohYanFen19-ICCV}.
The goal of these approaches has, however, not been to generalise to
{\em previously unseen scales\/} and they lack the structure necessary
for true scale invariance.
 
Examples of handcrafted scale-invariant hierarchical descriptors are
\cite{SifMal13-CVPR,Lin20-JMIV}. We are, here, interested in
combining scale invariance with learning. There exist some previous
work aimed explicitly at scale-invariant recognition in CNNs
\cite{XuXiaZhaYanZha14-arXiv,KanShaJac14-arXiv,MarKelLobTui18-arXiv,GhoGup19-arXiv,WorWel19-NeuroIPS}
These
approaches have, however, either not been evaluated for the task of
generalisation to scales {\em not present in the training set\/}
\cite{KanShaJac14-arXiv,MarKelLobTui18-arXiv,WorWel19-NeuroIPS} or only
across a very limited scale range
\cite{XuXiaZhaYanZha14-arXiv,GhoGup19-arXiv}. Previous scale-channel networks exist,
but are explicitly designed for multi-scale processing
\cite{FarCouNajLeC13-PAMI,NooPos16-PR} rather than scale invariance,
or have not been evaluated with regard to their ability to generalise to
unseen scales over any significant scale range
\cite{SerEigZhaMatFerLeC13-arXiv,XuXiaZhaYanZha14-arXiv}. 
A dual approach to scale-covariant scale-channel networks that, however, allows for
scale invariance and scale generalisation, is presented in
\cite{Lin21-SSVM,Lin22-JMIV}, based on transforming continuous CNNs expressed in
terms of continuous functions for the filter weights with respect to
scaling transformations.
Other scale-covariant or scale-equivariant approaches to deep networks have
also been recently proposed in \cite{Bek20-ICLR,SosSzmSme20-ICLR,ZhuQiuCalSapChe19-arXiv,SosMosSme21-BMVC}.

There is a large literature on approaches to achieve
rotation-covariant and rotation-invariant networks
\cite{DieFauKav16-ICML,LapSavBuhPol16-CVPR,WorGarTurBro17-CVPR,ZhoYeQiuJia17-CVPR,MarVolKomTui17-ICCV,CohWel17-ICLR,WeiGeiWelBooCoh18-NIPS,WeiHamSto18-CVPR,WorBro18-ECCV,CheHanZhoXu18-IP}
with applications to different domains, including astronomy
\cite{DieWilDam15-RoyAstro},
remote sensing \cite{CheZhoHan17-GeoRemoteSens},
medical image analysis
\cite{WanZheYanJinCheYin17-TBiomedHealth,BekLafVetEppPluDui18-MICCAI,LafBekPluDuiVet20-MedImAnal} 
and texture classification \cite{AndDep18-arXiv}.
There are also approaches to invariant networks based on formalism from
group theory \cite{CohWel16-ICML,PogAns16-book,KonTri18-ICML}.

\section{Theory of continuous scale-channel networks}
\label{sec-theory}

In this section, we will introduce a mathematical framework for
modelling and analysing scale-channel networks based on a continuous
model of the image space. This model enables straightforward analysis
of the covariance and invariance properties of the channel networks,
that are later approximated in a discrete implementation. We, here,
generalise previous analysis of invariance properties of channel
networks \cite{LapSavBuhPol16-CVPR} to scale-channel networks. We
further analyse covariance properties and additional options for
aggregating information across transformation channels. 
 
\subsection{Images and image transformations}

We consider images $f: \mathbb{R}^N \to \mathbb{R}$ that are measurable functions in $L_\infty(\bbbr^N)$ and denote this space of images as $V$. 
A \emph{group of image transformations} corresponding to a group $G$
is a family of image transformations $\calT_g$ ($g \in G$) with a
group structure, {\em i.e.\/}, fulfilling the group axioms of closure, identity, associativity and inverse.
We denote the combination of two group elements $g,h \in G$ by $gh$ and the cardinality of $G$ as $|G|$. 
Formally, a group $G$ induces an \emph{action on functions} by acting on the underlying space on which the function is defined (here the image domain). We are here  interested in the group of \emph{uniform scalings} around $x_0$ with the group action
\begin{align}
(\calS_{s, x_0} f ) (x') &= f(x), \quad\quad x' = S_s(x - x_0) + x_0 
\label{eq-scale-def},
\end{align}
where $S_s = \diag(s)$. For simplicity, we often assume $x_0=0$  and denote $\calS_{s,0}$ as $\calS_s$ corresponding to 
\begin{equation}
(\calS_s f)(x) = f(S_s^{-1} x) = f_s(x).
\label{eq-scale-def2}
\end{equation}
We will also consider the translation group with the action (where $\delta \in \bbbr^N$)
\begin{align}
(\calD_\delta f) (x') &= f(x), \quad\quad x' = x + \delta. 
\label{eq-trans-def}
\end{align}

\subsection{Invariance and covariance}

Consider a general feature extractor $\Lambda: V \to \mathbb{K}$ 
that maps an image $f \in V$ to a feature representation $y \in \mathbb{K}$. 
In our continuous model, $\mathbb{K}$ will typically correspond to a 
set of $M$  feature maps (functions) so that $\Lambda f \in V^M$.
This is a continuous analogue of a discrete convolutional feature map with $M$ features.

A feature extractor%
\footnote{With regard to the scale-channel networks
  that we develop later in this paper, note that $\Lambda$ should be seen as representing the entire family
of scale channels, not a
single-scale channel in isolation. An {\em invariant\/} feature
extractor $\Lambda$ will then correspond to the result of
max pooling or average pooling over all the scale channels.}
$\Lambda$ is {\em covariant\/}%
\footnote{In the deep learning literature, the notion of
  ``equivariance'' is also often used for this relationship, which is
  referred to as ``covariance'' in scale-space theory. In this paper,
  we use the terminology ``covariance'' to maintain consistency with
  the earlier scale-space literature \cite{Lin13-ImPhys}.}
to a transformation
group $G$ (formally to the group action) if there exists an
{\em input independent\/} transformation $\tilde{\calT}_g$ that can
align the feature maps of a transformed image with those of the
original image
\begin{equation}
\Lambda(\calT_g f) = \tilde{\calT}_g (\Lambda f) \quad\quad \forall g \in G, f \in V.
\label{eq-covariance}
\end{equation}
Thus, for a covariant feature extractor it is possible to predict the
feature maps of a transformed image from the feature maps of the
original image or, in other words, the order between feature extraction and transformation does not matter, as illustrated in the commutative diagram in 
Figure~\ref{fig-comm-diag}.

\begin{figure}[hbt]
  \[
    \begin{CD}
       {\Lambda} \, f @>{\tilde{\calT_g}}>> \Lambda(\calT_g f) = \tilde{\calT}_g (\Lambda f) \\
       \Big\uparrow\vcenter{\rlap{$\scriptstyle{{\Lambda}}$}} & & \Big\uparrow\vcenter{\rlap{$\scriptstyle{{\Lambda}}$}} \\
       f @>{\calT_g}>> {\calT_g f} 
    \end{CD}
 \]
 \caption{Commutative diagram for a covariant feature extractor
  $\Lambda$, showing how the feature map of the transformed image can be matched to the feature map of the original image by a transformation of the feature space. Note that $\tilde{\calT}_g$ will \emph{correspond to the same transformation} as $\calT_g$, but might take a different form in the feature space.
}
\label{fig-comm-diag}
\end{figure}

A feature extractor $\Lambda$ is {\em invariant\/} to a transformation group $G$ if the feature representation of a transformed image is \emph{equal to} the feature representation of the original image 
\begin{equation}
\Lambda (\calT_g f) = \Lambda (f)  \quad\quad \forall g \in G, f \in V.
\label{eq-invariance}
\end{equation} 
Invariance is thus a special case of covariance, where $\tilde{\calT_g}$ is the identity transformation. 

\subsection{Continuous model of a CNN}

Let $\phi:  V \to V^{M_k}$ denote a continuous CNN with $k$ layers and $M_i$ feature channels in layer $i$.
Let $\theta^{(i)}$ represent the transformation between layers $i-1$ and $i$ such that
\begin{align}
(\phi^{(i)} f)(x,c) &= (\theta^{(i)} \theta^{(i-1)} \cdots \theta^{(2)} \theta^{(1)} f)(x,c), 
\label{eq-phi_i-def}
\end{align} 
where $c \in \{1,2, \dots M_k\}$ denotes the feature channel and $\phi = \phi^{(k)}$.
We model the transformation $\theta^{(i)}$ between two adjacent layers
$\phi^{(i-1)}f$ and $\phi^{(i)}f$ as a convolution followed by the
addition of a bias term $b_{i,c} \in \bbbr$ and the application of a
pointwise non-linearity $\sigma_i:\bbbr \to \bbbr$:
\begin{multline}
(\phi^{(i)} f)(x, c)\\ =  \sigma_i \left( \sum_{m=1}^{M_{i-1}} \int_{\xi \in \bbbr^N } (\phi^{(i-1)}f)(x-\xi, m)\, g^{(i)}_{m,c}(\xi) \, d\xi + b_{i,c}
\right)
\label{eq-phi_integral}
\end{multline}
where $g^{(i)}_{m,c} \in L_1(\bbbr^N)$
denotes the convolution kernel that propagates information from
feature 
channel $m$ in layer $i-1$ to output feature channel $c$ in layer $i$.
A final fully connected classification layer with compact support can
also be modelled as a convolution combined with a non-linearity
$\sigma_k$ that represents {\em a softmax operation\/} over the
feature channels.

\begin{figure*}[hbpt]
	\begin{center}
		\includegraphics[width=0.98\textwidth]{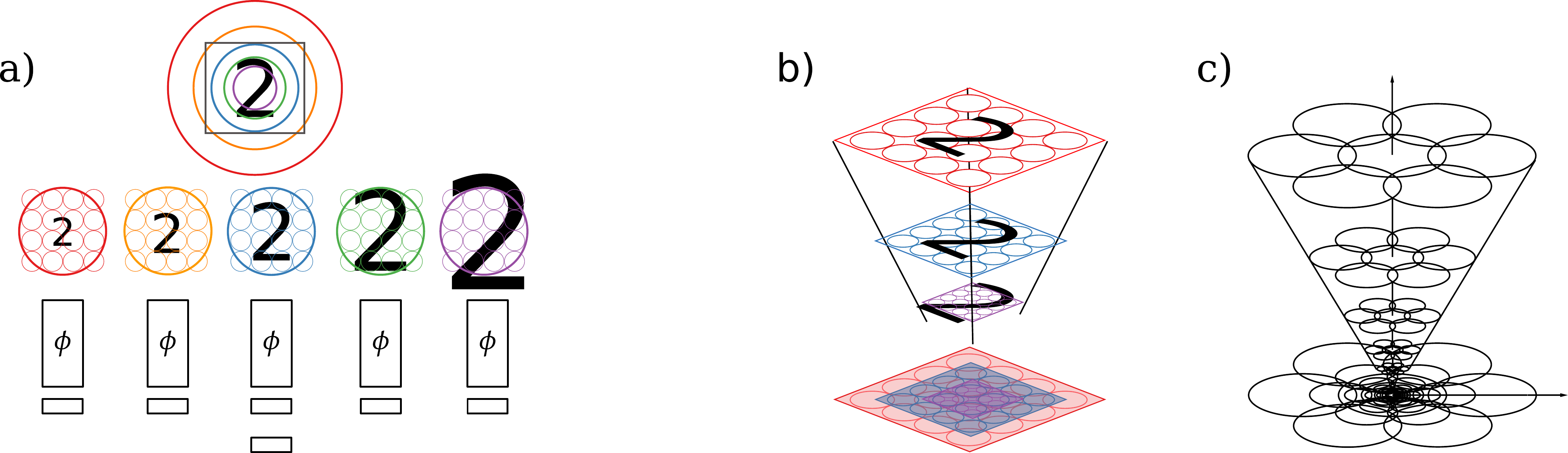} 
	\end{center}
	\caption{\emph{Foveated scale-channel networks.} (a) 
		Foveated scale-channel network that processes an image
                of the digit 2. Each scale channel has a fixed
                size receptive field/support region in relation to its rescaled image copy, but they will together process input regions
                corresponding to varying sizes in the original image
                (circles of corresponding colors).
		(b) This corresponds to a type of foveated processing,
                where the center of the image is processed with high
                resolution, which works well to detect small objects,
                while larger regions are processed using gradually
                reduced resolution, which enables detection of larger
                objects. (c) There is a close similarity between this
                model and the foveal scale space model \cite{CVAP166},
                which was motivated by a combination of regular scale
                space axioms with a complementary assumption of a
                uniform limited processing capacity at all scales.}
	\label{fig-foveated-processing}
\end{figure*}

\subsection{Scale-channel networks}
\label{sec-scale-channel-networks}

The key idea underlying \emph{channel networks} is to process
transformed copies of an input image in parallel, in a set of network
``channels" (subnetworks) with shared weights. For finite
transformation groups, such as discrete rotations, using one channel
corresponding to each group element and applying max pooling over the
channel dimension can give an invariant output. For continuous
but compact groups, invariance can instead be achieved for a discrete
subgroup. 

The scaling group is, however, neither finite nor compact. The key question that we address here
is whether a scale-channel network can still support invariant
recognition. 

We define a multi-column {\em scale-channel network\/} 
$\Lambda: V \to V^{M_k}$
for the group of scaling transformations $S$ by using a single base 
network $\phi: V \to V^{M_k}$ to define a set of {\em scale channels\/} $\{\phi_s \}_{s \in S}$
\begin{equation}
(\phi_s f)(x,c) = (\phi\, \calS_{s} f)(x,c) = (\phi f_s)(x,c),
\label{eq-phi_s-def}
\end{equation}
where each channel thus applies exactly the same operation to 
a scaled copy of the input image (see Figure~\ref{fig-foveated-processing}(a)). 
We denote the mapping from the input image to the scale-channel feature maps at depth $i$ as $\Gamma^{(i)}: V \to V^{M_i |S|}$
\begin{equation}
(\Gamma^{(i)} f)(x,c,s) =  (\phi^{(i)}_s f)(x, c)  = (\phi^{(i)} \calS_s f)(x,c).
\label{eq-gamma_s-def}
\end{equation}
A scale-channel network that is invariant to the continuous group of uniform scaling transformations $S = \{s \in \bbbr_+ \}$ 
can be constructed using an {\em infinite\/} set of scale channels $\{
\phi_s \}_{s \in S}$. The following analysis also holds for a set of
scale channels corresponding to a discrete subgroup of the group of
uniform scaling transformations, such that 
$S = \{\gamma^i |  i \in \bbbz\}$ for some $\gamma > 1$.

The final output $\Lambda f$ from the scale-channel network is an
aggregation across the scale dimension of the last layer scale-channel
feature maps. In our theoretical treatment, we combine the output of
the scale channels by the supremum
\begin{equation}
(\Lambda_{\sup} f)(x,c) = \sup_{s \in S} \left[  ( \phi_s f)(x,c) \right]
\label{eq-lambda_s-def}.
\end{equation}
Other permutation invariant operators, such as averaging operations,
could also be used. For this construction, the network output will be
invariant to {\em rescalings around $x_0=0$\/}
(global scale invariance). This architecture is appropriate  when
characterising a single centered object that might vary in scale and
it is the main architecture that we explore in this paper. 
Alternatively, one may instead pool over 
{\em corresponding image points\/} in the original image 
by operations of the form
\begin{equation}
\label{eq-lambda_s-skewed-def}
(\Lambda_{\sup}^{\local} f)(x, c) = \sup_{s \in S} \{ (\phi_s f)(S_s x, c) \}.
\end{equation}
This descriptor instead has the invariance property
\begin{equation}
  (\Lambda_{\sup}^{\local} f)(x_0, c) = (\Lambda_{\sup}^{\local} S_{s,
    x_0}f)(x_0, c) \quad \mbox{for all $x_0$},
\end{equation}
{\em i.e.\/}, when scaling around an
arbitrary image point, the output {\em at that specific point\/} does
not change (local scale invariance). This property makes it more
suitable to describe scenes with multiple objects. 

\subsubsection{Scale covariance}

Consider a scale-channel network $\Lambda$ (\ref{eq-lambda_s-def})
that expands the input over the group of uniform scaling transformations $S$. 
We can relate the feature map representation $\Gamma^{(i)}$ 
for a scaled image copy $\calS_t f$ for $t \in S$ and its original $f$
in terms of operator notation as
\begin{align}
&(\Gamma^{(i)} \calS_t f)(x,c,s) = (\phi_s^{(i)} \, \calS_t f)(x,c) \nonumber\\ 
&=(\phi^{(i)} \, \calS_s \, \calS_t f)(x,c) = (\phi^{(i)} \, \calS_{s t} f)(x,c) \nonumber\\ 
&= (\phi_{st}^{(i)} f)(x,c) = (\Gamma^{(i)} f)(x,c, s t),
\label{eq-scale-covariance}
\end{align}
where we have used the definitions (\ref{eq-phi_s-def}) and
(\ref{eq-gamma_s-def}) together with the fact that $S$ is a group. A
scaling of an image thus only results in a multiplicative shift in the
scale dimension of the feature maps. A more general and more rigorous
proof using an integral representation of the scale-channel network is
given in Section~\ref{sec-invariance-covariance-proof}. 

\subsubsection{Scale invariance}
\label{sec-scale-invariance}

Consider a scale-channel network $\Lambda_{\sup}$
(\ref{eq-lambda_s-def}) that selects the supremum over scales. We will
show that $\Lambda_{\sup}$ is scale invariant, {\em i.e.\/}, that
\begin{equation}
(\Lambda_{\sup}\, \calS_t f)(x,c) = (\Lambda_{\sup} f)(x,c).
\label{eq-scale-invariance}
\end{equation}
First, (\ref{eq-scale-covariance}) gives $\{\phi^{(i)}_s (\calS_t
f)\}_{s \in S} = \{ \phi_{st}^{(i)} (f) \}_{s \in S}$. 
Then, we note that $\{st\}_{s \in S} = St = S$. 
This holds both in the case when $S = \bbbr_+$ and in the case 
when $S = \{\gamma^i |  i \in \bbbz\}$.
Thus, we have
\begin{multline} 
\{(\phi_s^{(i)} \calS_t f)(x,c)\}_{s \in S}
= \{(\phi^{(i)}_{s t} f)(x,c)\}_{s \in S}\\ =  \{(\phi^{(i)}_{s} f)(x,c)\}_{s \in S}
\label{eq-phi_maps_scale},
\end{multline}
{\em i.e.\/}, \emph{the set} of outputs from the scale channels for a
transformed image is equal to the set of outputs from the 
scale channels for its original image.  
For any permutation invariant aggregation operator, 
such as the supremum, we have that
\begin{multline}
(\Lambda_{\sup}\, \calS_t f)(x,c) = \sup_{s \in S} \{(\phi^{(k)}_{s t} f)(x, c)\} \\
=   \sup_{s \in S} \{(\phi^{(k)}_{s} f)(x,c)\}  = (\Lambda_{\sup} f)(x, c),
\label{eq-lambda-scale-invariant}
\end{multline} 
and, thus, $\Lambda$ is invariant to uniform rescalings. 

\subsection{Proof of scale and translation covariance using an integral representation of a scale-channel network}
\label{sec-invariance-covariance-proof}

We, here, prove the transformation property 
\begin{equation}
(\Gamma^{(i)} h)(x, s, c)  = (\Gamma^{(i)} f)(x + S_s S_t x_1 - S_t x_2, st, c)
\end{equation}
of the scale-channel feature maps
under a more general combined scaling transformation and translation of the form
\begin{equation}
\label{eq-sc-transf-app-sc-cov-proof-noncent-offset}
h(x') = f(x) \quad \mbox{for} \quad x' = S_t (x - x_1) + x_2
\end{equation}
corresponding to
\begin{equation}
h(x) = f(S_t^{-1}(x - x_2) + x_1)
\label{eq-general_scaling_transformation}
\end{equation}
using an integral representation of the deep network.
In the special case when $x_1 = x_2 = x_0$, this corresponds to a
uniform scaling transformation around $x_0$ ({\em i.e.\/}, $S_ {x_0,s}$). 
With $x_1 = x_0$ and $x_2 = x_0 + \delta$, this
corresponds to a scaling transformation around $x_0$ followed by a
translation $\calD_\delta$. 

Consider a deep network $\phi^{(i)}$ (\ref{eq-phi_i-def}) 
and assume the integral representation (\ref{eq-phi_integral}), 
where we for simplicity of notation incorporate the offsets $b_{i,c}$ into the 
non-linearities $\sigma_{i,c}$. 
By expanding the integral representation 
of the rescaled image $h$ (\ref{eq-general_scaling_transformation}),
we have that that the feature representation in the scale-channel network is given by
 (with $M_0 = 1$ for a scalar input image):
 
 \begin{align}
 \begin{split}
 &  (\Gamma^{(i)} h)(x, s, c)  =  \{ \mbox{definition (\ref{eq-gamma_s-def})}  \}  
 = (\phi_s^{(i)} h)(x, c) 
 \end{split}\nonumber\\
 \begin{split}
 & = \{ \mbox{definition (\ref{eq-phi_s-def})} \} 
  = (\phi^{(i)} \, h_s)(x, c) 
 = \{ \mbox{equation~(\ref{eq-phi_i-def})} \}
 \end{split}\nonumber\\
 \begin{split}
 & = (\theta^{(i)} \theta^{(i-1)} \dots \theta^{(2)} \theta^{(1)}  h_s)(x, c) 
 = \{ \mbox{equation~(\ref{eq-phi_integral})} \}
 \end{split}\nonumber\\
 \begin{split}
 & = 
 \sigma_{i,c}
 \left(
 \sum_{m_i=1}^{M_{i-1}}
 \int_{\xi_i \in \bbbr^N} 
 \sigma_{i-1,m_i}
 \left(
 \sum_{m_{i-1}=1}^{M_{i-2}}
 \int_{\xi_{i-1} \in \bbbr^N} 
 \dots
 \right.
 \right.
 \end{split}\nonumber\\
 \begin{split}
 & \phantom{\sigma_i \vphantom{\left( \sum_{m_i=1}^{M_{i-1}} \right.)}} \quad
 \left.
 \left.
 \sigma_{1,m_2}
 \left(
 \sum_{m_1=1}^{M_0}
 \int_{\xi_1 \in \bbbr^N} 
 h_s(x - \xi_i - \xi_{i-1} -
 \dots - \xi_1) \, \times
 \right.
 \right.
 \right.
 \end{split}\nonumber\\
 \begin{split}
 & \phantom{\sigma_i \vphantom{\left( \sum_{m_i=1}^{M_{i-1}} \right.)}} \quad
 \left.
 \left.
 \left.
 \phantom{\left(\sum_{m_i=1}^{M_{i-1}} \right.)}
 g_{m_1,m_2}^{(1)}(\xi_1)  \, d\xi_1
 \right) \dots
 g_{m_{i-1},m_i}^{(i-1)}(\xi_{i-1})  \, d\xi_{i-1}
 \vphantom{\left( \sum_{m_i=1}^{M_{i-1}} \right.)} 
 \right)
 \right.
 \end{split}\nonumber\\
 \begin{split}
 \label{eq-exp-int-repr-sc-cov-proof-noncent-offset}
 \left.
 \vphantom{\left(\sum_{m_i=1}^{M_{i-1}} \right.)}\quad\quad
 g_{m_i,c}^{(i)}(\xi_i)  \, d\xi_i
 \vphantom{\left( \sum_{m_i=1}^{M_{i-1}} \right.)} \right).
 \end{split} 
 \end{align}
Under the scaling transformation
(\ref{eq-sc-transf-app-sc-cov-proof-noncent-offset}), 
the part of the integrand 
$h_s(x - \xi_i - \xi_{i-1} - \dots -\xi_1)$ transforms as follows:
\begin{align}
\begin{split}
& h_s(x - \xi_i - \xi_{i-1} - \dots - \xi_1)
\end{split}\nonumber\\
\begin{split}
& = \{ \mbox{$h_s(x) = h(S_s^{-1}x)$ according to definition (\ref{eq-scale-def2})}  \}
\end{split}\nonumber\\
\begin{split}
& = h(S_s^{-1} (x - \xi_i - \xi_{i-1} - \dots - \xi_1))
\end{split}\nonumber\\
\begin{split}
& = \{ \mbox{$h(x) = f(S_t^{-1} (x-x_2) + x_1)$ according to (\ref{eq-general_scaling_transformation}) } \}
\end{split}\nonumber\\
\begin{split}
& = f(S_t^{-1} S_s^{-1} ((x - \xi_i - \xi_{i-1} - \dots - \xi_1) - S_s x_2 + S_s S_t x_1)
\end{split}\nonumber\\
\begin{split}
& = \{ \mbox{$S_s S_t = S_{st}$ for scaling transformations} \}
\end{split}\nonumber\\
\begin{split}
& = f(S_{st}^{-1} ((x + S_s S_t x_1 - S_s x_2 - \xi_i - \xi_{i-1} - \dots - \xi_1))
\end{split}\nonumber\\
\begin{split}
& = \{ \mbox{$f_{st}(x) = f(S_{st}^{-1} x)$ according to definition (\ref{eq-scale-def2})} \}
\end{split}\nonumber\\
\begin{split}
& = f_{st}(x + S_s S_t x_1 - S_s x_2 - \xi_i - \xi_{i-1} - \dots - \xi_1).
\end{split}
\end{align}
Inserting this transformed integrand into the integral representation 
(\ref{eq-exp-int-repr-sc-cov-proof-noncent-offset})
gives
\begin{align}
\begin{split}
&  (\Gamma^{(i)} h)(x, s, c) =  
\end{split}\nonumber\\
\begin{split}
& = 
\sigma_{i,c} 
\left(
\sum_{m_i=1}^{M_{i-1}}
\int_{\xi_i \in \bbbr^N} 
\sigma_{i-1,m_i}
\left(
\sum_{m_{i-1}=1}^{M_{i-2}}
\int_{\xi_{i-1} \in \bbbr^N} 
\dots
\right.
\right.
\end{split}\nonumber\\
\begin{split}
& \phantom{\sigma_i \vphantom{\left( \sum_{m_i=1}^{M_{i-1}} \right.)}} \quad
\left.
\left.
\sigma_{1,m_2}
\left(
\sum_{m_1=1}^{M_0}
\int_{\xi_1 \in \bbbr^N} 
f_{st}(x + S_s S_t x_1 - S_s x_2  -
\right.
\right.
\right.
\end{split}\nonumber\\
\begin{split}
& \hphantom{\sigma_i \vphantom{\left( \sum_{m_i=1}^{M_{i-1}} \right.)}} \quad
\left.
\left.
\hphantom{\sigma_{1,m_2} \left(  \sum_{m_1=1}^{M_0}
	\int_{\xi_1 \in \bbbr^N} \right. } \quad\quad
\xi_i - \xi_{i-1} - \dots - \xi_1) \times
\right.
\right.
\end{split}\nonumber\\
\begin{split}
& \phantom{\sigma_i \vphantom{\left( \sum_{m_i=1}^{M_{i-1}} \right.)}} \quad
\left.
\left.
\left.
\phantom{\left(\sum_{m_i=1}^{M_{i-1}} \right.)}
g_{m_1,m_2}^{(1)}(\xi_1)  \, d\xi_1
\right) \dots
g_{m_{i-1},m_i}^{(i-1)}(\xi_{i-1})  \, d\xi_{i-1}
\vphantom{\left( \sum_{m_i=1}^{M_{i-1}} \right.)} 
\right)
\right.
\end{split}\nonumber\\
\begin{split}
\left.
\vphantom{\left(\sum_{m_i=1}^{M_{i-1}} \right.)}\quad\quad
g_{m_i,c}^{(i)}(\xi_i)  \, d\xi_i
\vphantom{\left( \sum_{m_i=1}^{M_{i-1}} \right.)} \right),
\end{split} 
\end{align}
which we recognise as
\begin{align}
\begin{split}
&  (\Gamma^{(i)} h)(x, s, c) 
\end{split}\nonumber\\
\begin{split}
& = (\theta^{(i)} \theta^{(i-1)} \dots \theta^{(2)} \theta^{(1)} f_{st})(x + S_s S_t x_1 - S_s x_2, c) 
\end{split}\nonumber\\
\begin{split}
& = (\phi^{(i)} \, f_{st})(x + S_s S_t x_1 - S_s x_2, c) 
\end{split}\nonumber\\
\begin{split}
&  = (\phi_{st}^{(i)} f)(x + S_s S_t x_1 - S_s x_2, c) 
\end{split}\nonumber\\
\begin{split}
&=  (\Gamma^{(i)} f)(x + S_s S_t x_1 - S_s x_2, st, c)
\end{split}
\end{align}
and which proves the result. Note that for a pure translation ($S_t = I$,  $x_1 = x_0 $  and $x_2 = x_0 + \delta$) this gives
\begin{align}
&(\Gamma^{(i)}\, \calD_\delta\, f)(x,c,s) = (\Gamma^{(i)} f)(x - S_s \delta, s, c).
\label{eq-translation-covariance-scale}
\end{align}
Thus, translation covariance is preserved in the scale-channel network
but the magnitude of the spatial shift in the feature maps will depend
on the scale channel. The discrete implementation and some additional design choices for discrete scale-channel networks are discussed in Section \ref{sec-discrete-scale-channels}, but, first, we will consider the relationship between continuous scale-channel networks and scale-space theory. 

\section{Relations between scale-channel networks and scale-space theory}
\label{sec-relations-scsp-theory}

This section describes relations between the presented scale-channel networks and concepts in scale-space theory, specifically (i)~a mapping between scaling the input image using multiple scaling factors, as used in scale-channel networks, or instead scaling the filters multiple times, as done in scale-space theory, and (ii)~a relationship to the normalisation over scales of scale-normalised derivatives, which holds if the learning algorithm for a scale-channel network would learn filters corresponding to Gaussian derivatives.

\subsection{Preliminaries 1: The Gaussian scale space}
\label{sec:gauss-prel}

In classical scale-space theory 
\cite{Iij62,Wit83,Koe84,KoeDoo92-PAMI,Lin93-Dis,Lin94-SI,Flo97-book,WeiIshImi99-JMIV,Haa04-book,DuiFloGraRom04-JMIV,Lin10-JMIV},
a multi-scale representation of an input image is created by
convolving the image with a set of rescaled and normalised Gaussian
kernels. The resulting {\em scale-space representation\/}
of an input image $f: \mathbb{R}^N \to \mathbb{R}$ is defined as
\cite{Lin93-Dis}:
\begin{equation}
L(x; \sigma) = \int_{u \in \bbbr^N} f(x - u)\,g(u; \sigma)\,du,
\end{equation}
where $g: \bbbr^N\times\bbbr^+ \to \bbbr$ denotes the (rotationally symmetric) Gaussian kernel
\begin{equation}
g(x;\sigma) =  \frac{1}{(\sqrt{2\pi} \sigma)^{N}}  e^{\frac{-x^2}{2 \sigma^2}},
\end{equation}
and we use $\sigma$ as the {\em scale parameter} compared to the
more commonly used $t = \sigma^2$. The original image/function is thus embedded into a family of functions parameterised by scale. The scale-space representation is scale covariant and the representation of an original image can be matched to that of a rescaled image by a spatial rescaling and a multiplicative shift along the scale dimension.
From this representation, a family of {\em Gaussian derivatives} can
be computed as
\begin{equation}
L_{x^{\alpha}}(x;\sigma) = \partial_{x^\alpha} L(x;\sigma) = ((\partial_{x^{\alpha}} g(\cdot;\; \sigma)) * f(\cdot))(x),
\end{equation}
where $n \in \mathbb{Z}$ and we use multi index notation 
$\alpha = (\alpha_1, \cdots \alpha_N)$ such that
$\partial_{x^\alpha} = \partial_{x^{\alpha^1}}
\cdots \partial_{x^{\alpha^N}}$. 

The \emph{scale covariance} property also transfers to such Gaussian derivatives, and 
these visual primitives have been widely used within the classical
computer vision paradigm to construct scale-covariant and
scale-invariant feature detectors and image descriptors
\cite{Lin97-IJCV,Lin98-IJCV,BL97-CVIU,ChoVerHalCro00-ECCV,MikSch04-IJCV,Low04-IJCV,BayEssTuyGoo08-CVIU,TuyMik08-Book,Lin13-ImPhys,Lin15-JMIV}. 

\subsection{Scaling the image {\em vs.\/}\ scaling the filter}
\label{sec:scaling-image-v-filter}

The scale-channel networks described in this paper are
based on a similar philosophy of processing an image at 
{\em all scales simultaneously\/}, although {\em the input image\/},
as opposed to the filter, is expanded over scales. We, here, consider
the relationship between multi-scale representations computed by
applying a set of {\em rescaled kernels\/} to a single-scale image and
representations computed by applying the same kernel to a set of
{\em rescaled images\/}. Since the scale-space representation can be computed using a single
convolutional layer, we compare with a single-layer
scale-channel network.
We consider the relationship between representations computed by:  
\begin{enumerate}[(i)]	
	\item 
	Applying a set of rescaled and scale-normalised filters (this
	corresponds to normalising filters to constant $L_1$-norm over
	scales) $h: \mathbb{R}^N \to \mathbb{R}$ 
	\begin{equation}\label{eq-scaled_and_normalised_filters}
	h_s(x) = \frac{1}{s^{N}} h(\frac{x}{s})
	\end{equation}
	to a fixed size input image $f(x)$: 
	\begin{equation}\label{eq-Lh}
	L_h(x;s) = (f*h_s)(x) = \int_{u \in \bbbr^N} f(u)\,h_s(x - u)\,du,
	\end{equation}
	where the subscript indicates that $h$ might not necessarily be
	a Gaussian kernel. If $h$ is a Gaussian then $L_h = L$. 
	\item 
	Applying a fixed size filter $h$ to a set of rescaled input images 
	\begin{equation}
	\label{eq-Mh}
	M_h(x;s) = (f_s*h)(x) = \int_{u \in \bbbr^N} f_s(u)\,h(x - u)\,du,
	\end{equation}
	with
	\begin{equation}
	f_s(x) = f(\frac{x}{s}).
	\end{equation}
	This is the representation computed by a single layer in a
	(continuous) scale-channel network. 
\end{enumerate}
It is straightforward to show that these representations are
computationally equivalent and related by a family of scale dependent
scaling transformations.
We compute using the change of variables $u =s\,v$, $du = s^{N} dv$:
\begin{align}\label{eq-filter-to-image}
L_h(x;s) &=  (f*h_s)(x) \nonumber \\
&= \int_{u \in \bbbr^N} f(x-u)\, \frac{1}{s^{N}}h(\frac{u}{s})\,du \nonumber \\
&= \int_{u \in \bbbr^N} f(x - s v )\, \frac{1}{s^{N}}h(v)\, s^{N}dv \nonumber \\
&= \int_{u \in \bbbr^N} f(s(\frac{x}{s}-v))\, h(v)\,dv \nonumber \\
&= \int_{u \in \bbbr^N} f_{s^{-1}}(\frac{x}{s}-v)\,h(v)\,dv \nonumber \\
&= (f_{s^{-1}} * h)(\frac{x}{s},s^{-1}).
\end{align}
Comparing this with (\ref{eq-Mh}),
we see that the two representations are related according to 
\begin{equation}\label{eq-l_to_m}
L_h(x;s) = M_h(\frac{x}{s};s^{-1}).
\end{equation}
We note that the relation (\ref{eq-l_to_m}) preserves \emph{the
	relative scale} between the filter and the image for each scale and
that both representations are scale covariant.
Thus, to convolve a set of rescaled images with a single-scale filter, is computationally equivalent
to convolving an image with a set of rescaled filters that are
$L_1$-normalised over scale. The two representations are related
through a \emph{spatial rescaling} and an \emph{inverse mapping of the
	scale parameter} $s \mapsto s^{-1}$.
Note that it is straightforward to show, using the integral
representation of a scale-channel network (\ref{eq-phi_integral}),
that a corresponding relation between scaling the image and scaling
the filters holds for a multi-layer scale-channel network as well. 

The result (\ref{eq-l_to_m}) implies that if a scale-channel network
learns a feature corresponding to a Gaussian with standard
deviation $\sigma$,
then the representation computed by the scale-channel network is
computationally equivalent to applying the family of kernels
\begin{equation}
h_s(x) = \frac{1}{s^N} h(\frac{x}{s}) = \frac{1}{(\sqrt{2\pi} s\sigma)^{N}}  e^{\frac{-x^2}{2 (s\sigma)^2}}
\end{equation}
to the original image, given the complementary scaling transformation
(\ref{eq-l_to_m}) with its associated inverse mapping of
the scale parameters $s \mapsto s^{-1}$.
Since this is a family of rescaled and $L_1$-normalised Gaussians, the
scale-channel network will compute a representation computationally
equivalent to a Gaussian scale-space representation. 
For discrete image data, a similar relation holds approximately,
provided that the discrete rescaling operation is a sufficiently good approximation of the continuous rescaling operation. 

\subsection{Relation between scale-channel networks and scale-normalised derivatives}

One way to achieve scale invariance within the Gaussian scale-space concept is to first perform \emph{scale
	selection}, {\em i.e.\/}, identify the relevant scale/scales, and then, {\em e.g.\/}, extract features at the identified
scale/scales. Scale selection can be done by comparing the magnitudes of
$\gamma$-normalised derivatives \cite{Lin97-IJCV,Lin98-IJCV}:
\begin{equation}
\partial_{\xi^{\alpha}} = \partial_{x^\alpha, \gamma-norm} = t^{|\alpha|\gamma/2} \, \partial_{x^{\alpha}} = \sigma^{|\alpha| \gamma} \, \partial_{x^{\alpha}} 
\end{equation}
over scales with $\gamma \in [0,1 ]$ as a free parameter and $|\alpha| = \alpha_1
+ \cdots + \alpha_N$. Such derivatives are likely to take maxima
at scales corresponding to the relevant physical scales of objects in
the image.
Although a multi-layer scale-channel network will compute more
complex {\em non-linear\/} features, it is enlightening to investigate whether
the network can learn to express operations similar to scale-normalised derivatives. This would increase our
confidence that scale-channel networks could be expected to work well
together with, {\em e.g.\/}, max pooling over scales.

We will, here, consider the maximally scale-invariant case for scale-normalised derivatives with $\gamma=1$ 
\begin{equation}
\partial_{\xi^{\alpha}} = \sigma^{|\alpha|}  \partial_{x^\alpha}.
\end{equation}
and show that scale-channel networks can indeed learn features equivalent to
such scale-normalised derivatives. 

\subsubsection{Preliminaries II: Gaussian derivatives in terms of Hermite polynomials}

As a preparation for the intended result, we will first establish a
relationship between Gaussian derivatives and probabilistic Hermite
polynomials.
The probabilistic Hermite polynomials $H e_n(x)$ are in 1-D defined by the
relationship
\begin{equation}
H e_n(x) = (-1)^n e^{x^2/2} \, \partial_{x^n} \left( e^{-x^2/2} \right)
\end{equation}
implying that
\begin{equation}
\partial_{x^n} \left( e^{-x^2/2} \right) =  (-1)^n H e_n(x) \, e^{-x^2/2} 
\end{equation}
and 
\begin{equation}
\partial_{x^n} \left( e^{-x^2/2\sigma^2} \right) =  (-1)^n H e_n(\frac{x}{\sigma}) \, e^{-x^2/2\sigma^2} \frac{1}{\sigma^n}.
\end{equation}
Applied to a Gaussian function in 1-D, this implies that
\begin{align}
\begin{split}
\partial_{x^n} \left( g(x;\; \sigma) \right) =
\end{split}\nonumber\\
\begin{split}
= \frac{1}{\sqrt{2 \pi} \sigma} \partial_{x^n} \left( e^{-x^2/2\sigma^2} \right) 
\end{split}\nonumber\\
\begin{split}
= \frac{1}{\sqrt{2 \pi} \sigma}  \frac{(-1)^n}{\sigma^n}  H e_n(\frac{x}{\sigma}) \, e^{-x^2/2\sigma^2}
\end{split}\nonumber\\
\begin{split}
\label{eq-gauss-der-herm-pol}
= \frac{(-1)^n}{\sigma^n} H e_n(\frac{x}{\sigma}) \, g(x;\; \sigma).
\end{split}
\end{align}

\subsubsection{Scaling relationship for Gaussian derivative kernels}
\label{sec:relation-to-scale-space2}

We, here, describe the relationship between scale-channel networks and
scale-normalised derivatives. Let us assume that the scale-channel network at some layer learns a kernel that corresponds to a Gaussian partial derivative at
some scale $\sigma$:
\begin{align}
\begin{split}
\partial_{x^{\alpha}} g(x;\; \sigma) =
\end{split}\nonumber\\
\begin{split}
= \partial_{x_1^{\alpha_1} x_2^{\alpha_2} \dots x_N^{\alpha_N}}
g(x;\; \sigma) 
= g_{x_1^{\alpha_1} x_2^{\alpha_2} \dots x_N^{\alpha_N}}(x;\; \sigma).
\end{split}
\end{align}
We will show that when this kernel is applied to all the scale
channels, this correspond to a normalisation over scales that is equivalent to scale-normalisation of Gaussian derivatives. 

For later convenience, we write this learned kernel as a scale-normalised
derivative at scale $\sigma$ for $\gamma = 1$ multiplied by some constant $C$:
\begin{equation}
h(x) = C \, \sigma^{\alpha_1 + \alpha_2 + \dots + \alpha_N} g_{x_1^{\alpha_1} x_2^{\alpha_2} \dots x_N^{\alpha_N}}(x;\; \sigma).
\end{equation}
Then, the corresponding family of equivalent kernels $h_s(x)$ in the dual
representation (\ref{eq-Lh}), 
which represents the same effect on the original image as applying the kernel $h(x)$ 
to a set of rescaled images $f_s(x) = f(x/s)$,
provided that a 
complementary scaling transformation and the inverse mapping of the
scale parameter $s \mapsto s^{-1}$ are performed, is given by
\begin{align}
\begin{split}
h_s(x) 
= \frac{1}{s^N} \, h(\frac{x}{s})
\end{split}\nonumber\\
\begin{split}
= \frac{C}{s^N} \, \sigma^{\alpha_1 + \alpha_2 + \dots + \alpha_N} g_{x_1^{\alpha_1} x_2^{\alpha_2} \dots x_N^{\alpha_N}}(\frac{x}{s};\; \sigma).
\end{split}
\end{align}
Using Equation~(\ref{eq-gauss-der-herm-pol}) with 
\begin{equation}
g(x;\; \sigma) = \frac{1}{(\sqrt{2 \pi} \sigma)^N} \, e^{-(x_1^2 + x_2^2 + \dots + x_N^2)/2\sigma^2} 
\end{equation}
in $N$ dimensions, we obtain
\begin{align}
\begin{split}
h_s(x) =\frac{C}{s^N} \, \sigma^{\alpha_1 + \alpha_2 + \dots + \alpha_N}
(-1)^{\alpha_1 + \alpha_2 + \dots + \alpha_N}
\end{split}\nonumber\\
\begin{split}
\phantom{=}
He_{\alpha_1}(\frac{x_1}{s\sigma}) \,  He_{\alpha_2}(\frac{x_2}{s\sigma}) \dots He_{\alpha_N}(\frac{x_N}{s\sigma}) 
\end{split}\nonumber\\
\begin{split}
\phantom{=}
\frac{1}{(\sqrt{2 \pi} \sigma)^N} \, e^{-(x_1^2 + x_2^2 + \dots + x_N^2)/2s^2\sigma^2} 
\frac{1}{\sigma^{\alpha_1 + \alpha_2 + \dots + \alpha_N} } 
\end{split}\nonumber\\
\begin{split}
=C \, (s \sigma)^{\alpha_1 + \alpha_2 + \dots + \alpha_N}
(-1)^{\alpha_1 + \alpha_2 + \dots + \alpha_N}
\end{split}\nonumber\\
\begin{split}
\phantom{=}
He_{\alpha_1}(\frac{x_1}{s\sigma}) \,  He_{\alpha_2}(\frac{x_2}{s\sigma}) \dots He_{\alpha_N}(\frac{x_N}{s\sigma}) 
\end{split}\nonumber\\
\begin{split}
\phantom{=}
\frac{1}{(\sqrt{2 \pi} s \sigma)^N} \, e^{-(x_1^2 + x_2^2 + \dots + x_N^2)/2s^2\sigma^2} 
\frac{1}{(s \sigma)^{\alpha_1 + \alpha_2 + \dots + \alpha_N} }.
\end{split}
\end{align}
Comparing with (\ref{eq-gauss-der-herm-pol}), we recognise this
expression as the scale-normalised derivative
\begin{equation}
h_s(x) = C \, (s \sigma)^{\alpha_1 + \alpha_2 + \dots + \alpha_N}
g_{x_1^{\alpha_1} x_2^{\alpha_2} \dots x_N^{\alpha_N}}(x;\; s \sigma)
\end{equation}
of order $\alpha = (\alpha_1, \alpha_2, \dots \alpha_N)$ at scale $s
\sigma$.

This means that if the scale-channel network learns a partial Gaussian
derivative of some order, then the application of that filter to all
the scale channels is computationally equivalent to 
{\em applying corresponding scale-normalised Gaussian derivatives\/}
to the original image at all scales.

While this result has been expressed for partial derivatives, a
corresponding results holds also for derivative operators that
correspond to directional derivatives of Gaussian kernels in arbitrary
directions.
This result can be easily understood from the expression for a
directional derivative operator $\partial_{e^n}$ of order 
$n = n_1 + n_2 + \dots + n_N$ 
in direction $e = (e_1, e_2, \dots, e_N)$ with 
$|e| = \sqrt{e_1^2 + e_2^2 + \dots + e_N^2} = 1$:
\begin{align}
\begin{split}
\partial_{e^n} g(x;\; \sigma) 
\end{split}\nonumber\\
\begin{split}
= (e_1 \, \partial_{x_1} + e_2 \, \partial_{x_2} + \dots +  e_{N} \, \partial_{x_N})^n g(x;\; \sigma)
\end{split}\nonumber\\
\begin{split}
= \sum_{\alpha_1 + \alpha_2 + \dots + \alpha_N = n}
{n \choose \alpha_1! \, \alpha_2! \, \dots \, \alpha_N!} 
\end{split}\nonumber\\
\begin{split}
\quad\quad\quad\quad\quad\quad
e_1^{\alpha_1} e_2^{\alpha_2} \dots e_N^{\alpha_N} \,
\partial_{x_1}^{\alpha_1} \partial_{x_2}^{\alpha_2} \dots \partial_{x_N}^{\alpha_N} 
g(x;\; \sigma)
\end{split}\nonumber\\
\begin{split}
= \sum_{\alpha_1 + \alpha_2 + \dots + \alpha_N = n}
{n \choose \alpha_1! \, \alpha_2! \, \dots \, \alpha_N!} 
\end{split}\nonumber\\
\begin{split}
\quad\quad\quad\quad\quad\quad
e_1^{\alpha_1} e_2^{\alpha_2} \dots e_N^{\alpha_N} \,
g_{x_1^{\alpha_1} x_2^{\alpha_2} \dots x_N^{\alpha_N}}(x;\; \sigma).
\end{split}
\end{align}
Since the scale normalisation factors $\sigma^{|\alpha|}$
for all scale-normal\-ised partial derivatives
of the same order $|\alpha| = \alpha_1 + \alpha_2 + \dots + \alpha_N = n$ are the same, it follows that all linear
combinations of partial derivatives of the same order are 
transformed by the same multiplicative scale normalisation factor, which
proves the result.

\subsection{Relations to classical scale selection methods}

Specifically, the scaling result for Gaussian derivative kernels implies that 
a scale-channel network that
combines the multiple scale channels by {\em supremum\/}, or for a discrete set of scale channels,
{\em max pooling\/} (see further Section~\ref{sec-discrete-scale-channels}), will be structurally similar to classical methods for
{\em scale selection\/}, which detect maxima over scale of
scale-normalised filter responses
\cite{Lin97-IJCV,Lin98-IJCV,Lin21-EncCompVis}. In the scale-channel networks, max pooling is,
however, done over more complex feature responses, already adapted to
detect specific objects, while classical scale selection is performed
in a class-agnostic way based on low-level features. This makes
max pooling in the scale-channel networks also closely related to more
specialised classical methods that detect maxima from the scales at
which a supervised classifier delivers class labels with the highest
posterior \cite{LiTaxLoo11-ScSp,LooLiTax09-LNCS}. 
Average pooling over the outputs of a discrete set of scale channels
(Section~\ref{sec-discrete-scale-channels}) is structurally similar to
methods for scale selection that are based on {\em weighted averages\/}
of filter responses at different scales
\cite{Lin12-JMIV,Lin15-JMIV}. Although there is no
guarantee that the learned non-linear features will, indeed, take
maxima for relevant scales, one might expect training to promote this,
since a failure to do so should be detrimental to the classification
performance of these networks.

\section{Discrete scale-channel networks}
\label{sec-discrete-scale-channels}

Discrete scale-channel networks are implemented by using a standard
discrete CNN as the base network $\phi$. For practical applications,
it is also necessary to restrict the network to include a finite
number of scale channels 
\begin{equation}
   \hat{S} = \{ \gamma^{i}\}_{-K_{min} \leq i   \leq K_{max} }. 
\end{equation}
The input image $f:\bbbz^2 \to \bbbr$ is assumed to
be of finite support. 
The outputs from the scale channels are, here,  
aggregated using, {\em e.g.\/}, max pooling
\begin{equation}
(\Lambda_{\max}f)(x,c) = \max_{s \in \hat{S}} \{(\phi_s f)(x,c,s) \} 
\label{eq-lambda-s-max}
\end{equation}
or average pooling
\begin{equation}
(\Lambda_{\avg} f)(x,c)= \mathop{\avg}_{s \in \hat{S}} \{(\phi_s f)(x,c,s) \}.
\label{eq-lambda-s-avg}
\end{equation}
We will also implement discrete scale-channel networks that
concatenate the outputs from the scale channels, followed by an
additional transformation $\varphi: \bbbr^{M_i |\hat{S}|} \to
\bbbr^{M_i}$ that mixes the information from the different channels
\begin{align} 
&(\Lambda_{\conc} f)(x, c) \nonumber \\
&=  \varphi \left( [(\phi_{s_1} f)(x,c), (\phi_{s_2} f)(x,c) \cdots  (\phi_{s_{|\hat{S}|}} f)(x,c) ] \right).
\label{eq-lambda-s-conc}
\end{align}
$\Lambda_{\conc}$ does not have any theoretical guarantees of invariance, but 
since scale concatenation of outputs from the scale channels has been
previously used with the explicit aim of scale-invariant recognition
\cite{XuXiaZhaYanZha14-arXiv}, we will evaluate that approach also here.

\subsection{Foveated processing}
\label{sec-foveated-operations}

A standard convolutional neural network $\phi$ has a finite support region $\Omega$ in the input. 
When rescaling an input image of fixed size/finite support in the
scale channels, it is necessary to decide how to process the resulting
images of varying size using a feature extractor with fixed
support. One option is to process regions of {\em constant size\/} in
the scale channels, corresponding to regions of {\em different sizes\/}
in the input image.
This results in {\em foveated image operations\/}, where a smaller
region around the center of the input image is processed at high
resolution, while gradually larger regions of the input image are
processed at gradually reduced resolution (see
Figures~\ref{fig-foveated-processing}(b)-(c)). Note how this implies that the scale channels will together process a covariant set of regions, so that for any object size there is always a scale channel with a support matching the size of the object.
We will refer to the foveated network architectures $\Lambda_{\max} $,
$\Lambda_{\avg} $ and $\Lambda_{\conc} $ as the FovMax network, the
FovAvg network and the FovConc network, respectively.

\begin{figure*}[hbtp]
	\begin{center}
		\includegraphics[width=0.9\textwidth]{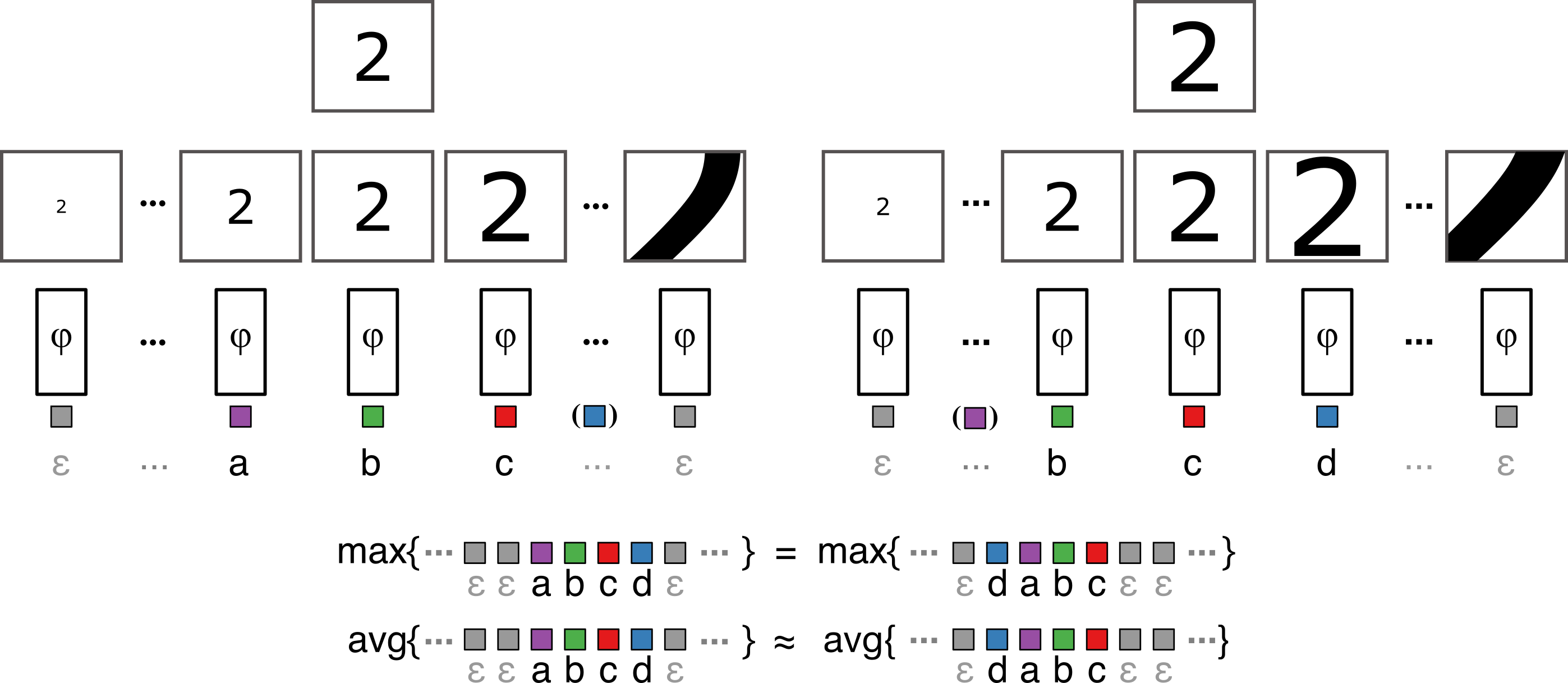} 
	\end{center}
	\caption{{\em An illustration of how discrete scale-channel networks approximate scale invariance over a finite scale range}. Consider a foveated scale-channel network combined with max or average pooling over the output from the scale channels. Since the same operation is performed in all the scale channels, when comparing the output for an original image (left) and a rescaled copy of this image (right), we see that the output code is just \emph{shifted} along the scale dimension. Thus, if the values taken at the edge of the scale range are small enough, then the maximum over scales will still be preserved between an original and a rescaled image. Correspondingly, for average pooling, there will in this case be no significant change of the mass of the feature response within the scale range spanned by the scale channels. Here, we illustrate the idea for a network that produces a scalar output, but the same argument is valid for vector valued output, where the only difference is that the pooling over the scale dimension is performed for each vector element separately.}
	\label{fig-visual-proof}
\end{figure*}

\begin{figure*}[hbtp]
	\begin{center}
		\includegraphics[width=1.0\textwidth]{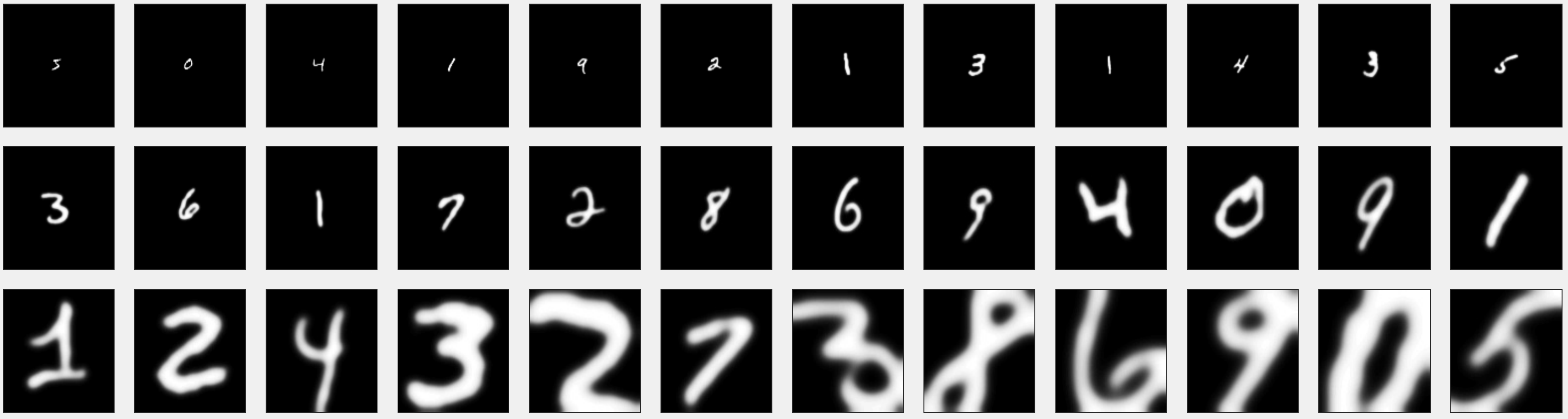} 
	\end{center}
	\caption{{\em Samples from the MNIST Large Scale dataset}: The
          MNIST Large Scale dataset is derived from the original MNIST
          dataset \cite{LecBotBenHaf98-ProcIEEE} and contains $112
          \times 112$ sized images of handwritten digits with scale
          variations of a factor of 16. The scale factors relative the
          original MNIST dataset are in the range $\frac{1}{2}$ (top
          left) to $8$ (bottom right).}
	\label{fig-dataset-mnist-scale}
\end{figure*}

\subsection{Approximation of scale invariance}

Foveated processing combined with max or average pooling will give an
approximation of scale invariance in the continuous model
(Section~\ref{sec-scale-invariance}) over {\em a limited scale range\/}. 
The numerical scale warpings of the input images in the
scale channels approximate continuous scaling transformations. A
discrete set of scale channels will approximate the representation for
a continuous scale parameter, where the approximation will be better
with a denser sampling of the scaling group. 

A possible source of problems will, however, arise due to boundary
effects caused by a finite scale interval. 
True scale invariance is only guaranteed for an infinite number of scale channels. In the case of max pooling over a finite set of scale channels, there is a risk that the maximum value over the scale channels moves in or out of the finite scale range covered by the scale channels. Correspondingly, for average pooling, there is a risk that a substantial part of
mass of the feature responses from the different
scale channels may move in or out of a finite scale interval.
The risk for such boundary effects would, however, be mitigated if the
network learns to suppress responses for both very zoomed in and
very zoomed out objects, so that the contributions from such image
structures are close to zero. 
As a design criterion for scale-channel networks, we therefore propose
to include at least a small number of scale channels both below
and above the effective training scales of the relevant image
structures. Further, we suggest \emph{training the network from scratch} as opposed to using pretrained weights for the scale channels. Then, we propose
that it should be likely that the network will learn to suppress responses for image structures that are far off in scale, since
the network would otherwise classify based on use of object views that
will hardly provide any useful information. An illustration providing
the intuition for how invariance can be achieved in the discrete
scale-channel networks is presented in Figure~\ref{fig-visual-proof}.

\subsection{Sliding window processing in the scale channels}
\label{sec-sliding-window}

An alternative option for dealing with varying image sizes is to, in
each scale channel, process the entire rescaled image by applying the
base network in {\em a sliding window manner\/}.
We, here, evaluate this option, but instead of evaluating
{\em the full network} anew at each image position,
we slide the classifier part of the network
({\em i.e.\/}, the last layer)
across the convolutional feature map.
This is considerably less computationally expensive and,
in the case of a network without subsampling by means of
strided convolutions (or max pooling),
the two approaches are equivalent.
Since strided convolution is used in the network,
it implies that we here trade some resolution in the
output for computational efficiency,
where it can be noted that a similar choice is made
in the OverFeat detector \cite{SerEigZhaMatFerLeC13-arXiv}.%
\footnote{A main difference between the OverFeat detector
  \cite{SerEigZhaMatFerLeC13-arXiv}
and our approach, however, is 
that the OverFeat detector uses a total effective stride of 32, 
whereas our network has a total effective stride of 4 (2 convolutional 
layers with stride 2 each). Because of 
the larger effective stride in the OverFeat detector, 
they apply their subsampling operation 
for every spatial offset in the last convolutional layer,
whereas we with our smaller effective stride do not 
need to, since the subsampled image representations are still at a 
satisfactory resolution.}

Concerning max pooling over space {\em vs.\/} over scale,
where according to the most original formulation, a sliding window
approach in a scale-space setting
would mean that the base network that performs integration
over scale should be applied and evaluated anew at all the visited
image positions, we, again for reasons of computational efficiency,
swap the ordering between max pooling
over space {\em vs.\/} over scale, and perform the max pooling over space before
the max pooling over scale, since we can then avoid the need for
incorporating an explicit mechanism for a
skewed/non-vertical pooling operation between corresponding image points
at different levels of scale according to (\ref{eq-lambda_s-skewed-def}).

The output from the scale channels can then be combined by max (or
average) pooling over space followed by max (or average) pooling over
scales%
\footnote{Concerning images of finite size, we make use of all the available image data
for computing the scale-channel representations used for the sliding
window approach, implying that more pixels are processed at a fine
scale compared to a coarse scale. This is in contrast to the
foveated representations, which are based on using the same number
pixels in the scale channels for every resolution.}
\begin{equation}
(\Lambda_{sw,\max} f)(c) = \max_{s \in S} \max_{x \in \Omega} \{ (\phi_s f)(x, c, s)  \}.
\label{eq-disc-sliding-window-pool}
\end{equation}
We will here only evaluate this architecture using max pooling only, which
is structurally similar to the popular multi-scale OverFeat detector
\cite{SerEigZhaMatFerLeC13-arXiv}. This network will be referred to as
the SWMax network. 

For this scale-channel network to support
invariance, it is not sufficient that boundary effects resulting from
using a finite number of scale channels are mitigated. When processing
regions in the scale channels corresponding to only a single region in
the input image, new structures can appear (or disappear) in this
region for a rescaled version of the original image. With a linear
approach, this might be expected to not cause problems,%
\footnote{When using linear template matching, the best matching
  pattern for a template learned during training will be a very
  similar image patch. Thus, when sliding a template across a matching
  object, it will take the maximum response when {\em centered\/} on
  the object. When using a non-linear method, however, there is no
  reason there could not be large responses for non centered views of
  familiar objects or completely novel patterns.}
since the best matching pattern will be the one corresponding to the template learned during training. 
For a deep neural network, however, there is no guarantee that there
cannot be strong erroneous responses for, {\em e.g.\/}, a partial view of a
zoomed in object. We are, here, interested in studying the effects
that this has on generalisation in the deep learning context. 

\begin{figure*}[h]
	\centering
	\subfloat[Subfigure 1 list of figures text][Standard CNN.]{
		\includegraphics[width=0.50\textwidth]{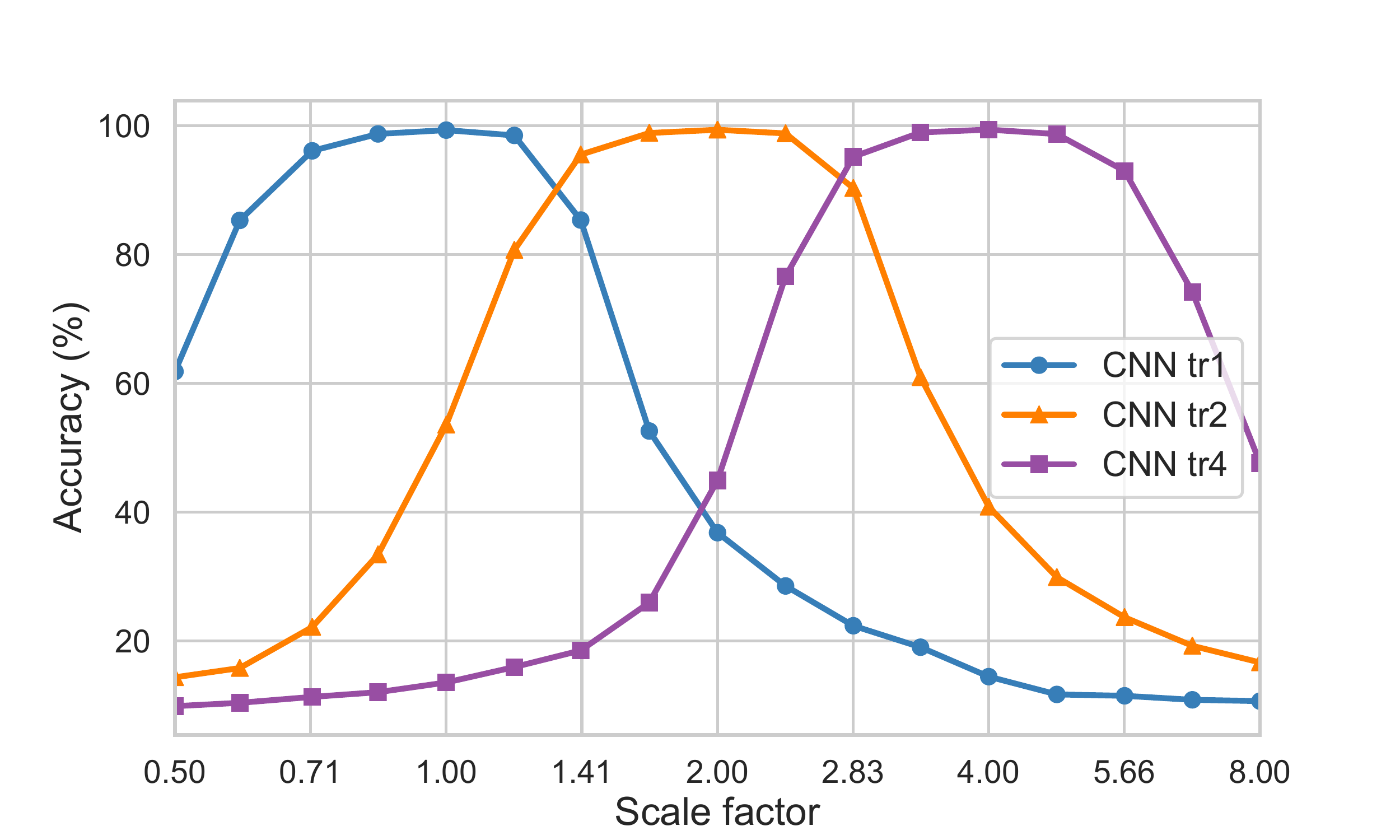}
		\label{fig-subfig1}
}
	\subfloat[Subfigure 2 list of figures text][The FovConc network.]{
		\includegraphics[width=0.50\textwidth]{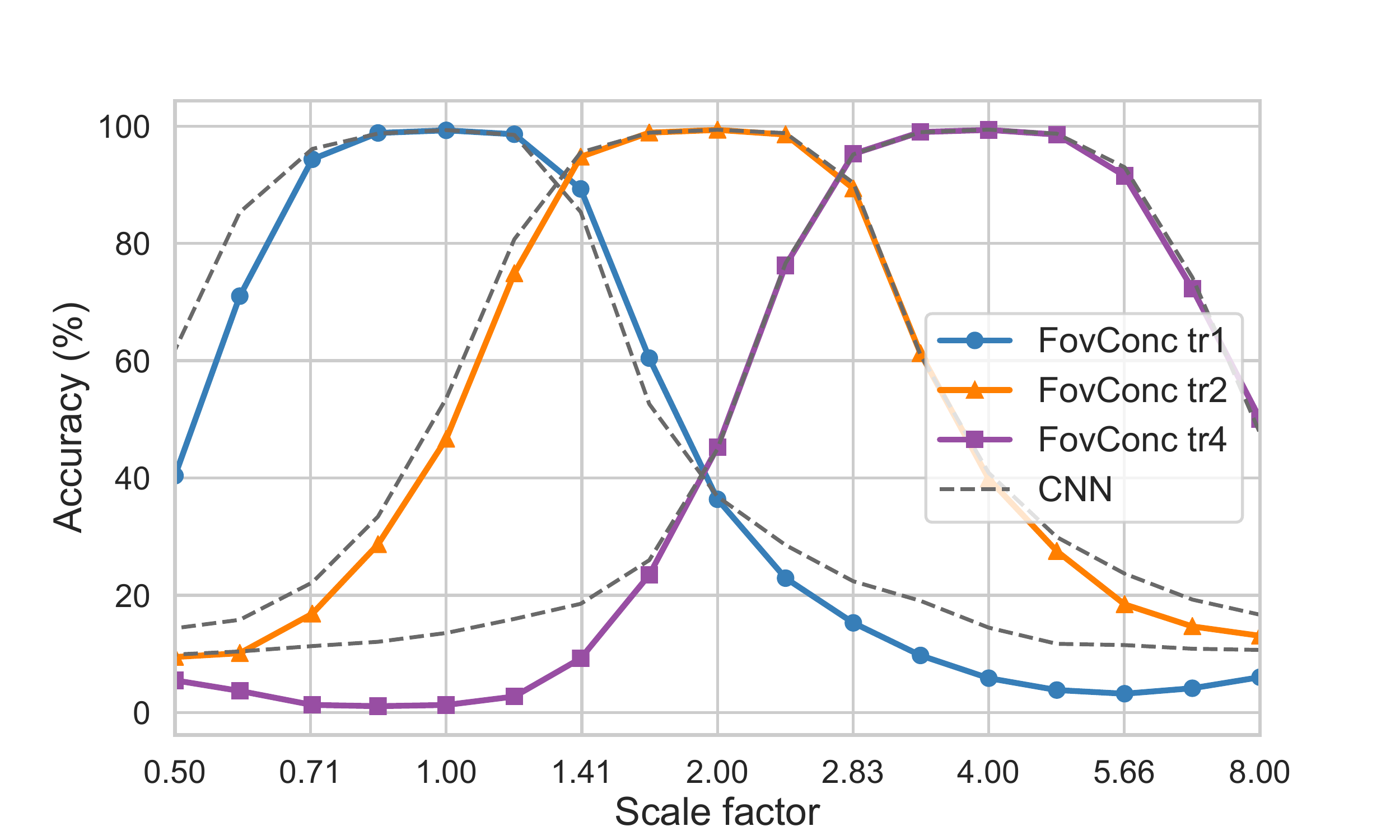}
		\label{fig-subfig2}
} \\
\vspace*{-1em}
	\subfloat[Subfigure 3 list of figures text][The FovMax and FovAvg networks]{
		\includegraphics[width=0.50\textwidth]{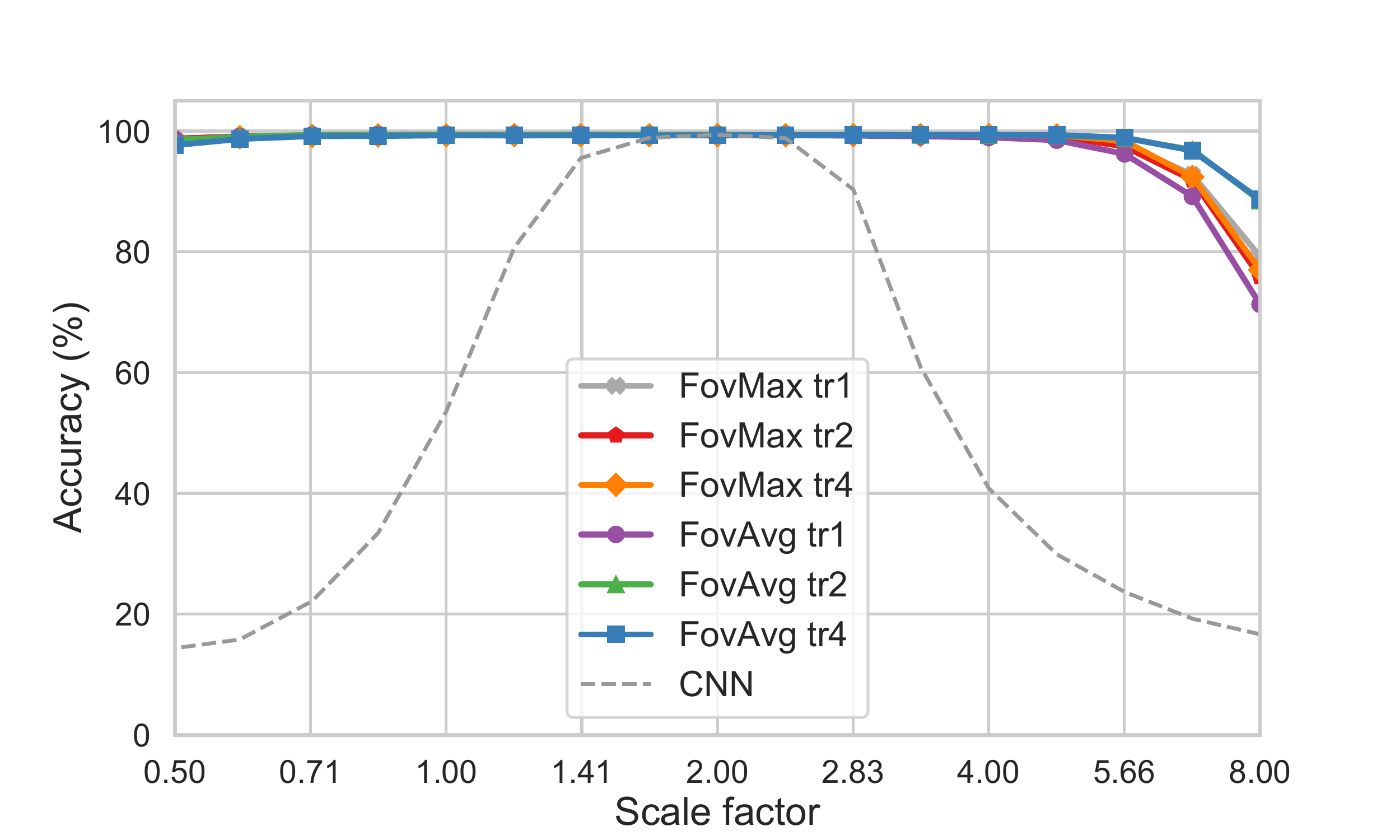}
		\label{fig-subfig3}
}
	\subfloat[Subfigure 4 list of figures text][The SWMax network]{
		\includegraphics[width=0.50\textwidth]{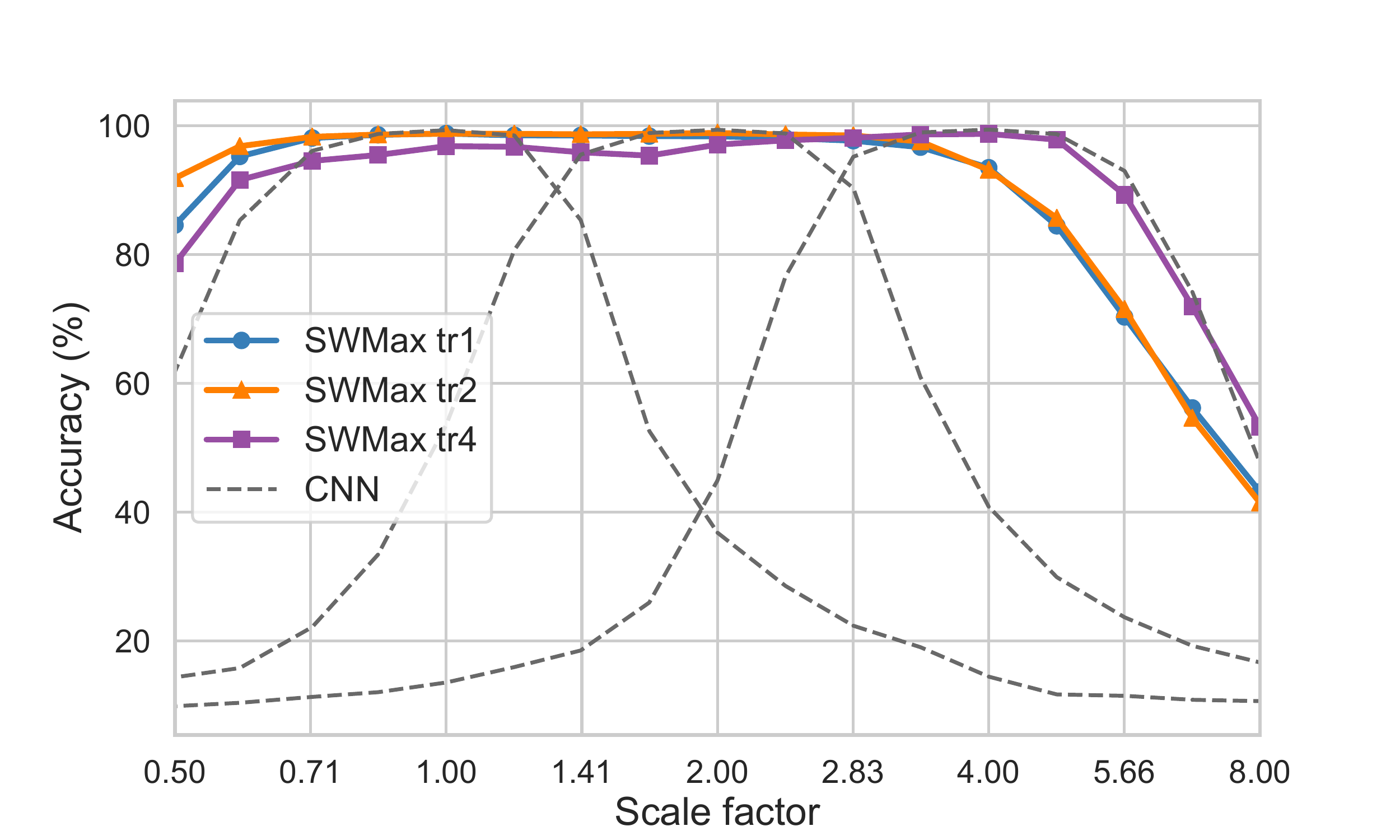}
		\label{fig-subfig4}
}
	\caption{{\em Generalisation ability to unseen scales for a
            standard CNN and the different scale-channel network
            architectures for the MNIST Large Scale dataset\/}. The networks are trained on digits of
          size~1 (tr1), size~2 (tr2) or size~4 (tr4) and evaluated
          for varying rescalings of the testing set. We note that the CNN
          (a) and the FovConc network (b) have poor generalisation
          ability to unseen scales, while the FovMax and FovAvg
          networks (c) generalise extremely well. The SWMax network
          (d) generalises considerably better than a standard CNN, but
          there is some drop in performance for scales not seen during
          training.}
	\label{fig-single-scale-generalisation}
\end{figure*}

\section{Experiments on the MNIST Large Scale dataset}
\label{sec-exp-MNISTLargeScale}

\subsection{The MNIST Large Scale dataset}
\label{sec-mnist-scale}

To evaluate the ability of standard CNNs and  scale-channel networks
to generalise to unseen scales over {\em a wide scale range\/}, we have
created a new version of the standard MNIST dataset
\cite{LecBotBenHaf98-ProcIEEE}. 
This new dataset, {\em MNIST Large Scale\/}, which is available online
\cite{JanLin20-MNISTLargeScale}, is composed of images of size 
$112 \times 112$  with scale variations of a factor 16 for 
scale factors $s \in [0.5, 8]$ relative to the original MNIST dataset  
(see Figure~\ref{fig-dataset-mnist-scale}). 
The training and testing sets for the different scale factors are created
by resampling the original MNIST training and testing sets using bicubic
interpolation followed by smoothing and soft thresholding to reduce
discretisation effects. Note that for scale factors $>4$, the full
digit might not be visible in the image. These scale values are
nonetheless included to study the limits of generalisation. More
details concerning this dataset are given in
Appendix~\ref{app-mnist-large-scale}.

\subsection{Network and training details}

In the experimental evaluation, we will compare five types of network
designs:
(i)~a (deeper) standard CNN 
(ii)~FovMax (max-pooling over the outputs from the scale channels),
(iii)~FovAvg (average pooling over the outputs from the scale channels),
(iv)~FovConc (concatenating the outputs from the scale channels) and
(v)~SWMax (sliding window processing in the scale channels combined with max-pooling over both space and scale).

{\em The standard CNN\/} is composed of 8 conv-batchnorm-ReLU blocks with $3 \times 3$ filters
followed by a fully connected layer and a final softmax layer. The
number of features/filters in each layer is
16-16-16-16-32-32-32-32-100-10. A stride of 2 is used in convolutional
layers 2, 4, 6 and 8. Note that this network is deeper and has more parameters than the networks used as base networks for the scale-channel networks. The reason for using a quite deep network is to
avoid a network structure that is heavily biased towards recognising
either small or large digits. A more shallow network would simply not have a receptive field large enough to enable recognising very large objects. The need for extra depth is thus a consequence of the scale preference built into a vanilla CNN architecture. Here, we are aware of this more structural problem of CNNs, but specifically aim to test scale generalisation for a network with a structure that would at least in principle enable scale generalisation.

{\em The FovMax, FovAvg, FovConc and SWMax \/}%
{\em scale-channel networks} are constructed using base networks for the scale channels with
4 conv-batchnorm-ReLU blocks with $3 \times 3$ filters followed by a fully connected layer and a
final softmax layer. The number of features/filters in each layer is 16-16-32-32-100-10. A
stride of 2 is used in convolutional layers 2 and 4. Rescaling within the scale channels is done with
bilinear interpolation and applying border padding or cropping as
needed. The batch normalisation layers are shared between the scale channels for the FovMax, FovAvg and FovConc networks. This implies that \emph{the same operation} is performed for all scales, to preserve scale covariance and enable scale invariance after max or average pooling.

We do not apply batch normalisation to the SW network, since this was shown to impair the performance. We believe that this is because the
  sliding window approach implies a \emph{change in the feature
  distribution} for this network when processing data of different sizes. 
  For the batch normalisation to function optimally, the data/feature 
  distribution should stay approximately the same, which is not the case for the SWMax network. 
        \footnote{Note that for the OverFeat detector \cite{SerEigZhaMatFerLeC13-arXiv} networks pretrained on ImageNet use a \emph{pretrained} base network which precludes the problem with training a sliding window scale-channel network with batch normalisation from scratch. For the larger scale ranges evaluated here, however, using networks with pretrained weights for the scale channels gives considerably worse generalisation performance.  We, here, tested two versions of batch normalisation: (i) normalising the feature responses jointly across all feature maps and (ii) normalising each channel separately. Neither of these options is scale invariant, the first because of the change in the feature distribution for the joint set of feature maps between inputs of different sizes and the second because the same operation is not applied for all feature channels. Both impaired the performance. We thus opt for evaluating the SWMax network with the best configuration we found, which corresponds to training the network from scratch without batch normalisation.}

For the FovAvg and FovMax networks, max pooling and average pooling, respectively, are performed across the logits outputs from the scale channels before the final softmax transformation and cross entropy loss. For the FovConc network, there is a fully connected layer that combines the logits outputs from the multiple scale channels before applying a final softmax transformation and cross entropy loss.

 All the scale-channel
architectures have around 70k parameters, whereas the baseline CNN
has around 90k parameters.

All the networks are trained with 50\,000 training samples from the
MNIST Large Scale dataset for 20 epochs using the Adam optimiser
with default parameters in PyTorch: $\beta_1 = 0.9$ and $\beta_2 = 0.999$.
During training, 15 \% dropout is applied to the first fully connected
layer. The learning rate starts at $3e^{-3}$ and decays with a factor
$1/e$ every second epoch towards a minimum learning rate of
$5e^{-5}$. For the SWMax network, the learning rate instead starts at $3e^{-4}$, since this produced better results in the absence of batch normalisation. Results are reported for the MNIST Large Scale testing set
(10\,000 samples) as the average of training each network using three
different random seeds. The remaining 10\,000 samples constitute a
validation set, which was used for parameter tuning. Parameter tuning was performed for a single-channel network, and the same parameters were used for the multi-channel networks and for the standard CNN.

Numerical performance scores for the results in
some of the figures 
to be reported are given in \cite{JanLin20-arXiv}.

\subsection{Generalisation to unseen scales}

We, first, evaluate the ability of the standard CNN and the different
scale-channel networks to generalise to previously unseen scales. We
train each network on either of the sizes 1, 2, and 4 
from the MNIST Large Scale dataset and evaluate the performance 
on the testing set for scale factors between $1/2$ and $8$. 
The FovMax, FovAvg and SWMax networks have 17 scale channels spanning
the scale range $[\frac{1}{2}, 8]$. The FovConc network has 3 scale
channels spanning the scale range $[1, 4]$.%
\footnote{The FovConc network has worse generalisation performance when
  including too many scale channels or spanning a too wide scale
  range. Since we are more interested in the best case rather than the
  worst case scenario, we, here, picked the best network out of a
  large range of configurations.} 
The results are presented in Figure~\ref{fig-single-scale-generalisation}. 
We, first, note that all the networks achieve similar top performance for
the scales seen during training. There are, however, large differences
in the abilities of the networks to generalise to unseen scales: 

\subsubsection{Standard CNN}

The standard CNN shows limited generalisation ability to unseen scales
with a large drop in accuracy for scale variations larger than a
factor $\sqrt{2}$. This illustrates that, while the network can
recognise digits of all sizes, a standard CNN includes no structural
prior to promote scale invariance. 

\subsubsection{The FovConc network}

The scale generalisation ability of the FovConc network is quite similar to
that of the standard CNN, sometimes slightly worse. The reason why
the scale generalisation is limited is that although the scale
channels share their weights and thus produce a scale-covariant output, when simply concatenating these outputs from the scale channels,
there is no structural constraint to support scale invariance. This is
consistent with our observation that spanning a too wide scale 
range (Section \ref{sec-dep-scale-range}) or using too many channels, the scale generalisation degrades
for the FovConc network (Section \ref{sec-scale-sampl-density}). For scales {\em not present during training\/}, 
there is, simply, no useful training signal to learn the correct
weights in the fully connected layer that combines
the outputs from the different scale channels. 
Note that our results are not contradictory to those previously
reported for a similar network structure
\cite{XuXiaZhaYanZha14-arXiv}, since they train on data that contain
natural scale variations and test over a quite narrow scale
range. What we do show, however, is that this network structure, although it enables multi-scale processing, is {\em not scale invariant\/}.

\begin{figure*}[h]
	\centering
	\subfloat[Subfigure 1 list of figures text][The FovAvg network]{
		\includegraphics[width=0.48\textwidth]{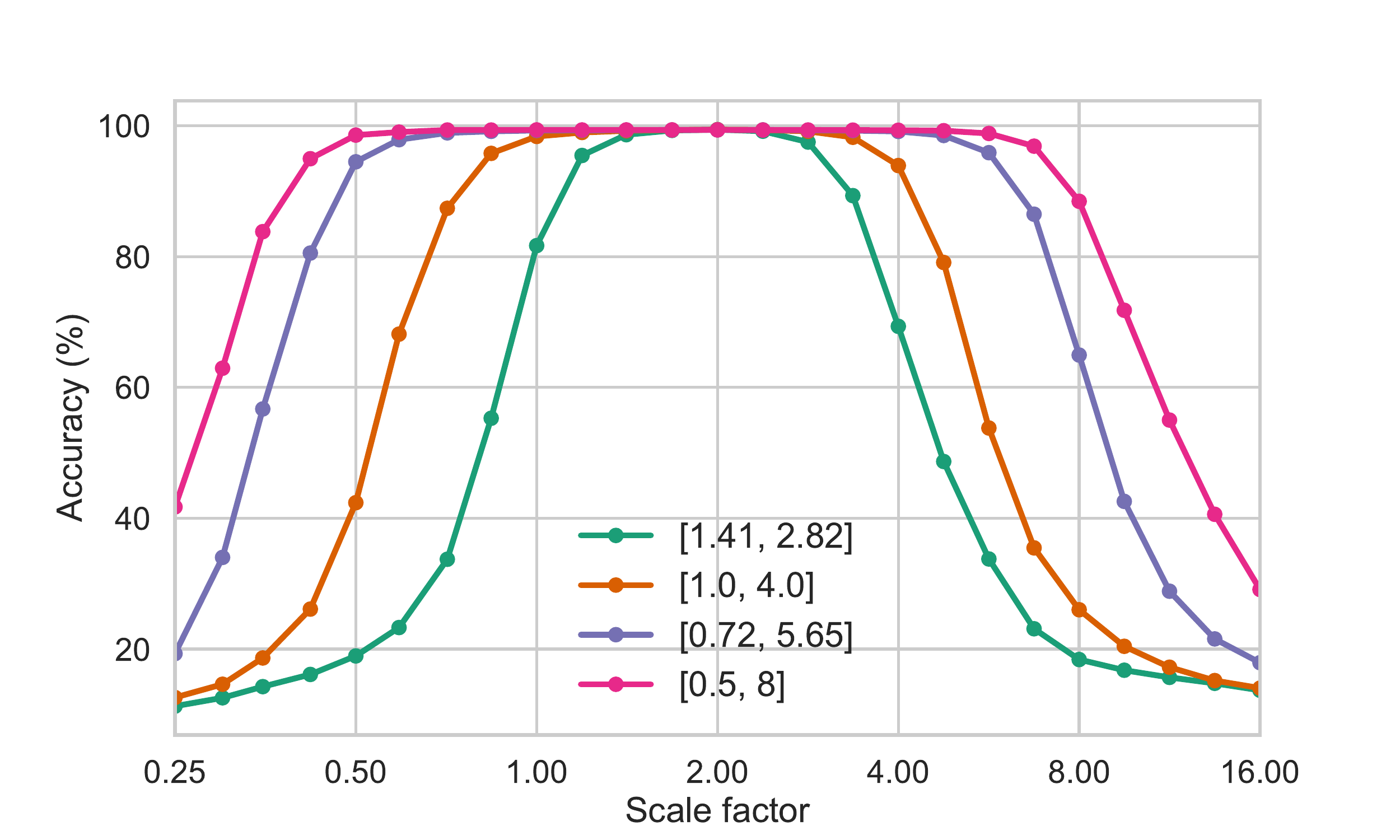}
	}
	\subfloat[Subfigure 2 list of figures text][The FovMax network]{
		\includegraphics[width=0.48\textwidth]{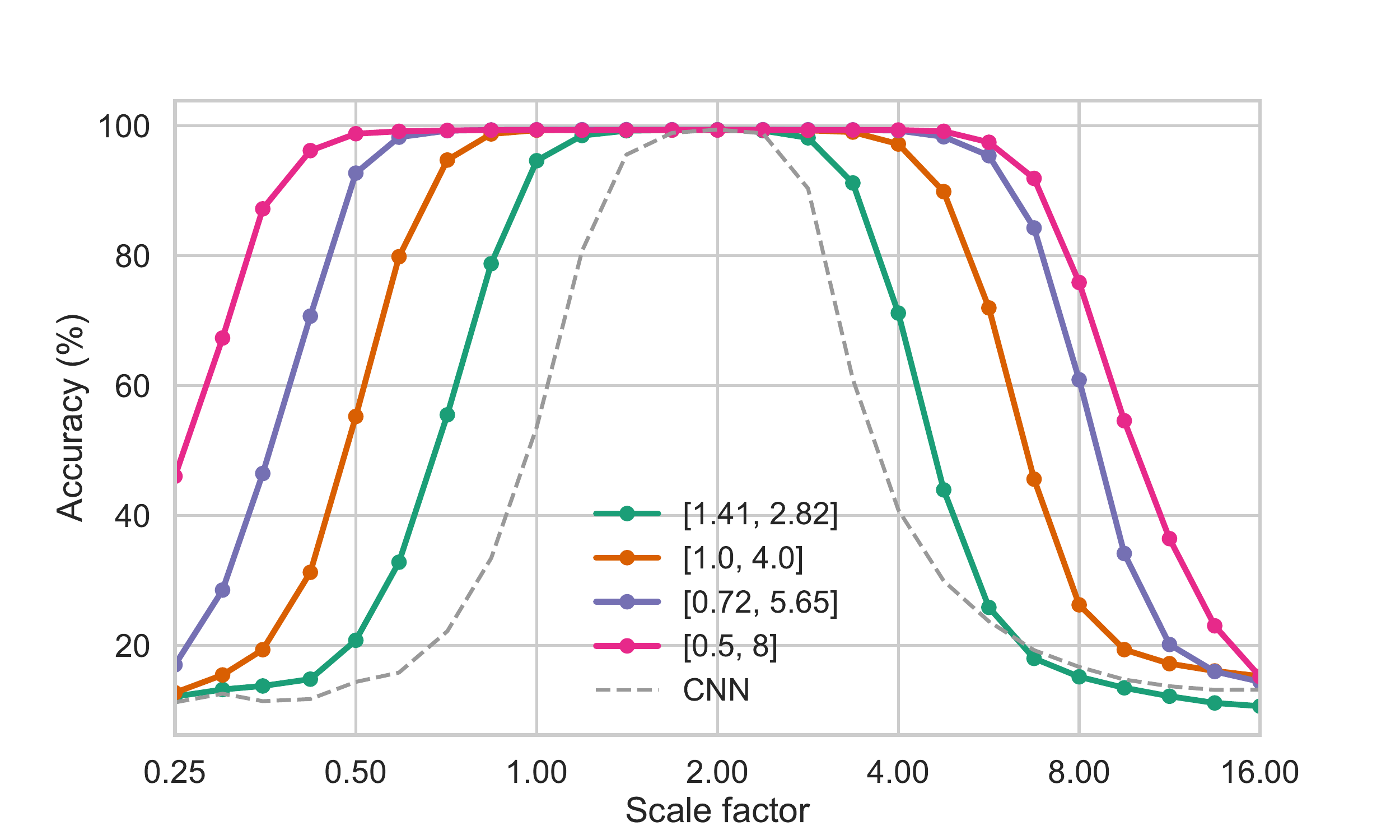}
	} \\
	\vspace*{-1em}
	\subfloat[Subfigure 3 list of figures text][The SWMax network]{
		\includegraphics[width=0.48\textwidth]{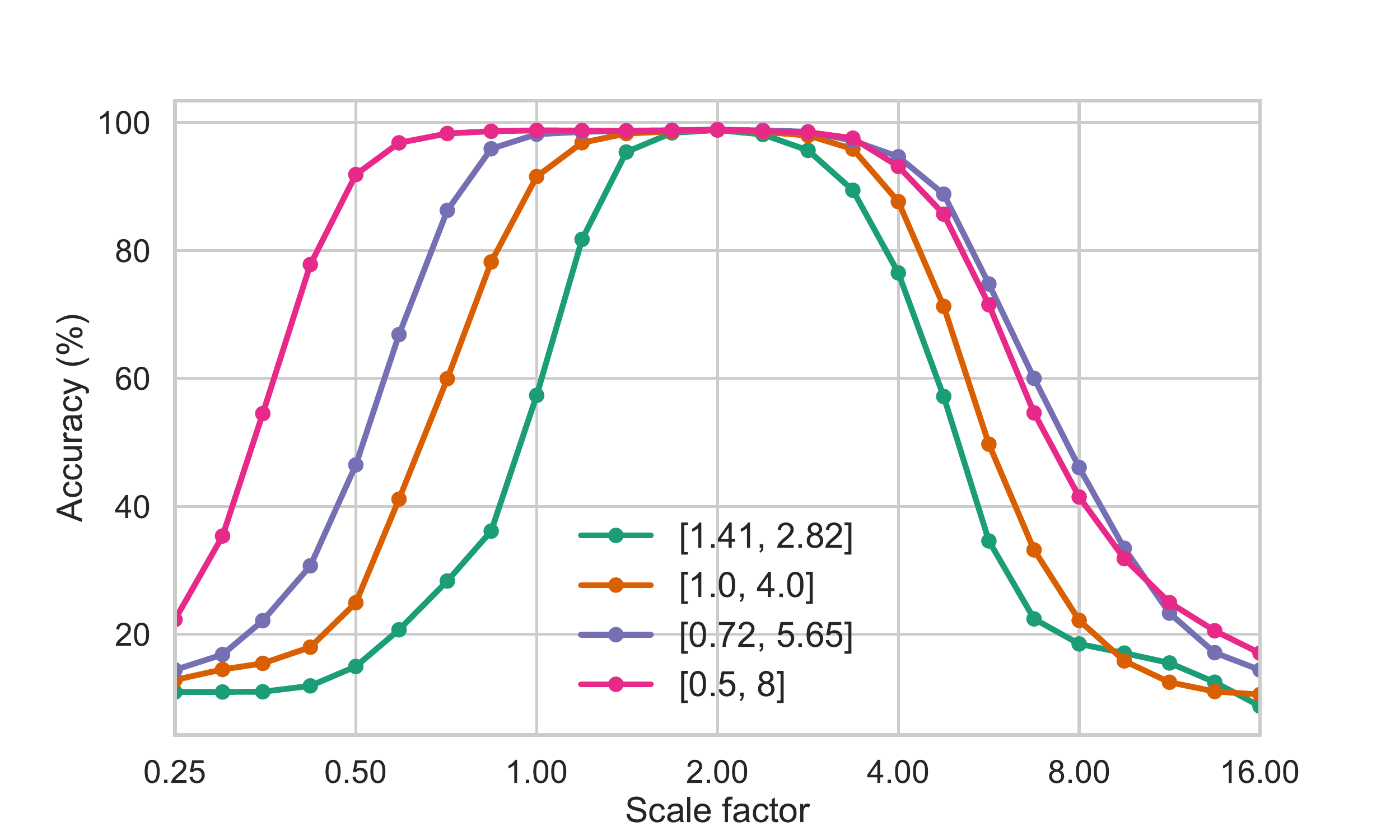}
	}
	\subfloat[Subfigure 4 list of figures text][The FovConc network]{
		\includegraphics[width=0.48\textwidth]{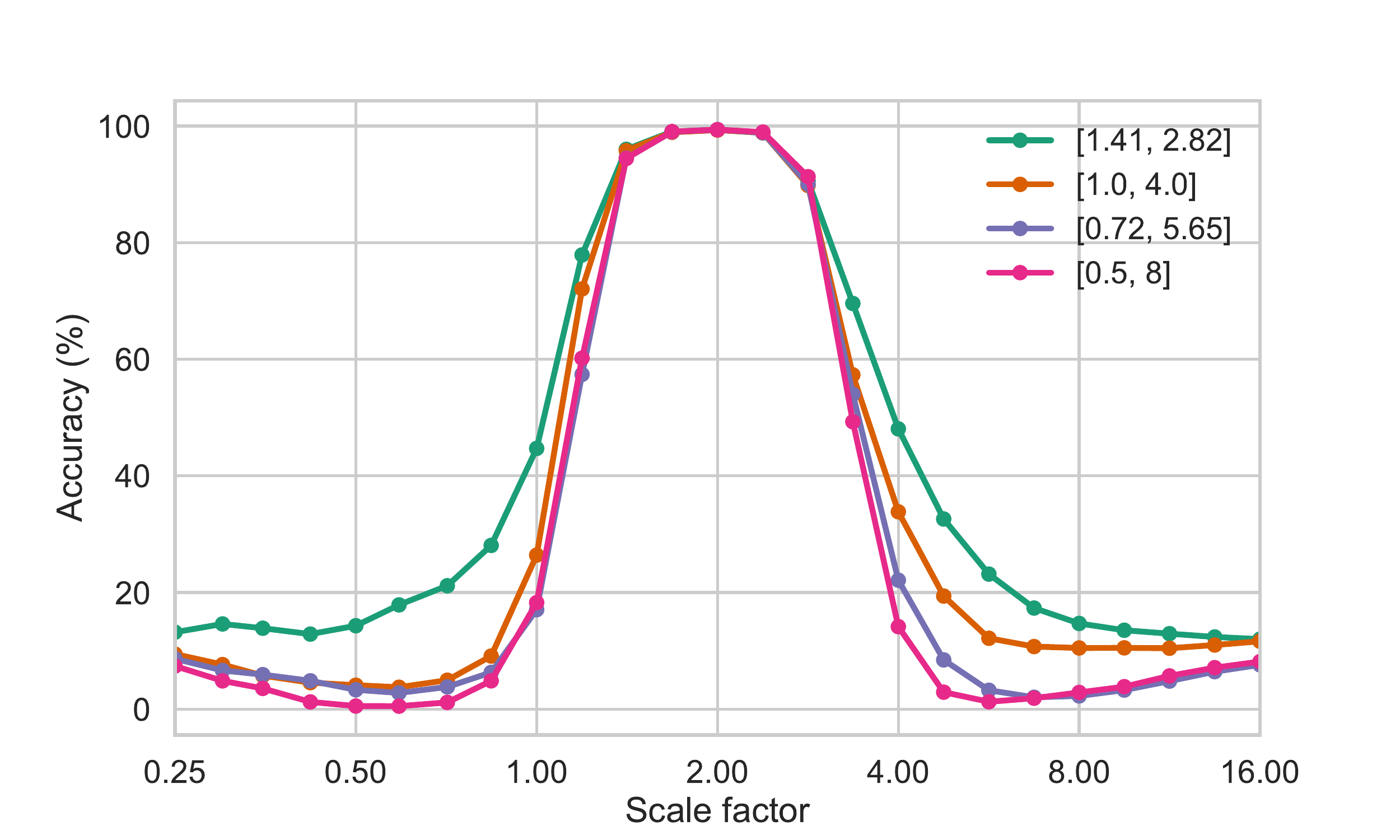}
	}
	\caption{{\em Dependency of the scale generalisation property
            on the scale range spanned by the scale channels:\/} 
		(a)--(b)~For the {\em FovAvg and FovMax\/} networks, the scale generalisation
                property is directly proportional to the scale range
                spanned by the scale channels, and there is no need to
                include training data for more than a single scale. 
		(c)~For the {\em SWMax\/} network, the scale generalisation is improved when including more scale channels, but the network does not generalise as well as the FovAvg and the FovMax networks. 
		(d)~For the {\em FovConc\/} network, the scale generalisation does
                actually become {\em worse\/} when including more
                scale channels (in the case of single-scale training), because there is no mechanism to
                support scale invariance when training the
                weights in the final fully connected layer that
                combines the different scale channels.
	}
	\label{fig-scale-range}
\end{figure*}

\begin{figure*}[h]
	\centering
	\subfloat[Subfigure 1 list of figures text][The FovAvg network]
	{
		\includegraphics[width=0.48\textwidth]{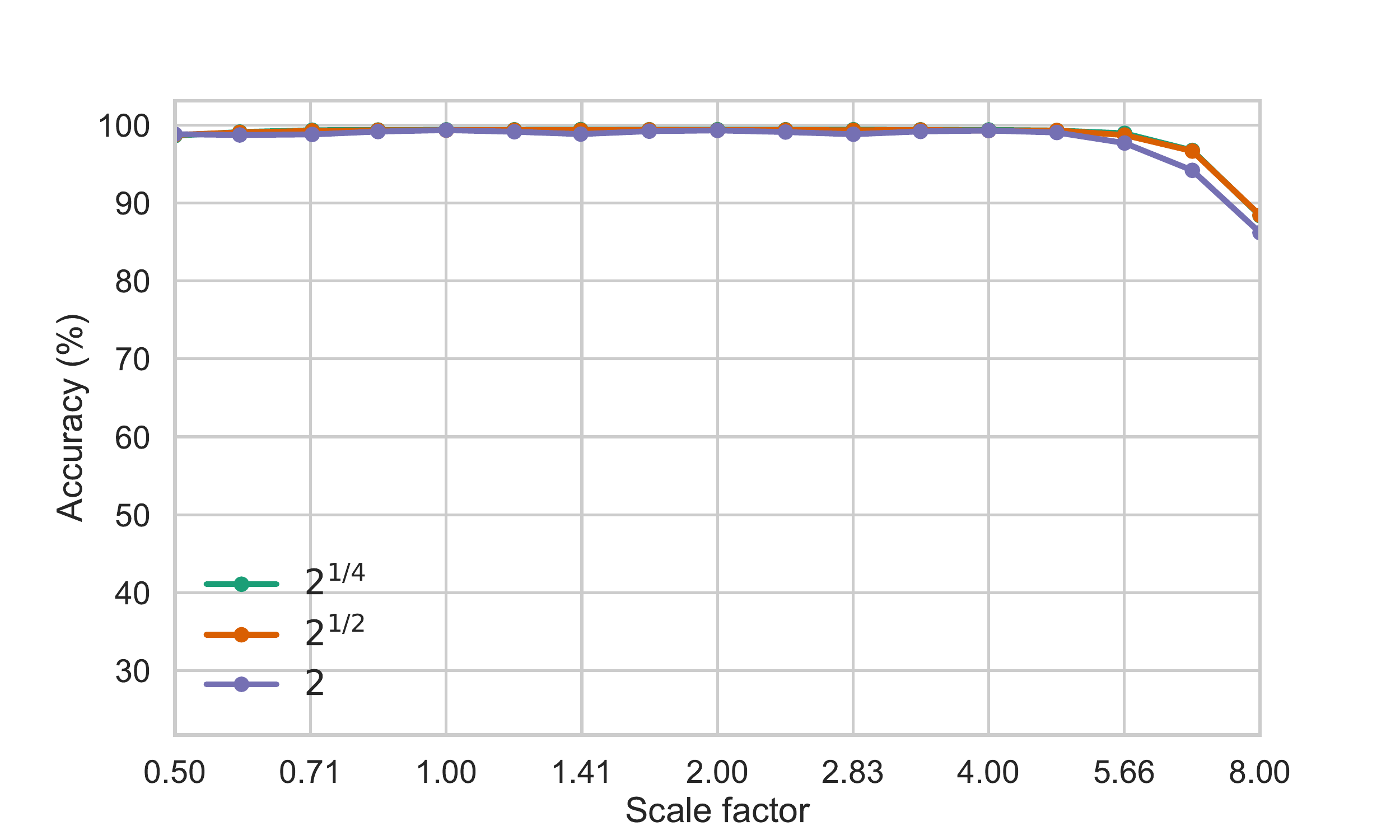}
	}
	\subfloat[Subfigure 2 list of figures text][The FovMax network]{
		\includegraphics[width=0.48\textwidth]{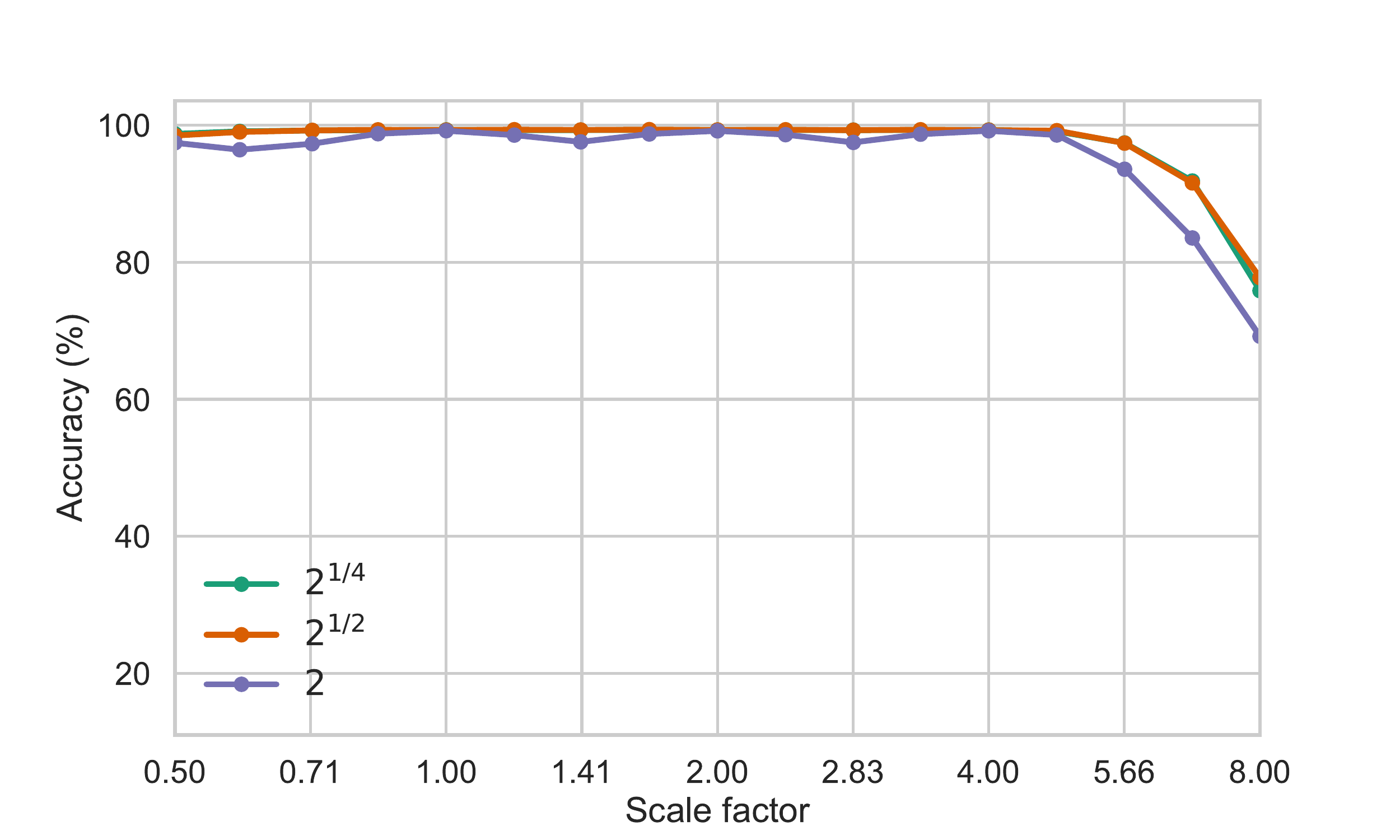}
	} \\
	\vspace*{-1em}
	\subfloat[Subfigure 3 list of figures text][The SWMax network]{
		\includegraphics[width=0.48\textwidth]{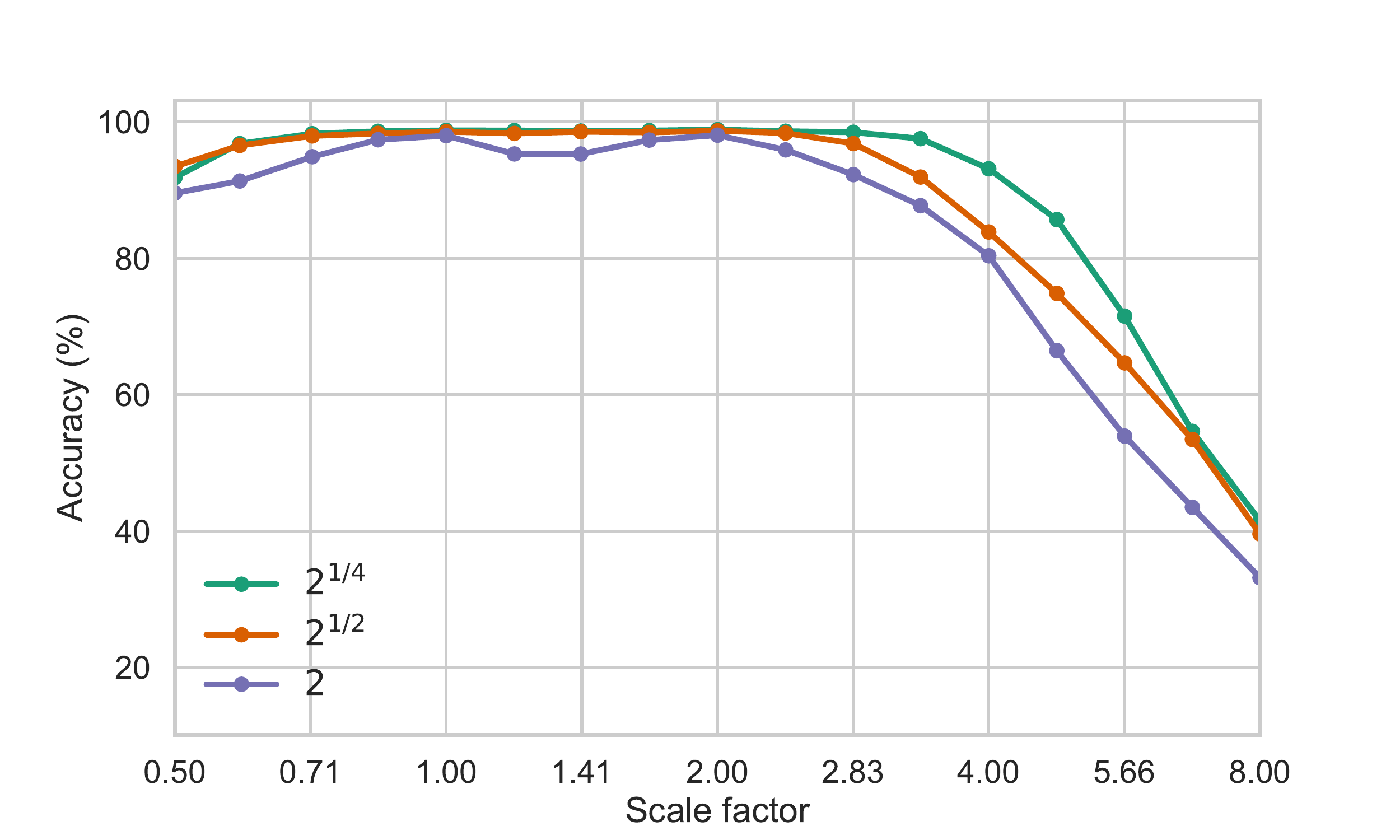}
	}
	\subfloat[Subfigure 4 list of figures text][The FovConc network]{
		\includegraphics[width=0.48\textwidth]{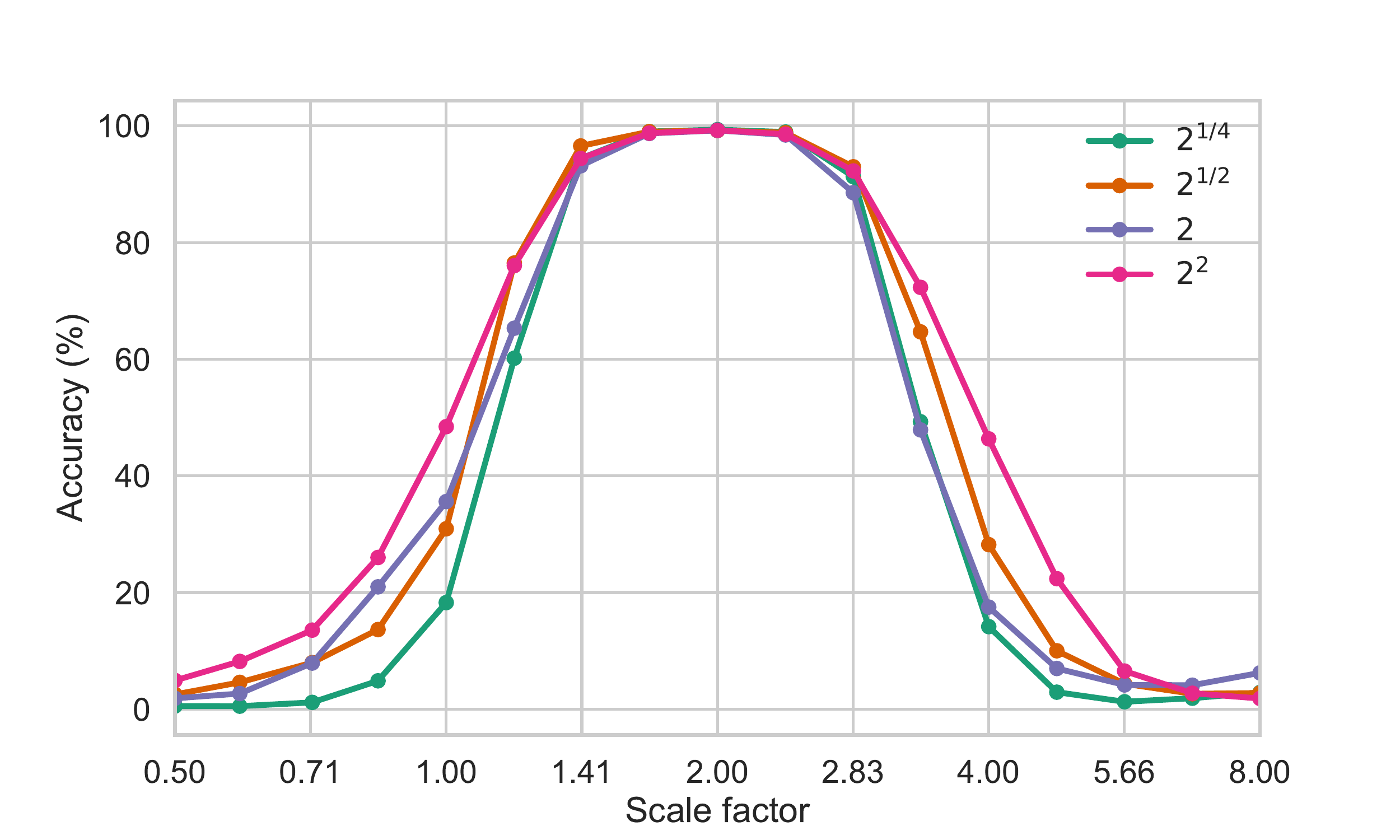}
	}
	\caption{{\em Dependency of the scale generalisation property
            on the scale sampling density:\/} 
		(a)--(b) For the FovAvg and FovMax networks, the
                overall scale generalisation is very good for all the
                studied scale sampling rates, although it becomes
                noticeably better for $2^{1/2}$ compared to $2$.
                For a more close up look regarding the {\em
                  FovAvg and FovMax networks\/}, 
                see Figure~\ref{fig-FovAvg-FovMax-scale-sampling-density}.
                (c)~The {\em SWMax\/} network is more sensitive to how densely the
                scales are samples compared to the FovAvg and the FovMax networks, and
                the sensitivity to the scale sampling density is larger when
                observing objects that are {\em larger\/} than those seen during
                training,  as compared to when observing objects that
                are {\em smaller\/} than those seen during training. 
                (d)~The {\em FovConc\/} network actually generalises worse with a denser
                sampling of scales.}
	\label{fig-scale-sampling-density}
\end{figure*}

\begin{figure*}[hbpt]
	
		\centering
	\subfloat[Subfigure 1 list of figures text][The FovAvg network]
	{
		\includegraphics[width=0.49\textwidth]{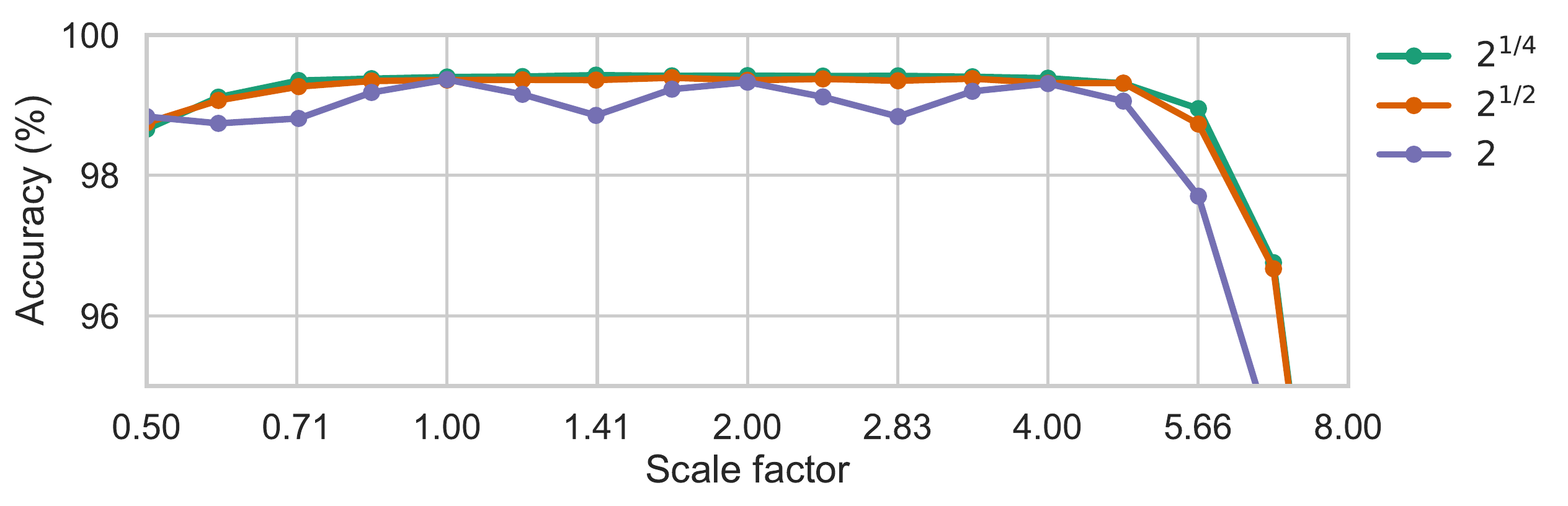} 
	}
	\subfloat[Subfigure 2 list of figures text][The FovMax network]{
			\includegraphics[width=0.49\textwidth]{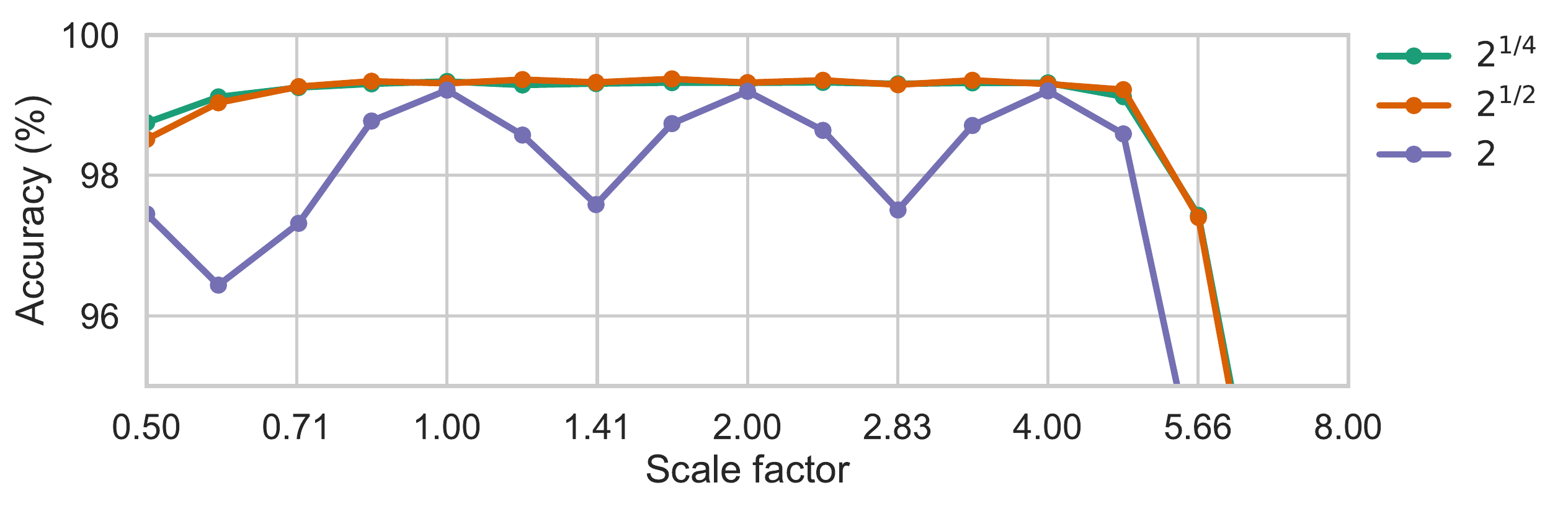}  
	} 
	\caption{{\em Dependency of the scale generalisation property
            on the scale sampling density for the FovAvg and FovMax
            networks:\/}
          FovMax and FovAvg networks spanning the scale
          range $[\frac{1}{4},8]$ were trained with varying spacing
          between the scale channels, either $2$, $2^{1/2}$ or
          $2^{1/4}$. All the networks were trained on size 2. There is a
          significant increase in the performance when reducing the
          spacing between the scale channels from $2$ to $2^{1/2}$,
          while the effect of a further reduction to $2^{1/4}$ is
          very small.}
	\label{fig-FovAvg-FovMax-scale-sampling-density}
      \end{figure*}

\subsubsection{The FovAvg and FovMax networks}

We note that the FovMax and FovAvg networks generalise very well,
independently of what size the network is trained on. The maximum
difference in performance in the size range $[1, 4]$ between training
on size~1, size~2  or size~4  is less than 0.2 percentage points for
these network architectures.
Importantly, this shows that, if including a large enough number of
sufficiently densely distributed scale channels and training the networks from scratch, boundary
effects at the scale boundaries do not prohibit invariant
recognition.

\subsubsection{The SWMax network}

We note that the SWMax network generalises considerably better than a
standard CNN, but there is some drop in performance for sizes not
seen during training. 
We believe that the main reason for this is, here, that since all the
scale channels are processing a fixed sized region in the input image
(as opposed to for foveated processing), new structures might leave or
enter this region when an image is rescaled. This might lead to
erroneous high responses for unfamiliar views
(see Section~\ref{sec-sliding-window}). 
We also noted that the SWMax networks are harder to train (more
sensitive to learning rate etc) compared to the foveated network
architectures as well as more computationally expensive. 
Thus, while the FovMax and Fov\-Avg networks still are easy to train and
the performance is not degraded when spanning a wide scale range, the
SWMax network seems to work best for spanning a more limited scale
range, where fewer scale channels are needed (as was indeed the use
case in \cite{SerEigZhaMatFerLeC13-arXiv}). 

\subsection{Dependency on the scale range spanned by the scale channels}
\label{sec-dep-scale-range}

Figure~\ref{fig-scale-range} shows the result of experiments to
investigate the sensitivity of the scale generalisation properties to
how wide range of scale values is spanned by the scale channels.
For all the experiments, we have used a scale sampling ratio of
$\sqrt{2}$ between adjacent scale channels.
All the networks were trained on the single size~2 and were
tested for all sizes between $\frac{1}{2}$ and 8.
The scale interval was varied between the four choices
$[\sqrt{2}, 2\sqrt{2}]$, $[1, 4]$, $[1/\sqrt{2}, 4\sqrt{2}]$ and
$[\frac{1}{2}, 8]$.

\subsubsection{The FovAvg and FovMax networks}

For the FovAvg and FovMax networks, the scale generalisation
properties are directly connected to how wide a scale 
range is spanned by the scale channels. By including more scale
channels, these networks generalise over a wider scale range, without
any need to include training data for more than a single scale.
The scale generalisation property will, however, be limited by the image
resolution for small testing sizes and by the fact that the full object
is not visible in the image for larger testing sizes.

\subsubsection{The SWMax network}

For the SWMax network, the scale generalisation property is improved
when including more scale channels, but the network does not
generalise as well as the FovAvg and the FovMax networks. It is also
noticeable that scale generalisation is harder when for large testing sizes compared to
small testing sizes. This is probably because of the problem with unfamiliar
partial views present for sliding window processing becoming more
pronounced for large testing sizes.

\subsubsection{The FovConc network}

For the FovConc network, the scale generalisation is actually
{\em worse\/} when including more scale channels. This phenomenon can
be understood by considering that the weights in the
fully connected layer, which combines information from the concatenated
scale channels output, are not controlled by any invariance
mechanism.
Indeed, the weights corresponding to scales not present during
training may take arbitrary values without any significant impact on
the training error.
Incorrect weights for unseen scales will, however, imply very poor generalisation to those scales.

\subsection{Dependency on the scale sampling density}
\label{sec-scale-sampl-density}

Figure~\ref{fig-scale-sampling-density} and
Figure~\ref{fig-FovAvg-FovMax-scale-sampling-density} show the result of 
experiments to investigate the sensitivity of the scale generalisation
property to the sampling density of the scale channels.
All the networks were trained on size~2, with the scale channels spanning
the scale range $[\frac{1}{2}, 8]$, and with a varying spacing between the
scale channels: either $2$, $2^{1/2}$ or $2^{1/4}$.
For the FovConc network, we also included the spacing $2^2$.

The number of scale channels for the different sampling densities were
for the $2^2$ spacing: 3~channels, for the $2$ spacing:  5~channels,
for the $2^{1/2}$ spacing: 9~channels and for the $2^{1/4}$ spacing:
17~channels.

\subsubsection{The FovAvg and FovMax networks}

For both the FovAvg and FovMax networks, the accuracy is
considerably improved when decreasing the ratio between adjacent
scale levels from a factor $2$ to a factor of $2^{1/2}$,
while a further reduction to $2^{1/4}$ 
provides very low additional benefits.%
\footnote{This result is consistent with results about scale sampling
  in classical scale-space theory, where it is known that uniform
  scale sampling in units of effective scale $\tau = \log \sigma$
  \cite{Lin92-PAMI} is the
  natural scale sampling strategy, and a scale sampling ratio of
  $\sqrt{2}$ often leads to substantially better performance than a scale
  sampling ratio of 2 in classical scale-space algorithms.}

\subsubsection{The SWMax network}

The SWMax network is more sensitive to how densely the scale levels are
sampled compared to the FovAvg and FovMax networks. This
sensitivity to the scale sampling density is larger, when observing
objects {\em of larger size\/} than those seen during training, as
compared to when observing objects {\em of smaller size\/} than
those seen during training.

This, again, illustrates the problem due to
partial views of objects, which will be present at some scales but not
at others, are more severe when observing larger size objects than
seen during training.

\subsubsection{The FovConc network}

The FovConc network does actually generalise worse with a denser sampling
of scales. In fact, none of the network versions generalises better
than a standard CNN. The reason for this is probably that for a dense
sampling of scales, there is no need for the last fully connected
layer, which processes the concatenated outputs from all the scale
channels, to include information from scales further away from the
training scale. Thus, the weights corresponding to such scales may
take arbitrary values without affecting the accuracy during the
training process, thereby implying very poor generalisation to
previously unseen scales.

\begin{figure}[hbpt]
	\begin{tabular}{c}
		\includegraphics[width=0.51\textwidth]{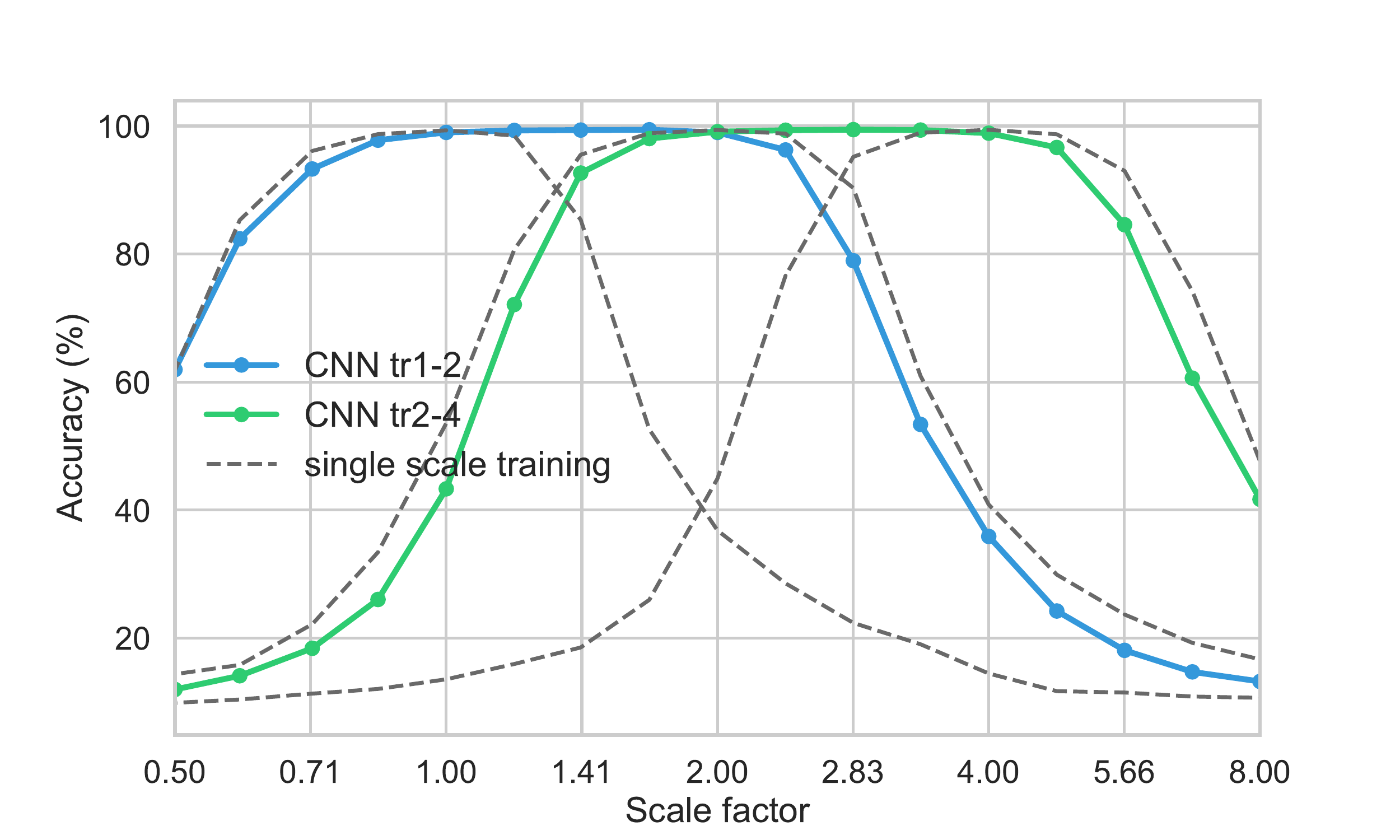}  
	\end{tabular}
	\caption{{\em Comparing multi-scale vs.\ single-scale training
            for a vanilla CNN\/}. Training is here performed over the
          size ranges $[1, 2]$ and $[2, 4]$, respectively. The scale generalisation when
          trained on single size training data is presented as dashed
          grey lines for training sizes~1, 2 and~4, respectively. As can be seen from the results,
          training on multi-scale training
          data does not improve the scale generalisation ability of the CNN for
          sizes {\em outside the size range the network is trained on\/}.}
        \label{fig-multi-scale-training-CNN}
\end{figure}

\begin{figure}[hbpt]
   \begin{tabular}{c}
	\includegraphics[width=0.50\textwidth]{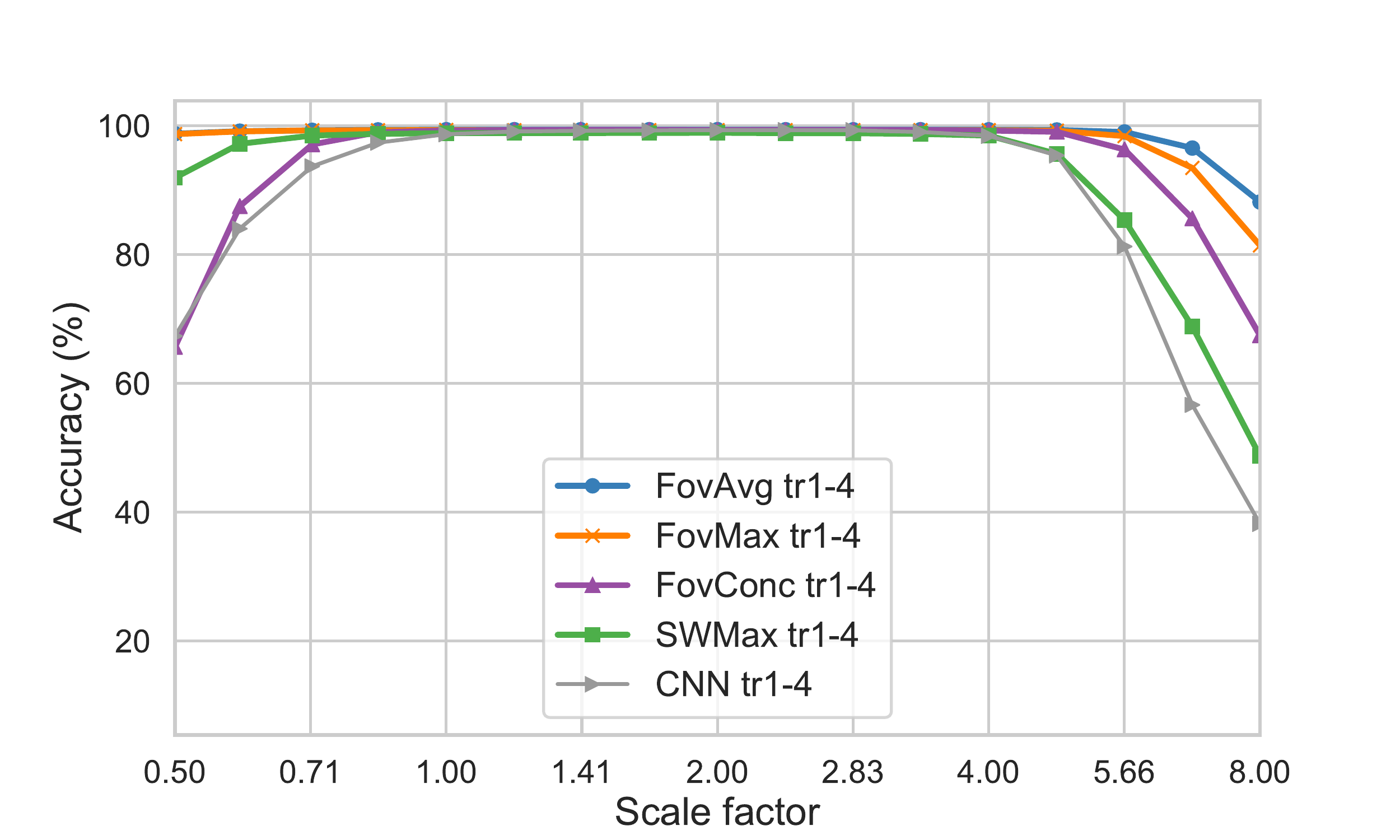}  \\
	\includegraphics[width=0.50\textwidth]{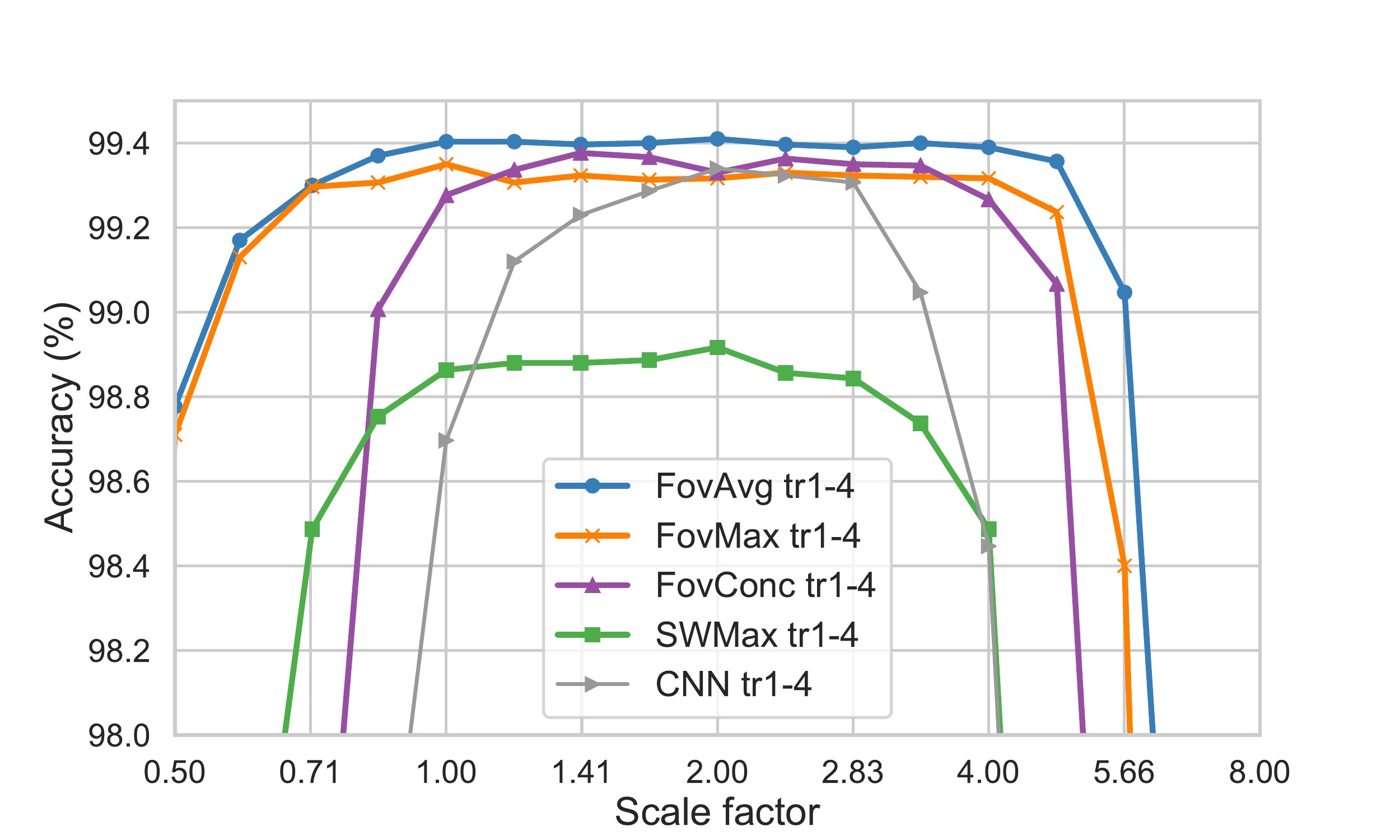}  
   \end{tabular}
   \caption{{\em Results of multi-scale training for the scale-channel
       networks with training sizes uniformly distributed on the size
       range $[1, 4]$ (with the uniform distribution on a logarithmic scale)\/.}
     These two figures show the same experimental results,
     where the second figure is zoomed in, to make comparisons between the networks more visible.
     The presence of multi-scale training data substantially improves the
     performance of the CNN, the FovConc network and the SWMax
     network. The difference in performance between
     single-scale training and multi-scale training is almost
     indiscernable for the FovAvg and FovMax networks.
     The overall best performance is 
     obtained for the FovAvg network.}
      \label{fig-multi-scale-training-scale-ch-networks}
\end{figure}
   
\subsection{Multi-scale {\em vs.\/}\ single-scale training}
\label{sec-multi-scale_vs_singlescale}

All the scale-channel architectures support multi-scale processing
although they might not support scale invariance. We, here, test the
performance of the different scale-channel networks when training on
multi-scale training data. For the standard CNN, we also explicitly explore how generalisation is affected when training on a smaller scale range to see how this affects generalisation outside the scale range trained on. 

\subsubsection{Limits of generalisation for a standard CNN}

If including data multi-scale data within a some range, could a CNN learn to ``extrapolate" outside this scale range? Figure~\ref{fig-multi-scale-training-CNN} shows the result of training
the standard CNN on training data with multiple sizes uniformly distributed over the scale ranges $[1, 2]$ and $[2, 4]$, respectively, and testing on all sizes
over the range $[\frac{1}{2}, 8]$. (The size distributions are uniform on a
logarithmic scale.)

Training on multi-scale training
data does not improve the scale generalisation ability of the CNN for
scales {\em outside the scale range the network is trained on}.
The network can, indeed, learn to recognise digits of different
sizes. But just because it might learn that an object of
size~1 is the same as the same object of size 2, this does
not at all imply that it will recognise the same object if it has
size~4. In other words, the scale generalisation ability within a
subrange does not transfer to outside that range.

\begin{table*}[tp]
\setlength{\tabcolsep}{6pt} 
\begin{center}
  {\em Compact performance measures regarding scale generalisation on
    the MNIST Large Scale dataset}

  \medskip

\begin{tabular}{l r r r r r r r}
\hline
\ Scale range & $[1/2,1]$ & $[1,4]$  & $[4,8]$ & $[1/2, 4]$ & $[1/2,8]$  \\ 
\hline
FovAvg 17ch tr1  &  99.15 & 99.27 & 90.82 & 99.22 & 96.76\\
FovAvg 17ch tr2  &  99.14 & 99.36 & 96.55 & 99.27 & 98.47\\
FovAvg 17ch tr4  &  98.78 & 99.31 & 96.61 & 99.11 & 98.36\\
FovAvg 17ch mean(tr1, tr2, tr4)  &  99.02 & 99.32 & 94.66 & 99.20 & 97.86\\
FovAvg 17ch tr14  &  99.20 & 99.40 & 96.50 & 99.32 & 98.49\\
\hline
FovMax 17ch tr1  &  99.15 & 99.35 & 93.70 & 99.27 & 97.63\\
FovMax 17ch tr2  &  99.15 & 99.31 & 92.72 & 99.25 & 97.32\\
FovMax 17ch tr4  &  99.03 & 99.30 & 93.26 & 99.20 & 97.45\\
FovMax 17ch mean(tr1, tr2, tr4)  &  99.11 & 99.32 & 93.23 & 99.24 & 97.47\\
FovMax 17ch tr14  &  99.16 & 99.32 & 94.37 & 99.26 & 97.82\\
\hline
FovConc 3ch tr1  &  80.76 & 48.64 & 4.61 & 57.10 & 44.68\\
FovConc 3ch tr2  &  22.35 & 78.17 & 22.71 & 59.12 & 49.55\\
FovConc 3ch tr4  &  2.57 & 50.20 & 82.36 & 35.64 & 45.63\\
FovConc 3ch mean(tr1, tr2, tr4)  &  35.23 & 59.00 & 36.56 & 50.62 & 46.62\\
FovConc 17ch tr14  &  89.70 & 99.33 & 89.54 & 95.63 & 93.63\\
\hline
SWMax 17ch tr1  &  95.06 & 97.60 & 69.52 & 96.53 & 88.77\\
SWMax 17ch tr2  &  96.87 & 97.96 & 69.28 & 97.48 & 89.44\\
SWMax 17ch tr4  &  91.40 & 97.23 & 82.21 & 95.02 & 91.04\\
SWMax 17ch mean(tr1, tr2, tr4)  &  94.44 & 97.60 & 73.67 & 96.34 & 89.75\\
SWMax 17ch tr14  &  97.05 & 98.82 & 79.40 & 98.13 & 92.60\\
\hline
CNN tr1  &  88.26 & 50.78 & 11.85 & 61.46 & 49.64\\
CNN tr2  &  27.87 & 79.88 & 26.08 & 61.90 & 52.60\\
CNN tr4  &  11.45 & 54.35 & 82.59 & 40.99 & 49.79\\
CNN mean(tr1, tr2, tr4)  &  42.53 & 61.67 & 40.17 & 54.78 & 50.68\\
CNN tr14  &  88.23 & 99.09 & 73.98 & 94.94 & 88.57\\
  \hline
\end{tabular}
\end{center}

\caption{{\em Average classification accuracy (\%) over different
    size ranges of the testing data.\/}
For each type of network (FovAvg, FovMax, FovConc, SWMax or CNN), this
table shows the average classification accuracy over different ranges
of the size of the testing data in the MNIST Large Scale datasets, for
networks trained by single-scale training for either of the training sizes
1, 2 or 4 (denoted tr1, tr2, tr4) or multi-scale training data spanning the
scale range $[1, 4]$ (denoted tr14).  The rows labelled ``mean(tr1,
tr2, tr4)'' give the average value for the training sizes 1, 2 and 4. The reported accuracy is the average of the accuracy for multiple test sizes within the size ranges
$[1/2,1], [1,4], [4,8]$, $[1/2,4]$ and $[1/2,8]$ with spacing $2^{1/4}$ between consecutive sizes.} 
\label{tab-benchmarks}
\end{table*}

\begin{table}[hbtp]
  \begin{tabular}{lr}
  \hline
  \multicolumn{2}{c}{\em Single-scale training evaluated over testing sizes in $[1, 4]$} \\
    \hline
    FovAvg mean(tr1, tr2, tr4) & 99.32 \% \\
    FovMax mean(tr1, tr2, tr4) & 99.32 \% \\
    SWMax mean(tr1, tr2, tr4) & 97.60 \% \\
    CNN mean(tr1, tr2, tr4)  & 61.67 \% \\
    FovConc mean(tr1, tr2, tr4) & 59.00 \% \\
    \hline
    \end{tabular}
 \caption{Relative ranking of the different networks for single-scale
   training at either of the training sizes 1, 2 or~4 evaluated over the testing size interval $[1, 4]$.}
 \label{tab-benchmark-single-scale-training-1-4}

 \bigskip
  \bigskip
 
  \begin{tabular}{lr}
  \hline
  \multicolumn{2}{c}{\em Multi-scale training evaluated over testing sizes in $[1, 4]$} \\
    \hline
    FovAvg tr14 & 99.40 \% \\
    FovConc tr14 & 99.33 \% \\
    FovMax tr14 & 99.32 \% \\
    CNN tr14 & 99.09 \% \\
    SWMax tr14 & 98.82 \% \\
    \hline
    \end{tabular}
 \caption{Relative ranking of the different networks for multi-scale
   training over the training size interval $[1, 4]$ evaluated over the testing size interval $[1, 4]$.}
  \label{tab-benchmark-multi-scale-training-1-4}
\end{table}

\begin{table}[hbtp]
  \begin{tabular}{lr}
  \hline
  \multicolumn{2}{c}{\em Single-scale training evaluated over testing sizes in $[1/2, 4]$} \\
    \hline
    FovMax mean(tr1, tr2, tr4) & 99.24 \% \\
    FovAvg mean(tr1, tr2, tr4) & 99.20 \% \\
    SWMax mean(tr1, tr2, tr4) & 96.34 \% \\
    CNN mean(tr1, tr2, tr4)  & 54.78 \% \\
    FovConc mean(tr1, tr2, tr4) & 50.62 \% \\
    \hline
    \end{tabular}
 \caption{Relative ranking of the different networks for single-scale
   training at either of the training sizes 1, 2 or~4 evaluated over the testing size interval $[1/2, 4]$.}
 \label{tab-benchmark-single-scale-training-1/2-4}

 \bigskip
 \bigskip
 
  \begin{tabular}{lr}
  \hline
  \multicolumn{2}{c}{\em Multi-scale training evaluated over testing sizes in $[1/2, 4]$} \\
    \hline
    FovAvg tr14 & 99.32 \% \\
    FovMax tr14 & 99.26 \% \\
    SWMax tr14 & 98.13 \% \\
    FovConc tr14 & 95.63 \% \\
    FovConc tr14 & 96.32 \% \\
    CNN tr14 & 94.94 \% \\
    \hline
    \end{tabular}
 \caption{Relative ranking of the different networks for multi-scale
   training over the training size interval $[1, 4]$ evaluated over the testing size interval $[1/2, 4]$.}
  \label{tab-benchmark-multi-scale-training-1/2-4}
\end{table}

\subsubsection{Multi-scale training}

Figure~\ref{fig-multi-scale-training-scale-ch-networks} shows the
result of performing multi-scale training over the size range $[1, 4]$
for the scale-channel networks FovMax, FovAvg, FovConc and SWMax as well as the standard CNN.
Here, the same scale-channel setup with 17~channels
spanning the scale range $[\frac{1}{2}, 8]$ is used for all the scale-channel
architectures. When multi-scale training data is used, the advantage of using scale channels spanning a larger scale range no longer incurs a penalty for the FovConc network, since the correct weights can be learned in the fully connected layer. 

We note that the difference between training on multi-scale and single-scale data
is striking both for the FovConc network and the standard CNN. It can,
however, be noted that the FovConc network works well in this
scenario, especially for the scale range included in the training
set. Outside this scale range, we note somewhat better generalisation
compared to the CNN, while the generalisation is still worse than for the FovAvg and FovMax networks. The FovConc network does, after all, include a mechanism for
multi-scale processing and when trained on multi-scale training data,
the lack of invariance mechanism in the fully connected layer is less
of a problem.

For the SWMax network, including multi-scale data improves
the scale generalisation somewhat compared to single-scale training.
The SWMax network does, however, have worse performance for spanning larger scale ranges compared
to the other networks. The reason behind this is probably that the
multiple views produced in the different scale channels
indeed makes the problem harder for this network compared to the
foveated networks, which only need to process centered digit views.
 
The difference in scale generalisation ability between training on a single
scale or multi-scale image data is on the other hand almost indiscernible for the FovMax
and FovAvg networks (less than 0.1 \% difference in accuracy), illustrating the strong scale invariance properties of these networks.

\subsection{Compact benchmarks regarding the scale generalisation performance}

Table~\ref{tab-benchmarks} gives compact performance
measures of the generalisation performance of the different types of
networks considered in the experiments on the MNIST Large Scale
dataset.
For each type of network (FovAvg, FovMax, FovConc, SW or CNN), the
table gives the average classification accuracy over different ranges
of the size of the testing data, for networks trained by single-scale
training, for either of the training sizes
1, 2 or 4 or multi-scale training data spanning the
scale range $[1, 4]$.

Tables~\ref{tab-benchmark-single-scale-training-1-4}--\ref{tab-benchmark-multi-scale-training-1/2-4}
gives relative
  ranking of the different networks on specific subsets of this data,
  which can be treated as benchmarks regarding scale generalisation
  for the MNIST Large Scale dataset.
  As can be seen from these tables, the FovAvg and FovMax networks
  have the overall best performance scores of these networks, both for
  the cases of single-scale training and multi-scale training.
  
  The FovConc, CNN and SWMax networks are very much improved by
  multi-scale training, whereas the FovAvg and FovMax networks perform
  almost as well for single-scale training as for
  multi-scale training.

\begin{figure}[hbpt]
	\begin{center}
		\includegraphics[width=0.48\textwidth]{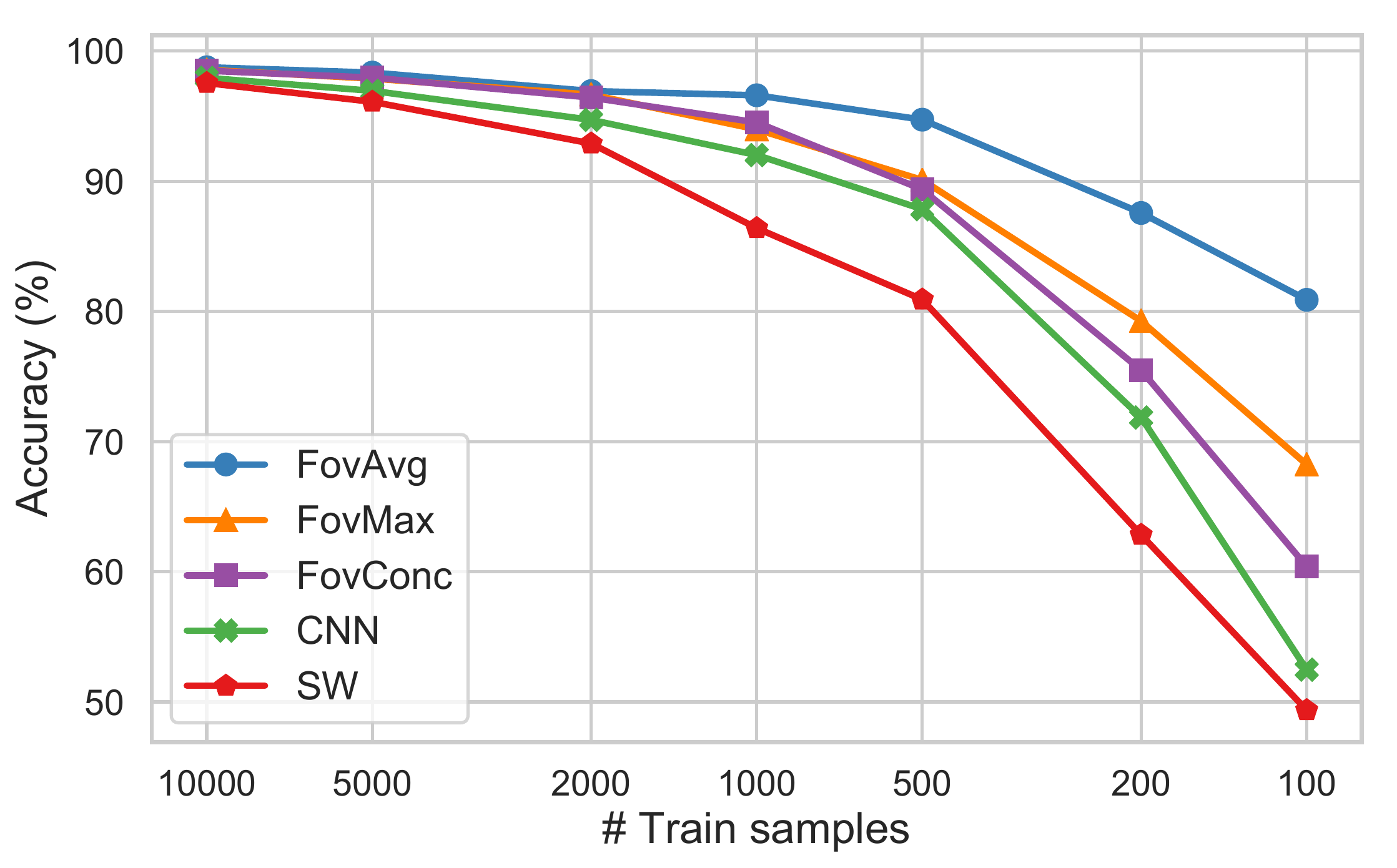}		
	\end{center}
	\caption{{\em Training  with smaller training sets with large
            scale variations\/}. All the network architectures are evaluated
          on their ability to classify data with large scale
          variations, while reducing the number of training
          samples. Both the training and the testing sets here span the size
          range $[1,4]$. The Fov\-Avg network shows the highest
          robustness when decreasing the number of training samples
          followed by the FovMax network. 
         The FovConc network also shows a small improvement over the
         standard CNN.}
	\label{fig-mnist_few_samples}
\end{figure}

\subsection{Generalisation from fewer training samples}

Another scenario of interest is when the training data does span a
relevant range of scales, but there are few training samples. Theory
would predict a correlation between the performance in this scenario
and the ability to generalise to unseen scales. 

To test this prediction, we trained the standard CNN and the different
scale-channel networks on multi-scale training data spanning the size
range $[1,4]$, while gradually reducing the number of samples in the
training set. Here, the same scale-channel setup with 17~channels
spanning the scale range $[\frac{1}{2}, 8]$ was used for all the
architectures. The results are presented in
Figure~\ref{fig-mnist_few_samples}.
We can note that the FovConc network
shows some improvement over the standard CNN. The SWMax network, on
the other hand, does not, and we hypothesise that when using fewer
samples, the problem with partial views of objects (see
Section~\ref{sec-sliding-window}) might be more severe.
Note that the way the OverFeat detector is used in the original study
\cite{SerEigZhaMatFerLeC13-arXiv} is more similar to our single-scale
training scenario, since they use base networks pre-trained on
ImageNet.  The FovAvg and FovMax networks show the highest robustness
also in this scenario. This illustrates that these networks can give
improvements when multi-scale training data is available, but there are
few training samples. 

\begin{figure*}[hbtp]
\begin{center}
   \begin{tabular}{cc}
     {\em Selected scale channels: FovAvg network trained for size 1}
     & {\em Selected scale channels: FovMax network trained for size 1} \\
     \includegraphics[width=0.27\textheight]{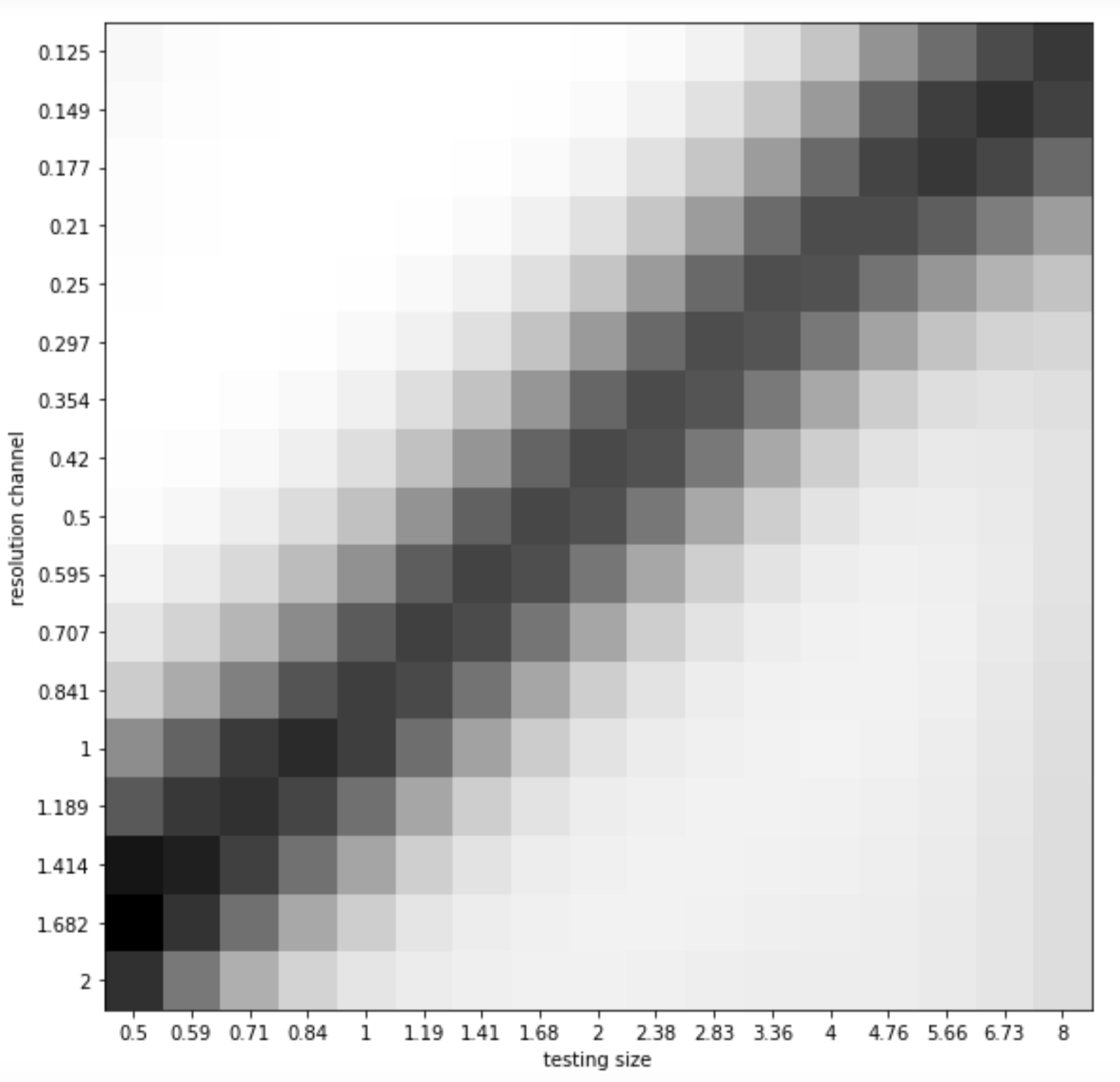}
     & \includegraphics[width=0.27\textheight]{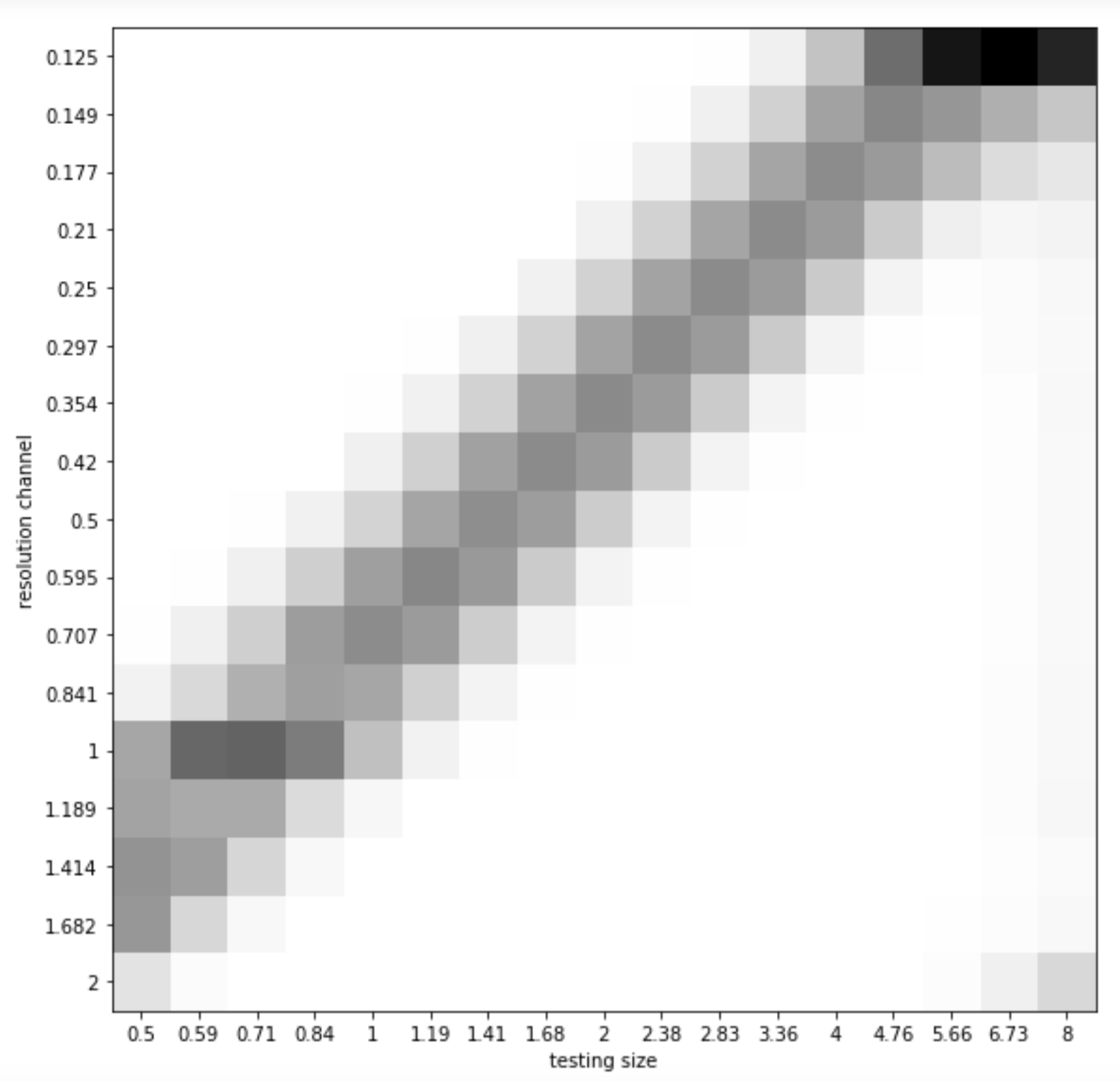}\\
     \\
     {\em Selected scale channels: FovAvg network trained for size 2}
     & {\em Selected scale channels: FovMax network trained for size 2} \\
     \includegraphics[width=0.27\textheight]{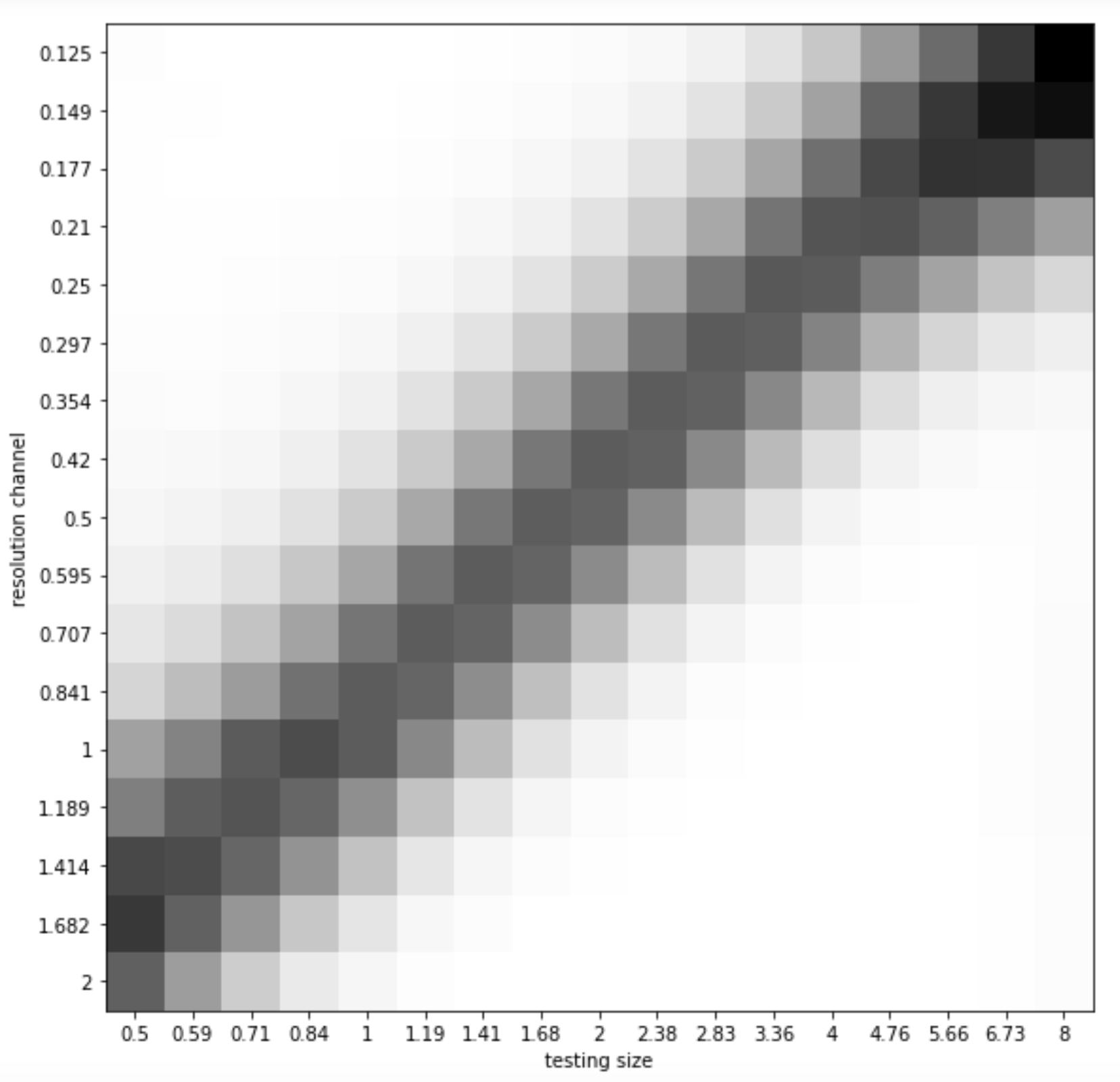}
     & \includegraphics[width=0.27\textheight]{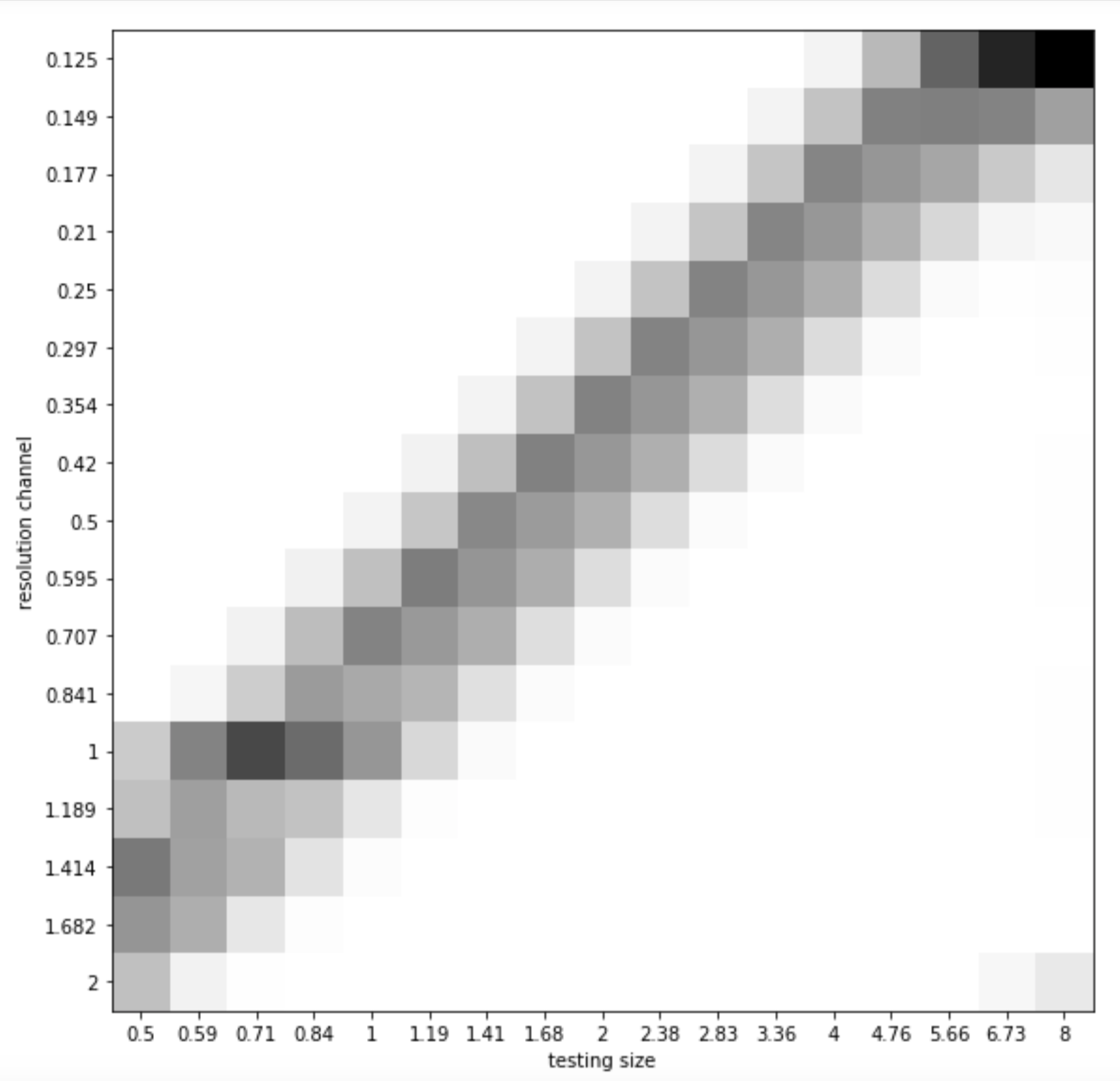}\\
      \\
     {\em Selected scale channels: FovAvg network trained for size 4}
     & {\em Selected scale channels: FovMax network trained for size 4} \\
     \includegraphics[width=0.27\textheight]{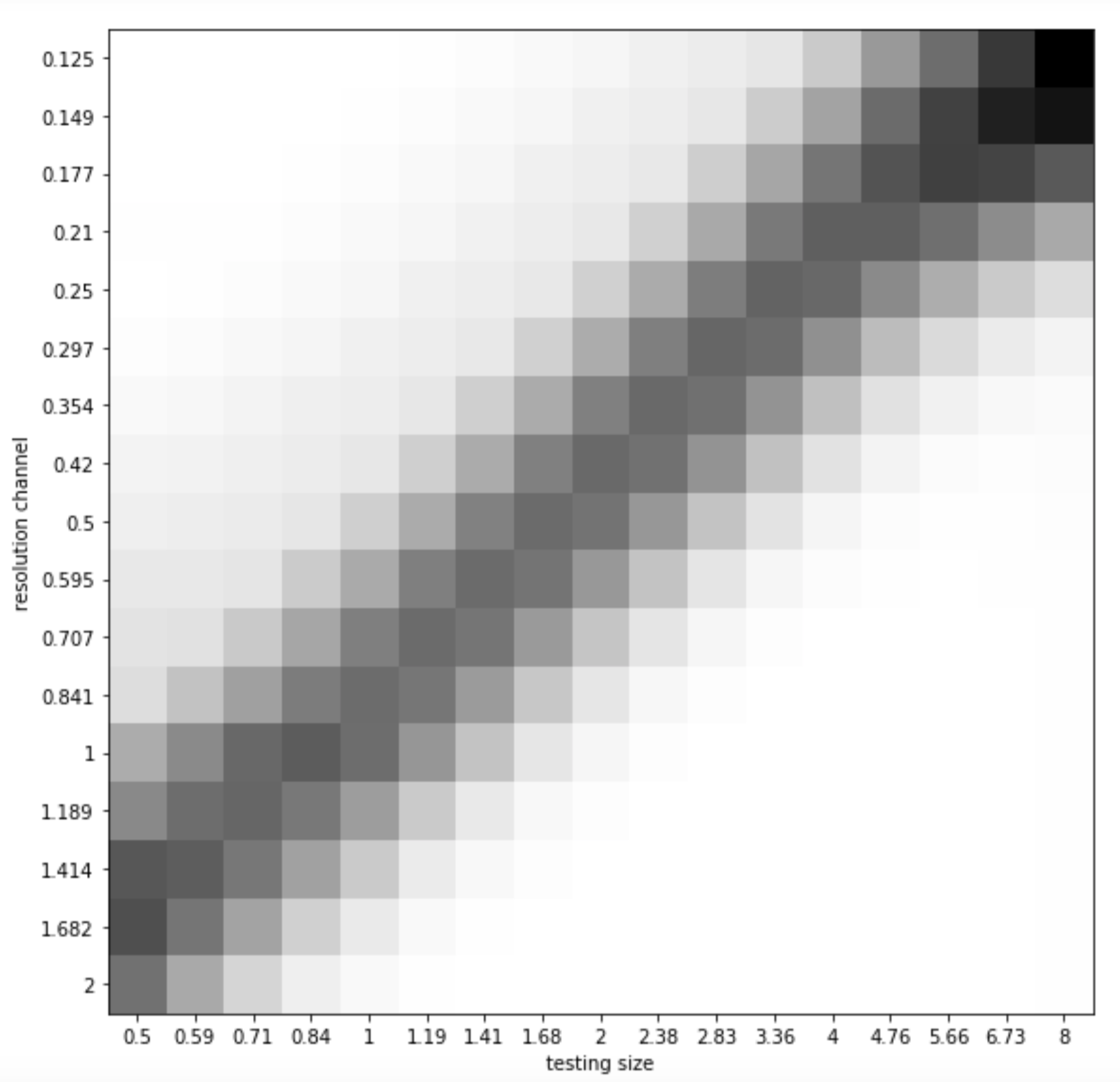}
     & \includegraphics[width=0.27\textheight]{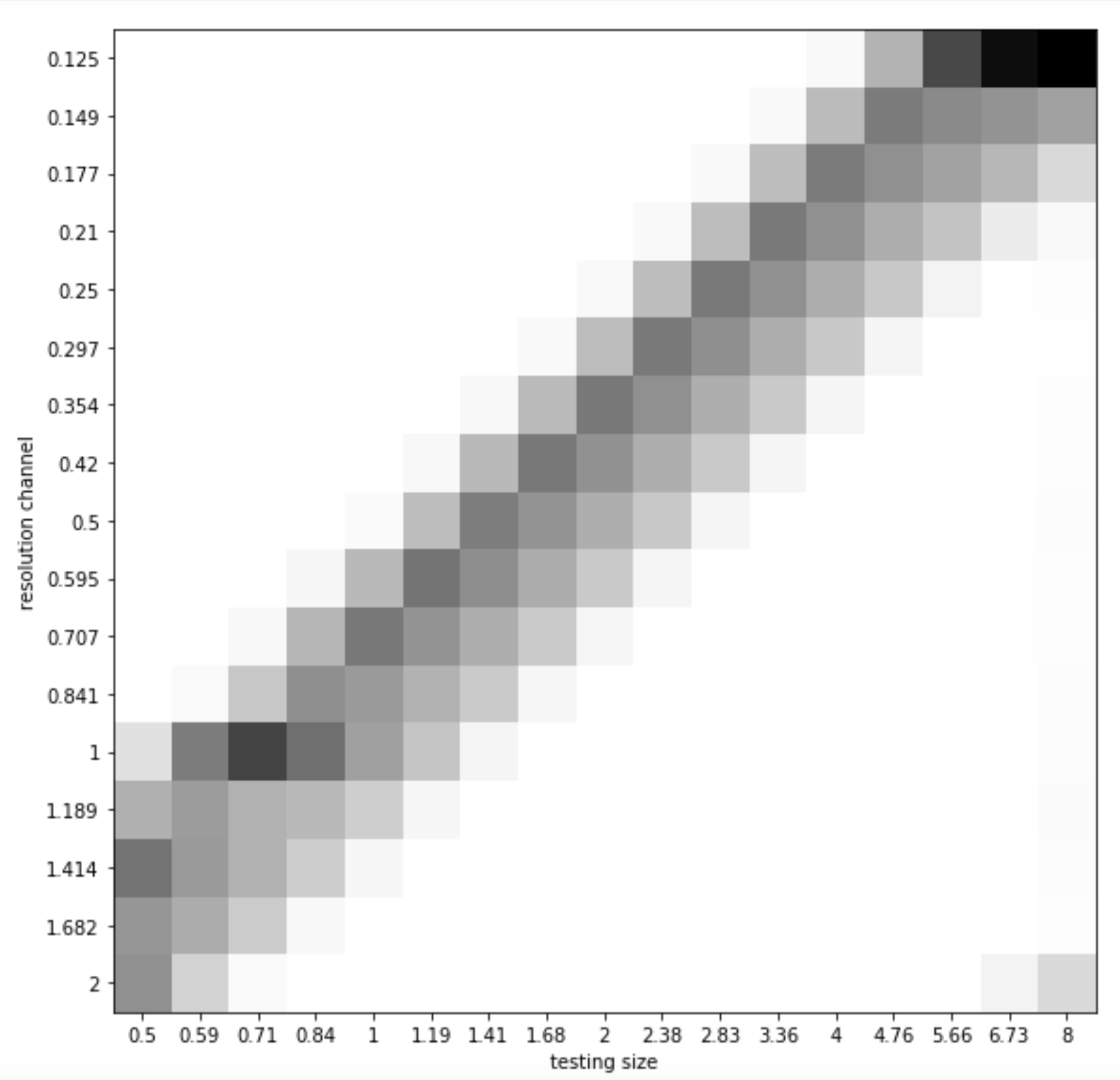}\\
    \end{tabular}
  \end{center}
  \caption{\emph{Visualisation of the scale selection properties of the  scale-invariant FovAvg and FovMax
    networks, when training the network for each one of the sizes 1, 2 and 4}.
For each testing size, shown on the
    horizontal axis with increasing testing sizes towards the right, the vertical axis displays a histogram of the relative contribution of the scale channels to the winning classification, with the
    lowest scale at the bottom and the highest scale at the top. As can be
    seen from the figures, there is a general tendency of the
    composed classification scheme to select coarser scale levels with
  increasing size of the image structures, in agreement with the
  conceptual similarity to classical methods for scale selection based
on detecting local extrema over scale or performing weighted averaging over
scale of scale-normalised derivative responses. (In these figures, the
resolution parameter on the vertical axis represents the inverse of
scale. Note that the grey-levels in the histograms are not directly
comparable, since the grey-levels for each histogram are normalised
with respect to the maximum and minimum values in that histogram.)}
   \label{fig-sc-sel-FovAvg-FovMax}
 \end{figure*}

\begin{figure*}[hbtp]
\begin{center}
   \begin{tabular}{cc}
     {\em Selected scale channels: FovConc network trained for size 1}
     & {\em Selected scale channels: SWMax network trained for size 1} \\
     \includegraphics[width=0.27\textheight]{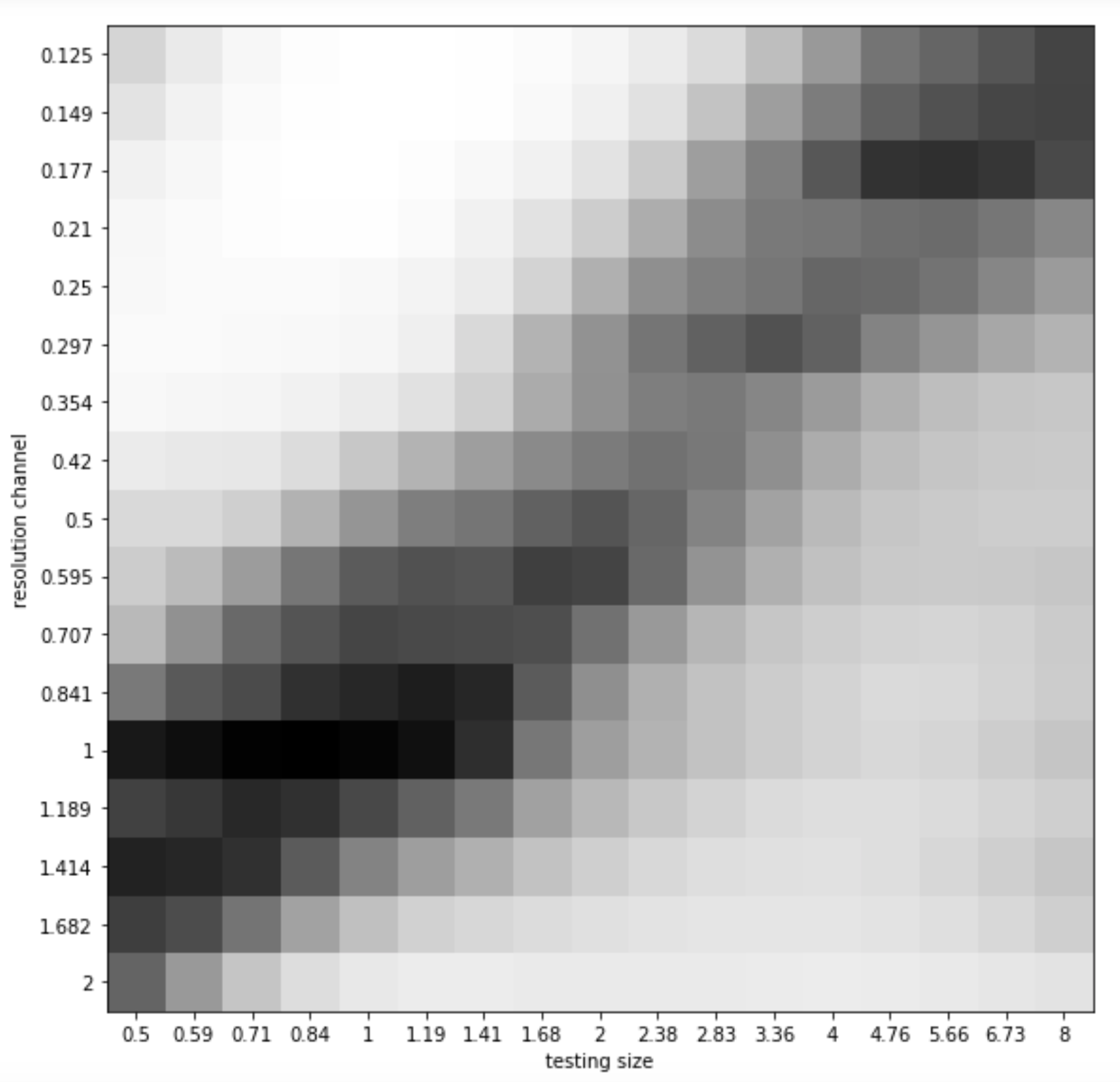}
     & \includegraphics[width=0.27\textheight]{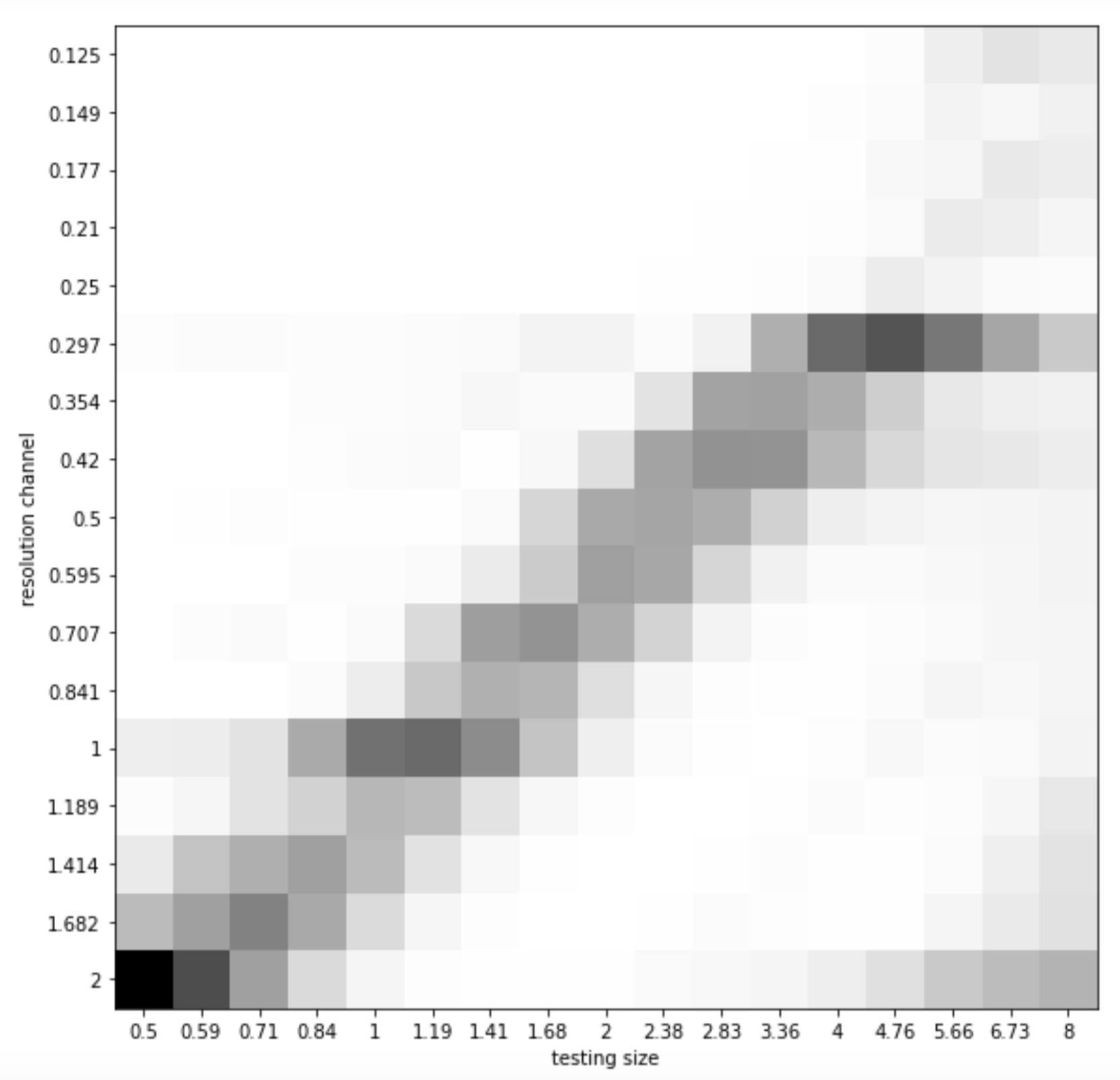}\\
     \\
     {\em Selected scale channels: FovConc network trained for size 2}
     & {\em Selected scale channels: SWMax network trained for size 2} \\
     \includegraphics[width=0.27\textheight]{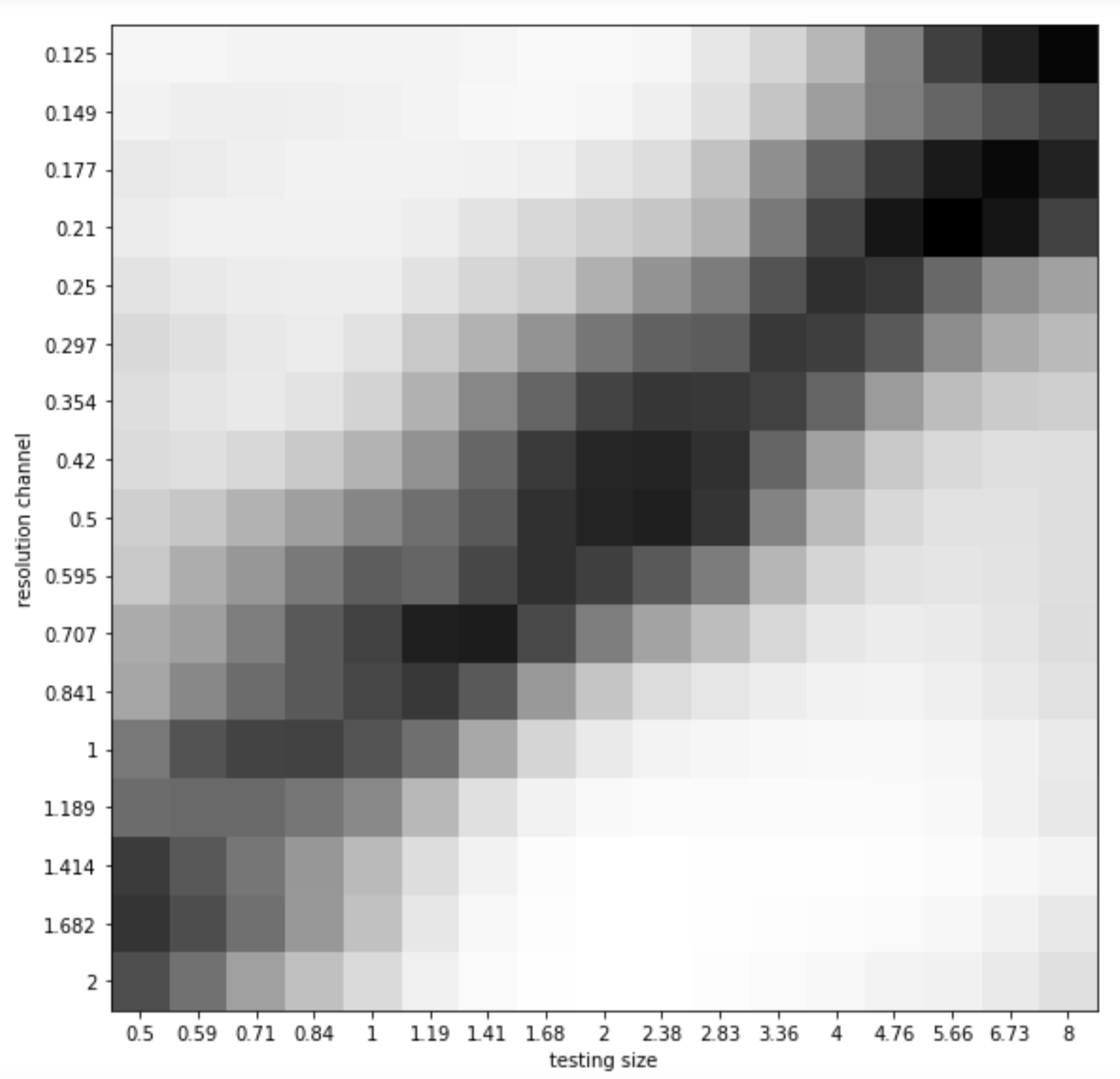}
     & \includegraphics[width=0.27\textheight]{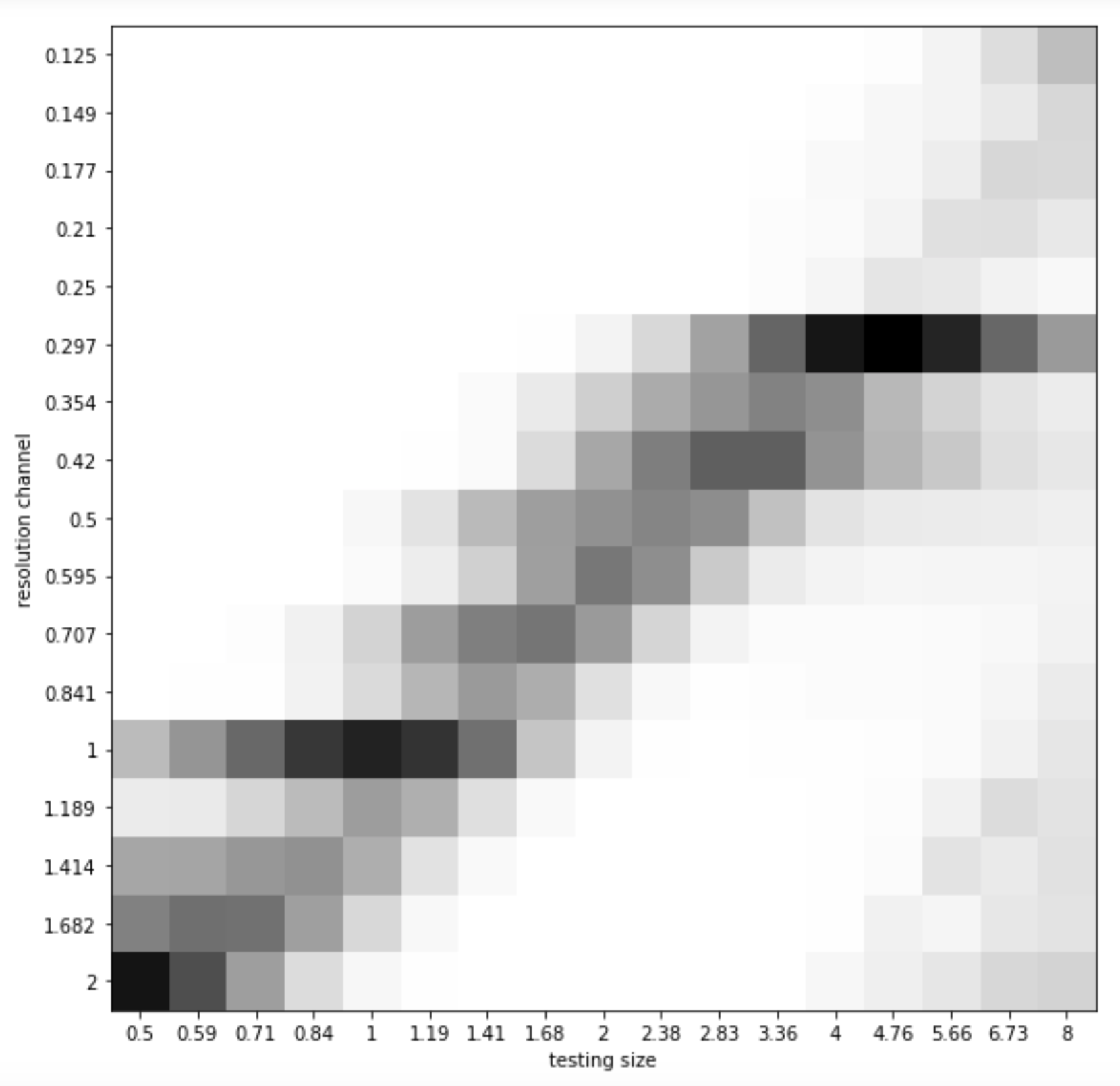}\\
      \\
     {\em Selected scale channels: FovConc network trained for size 4}
     & {\em Selected scale channels: SWMax network trained for size 4} \\
     \includegraphics[width=0.27\textheight]{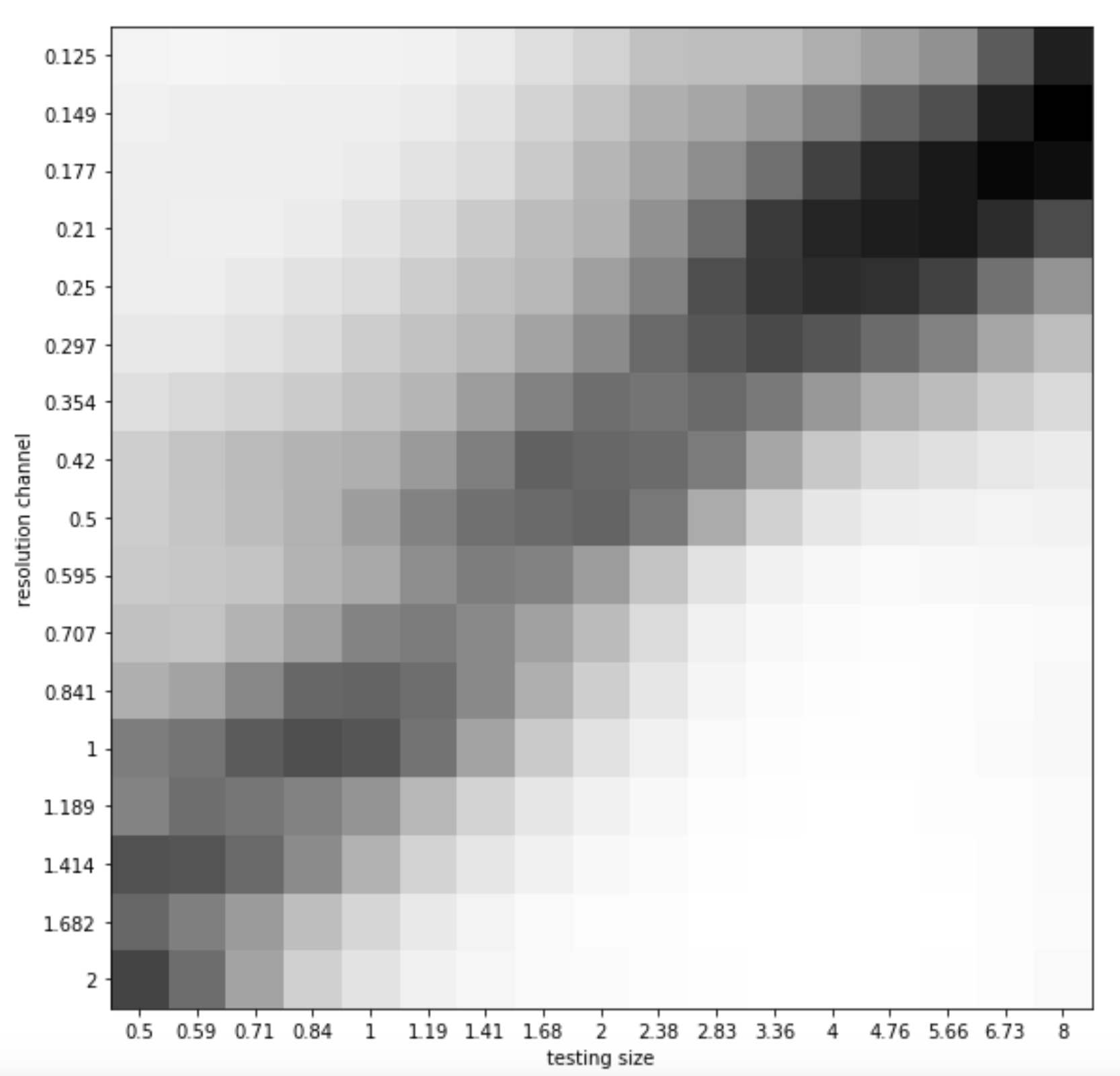}
     & \includegraphics[width=0.27\textheight]{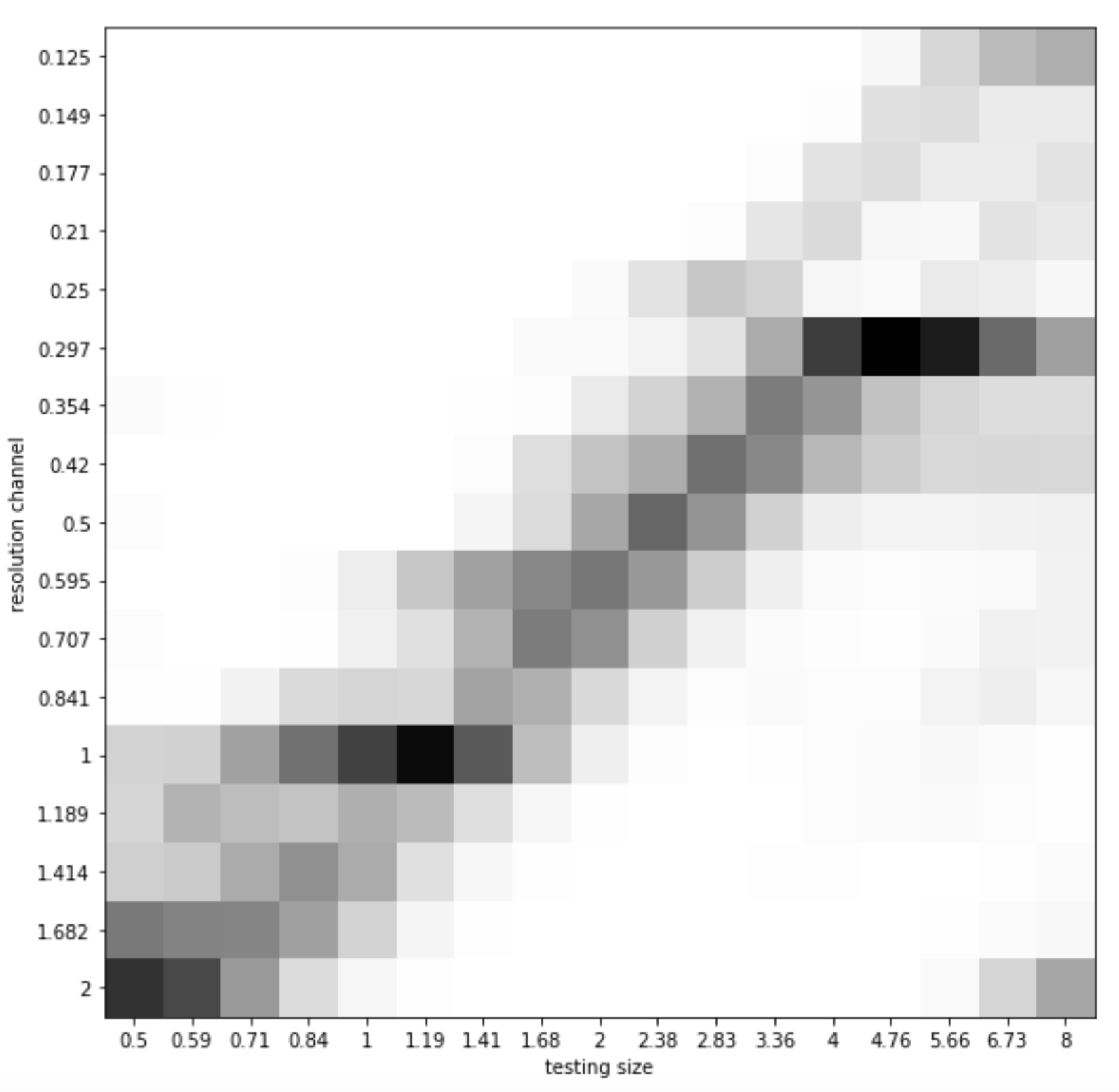}\\
    \end{tabular}
  \end{center}
  \caption{\emph{Visualisation of the scale selection properties of the not scale-invariant FovConc and SWMax
    networks, when training the network for each one
    of the sizes 1, 2 and 4}. For each testing size, shown on the
    horizontal axis with increasing testing sizes towards the right, the vertical axis displays a histogram of the relative contribution of the scale channels to the winning classification, with the
    lowest scale at the bottom and the highest scale at the top. As can be
    seen from the figures, the relative contributions from the
    different scale levels do not as well follow a linear dependency
    on the size of the input structures as for the scale-invariant FovAvg and FovMax
    networks. Instead, for the FocConc network, there is a bias
    towards the size of image structures used for training, whereas
    for the SWMax network some scale levels dominate for fine-scale or
    coarse-scale sizes in the testing data. (In these figures, the
resolution parameter on the vertical axis represents the inverse of
scale. Note that the grey-levels in the histograms are not directly
comparable, since the grey-levels for each histogram are normalised
with respect to the maximum and minimum values in that histogram.)}
   \label{fig-sc-sel-FovConc-SWMax}
\end{figure*}

\subsection{Scale selection properties}

One may ask, how do the scales ``selected" by the networks, {\em i.e.\/}, the scales that contribute the most to the feature response of the winning digit class, vary with the size of the object in the image? We, here, investigate the relative contributions from the different scale channels to the classification decision
and how they vary with the object size. For this purpose, we train the FovAvg,
FovMax, FovConc and SWMax networks on the MNIST Large Scale dataset for each one of the different training sizes 1, 2 and 4 and then accumulate histograms that quantify the contribution from the different scale channels over a range of image sizes in the testing data.

The histograms are constructed as follows:
\begin{itemize}
    \item \emph{FovMax}: We identify the scale channel that provides the maximum value for the winning digit class and increment the histogram bin corresponding to this scale channel with a unit increment.
    \item \emph{FovAvg}: The FovAvg network aggregates contributions from multiple scale channels for each classification decision. For the winning digit class, we, consider the \emph{relative contributions} from the different scale channels and increment each histogram bin with the corresponding fraction of unity of this contribution. The contribution is measured as the absolute value of the feature response before average pooling.
    \item \emph{FovConc}: We compute the relative contribution from each scale channel as the sum of the weights in the fully connected layer corresponding to the winning digit class \emph{and} the specific scale channel, multiplied by the feature values corresponding to the output from that scale channel. We increment each histogram bin with the fraction of unity corresponding to the absolute value of the relative contribution from each scale channel. 
    \item \emph{SWMax}: We identify the scale channel that provides the maximum value for the winning digit class and increment the histogram bin corresponding to this scale channel with a unit increment. 
\end{itemize}
The procedure is repeated for all the testing sizes in the MNIST Large Scale dataset, resulting in two-dimensional scale
selection histograms, which show what scale channels contribute to the classification output
as function of the size of the image structures in the testing data. The histograms are presented in
Figures~\ref{fig-sc-sel-FovAvg-FovMax}--\ref{fig-sc-sel-FovConc-SWMax}.
As can be seen in Figure~\ref{fig-sc-sel-FovAvg-FovMax}, for the FovAvg and FovMax networks, the selected scale levels do very well follow a linear trend  in the sense that the selected scale levels are proportional to the size of the image structures in the testing data.%
\footnote{A certain bias that can be observed for the FovMax and SWMax
networks, is that there is a stronger peak in the histogram scale channels for
scale channel 1 for small testing sizes, than for the neighbouring
scale channels. A possible explanation for this effect is that for
scale channel 1 there will not be any effective initial interpolation stage as
for the other scale channels, which implies that there is no
additional interpolation blur for this scale channel as for the other scale channels, in turn implying a stronger
response for this scale channel compared to the neighbouring scale
channels. A certain bias towards scale channel 1 can also be observed
for the FovConc network. For the FovAvg network, which is also the
network that performs clearly best out of these four networks, the
bias towards scale channel 1 is, however, very minor. In retrospect,
the bias towards scale channel 1 for the other networks could point to
replacing the initial bilinear interpolation stage by some other
interpolation method, and/or to add a small complementary smoothing stage after
the interpolation stage, to ensure that the sum of the effective interpolation blur
and the added complementary blur remains approximately the same for neighbouring
scale channels.}
The scale selection histograms are also largely similar, irrespective
of whether the training is performed for size 1, 2 or 4, illustrating
that the scale-invariant properties of the FovAvg and FovMax networks in the
continuous case transfer very well to the discrete implementation.

In this respect, the resulting scale-selection properties of the
FovAvg and FovMax networks share similarities to classical methods for
scale selection based on local extrema over scale or weighted
averaging over scale of scale-normalised derivative responses
\cite{Lin97-IJCV,Lin98-IJCV,Lin12-JMIV,Lin15-JMIV,Lin21-EncCompVis}. This makes sense in light of the result that the scaling properties of
the filters applied to the scale channels are similar to the scaling
properties of scale-normalised Gaussian derivatives (see  Section~\ref{sec:relation-to-scale-space2}). The
approach for the FovMax network is also closely related to the scale
selection approach in \cite{LooLiTax09-LNCS,LiTaxLoo12-IVC} based on choosing the scales at which a supervised
classifier delivers class labels with the highest posterior.

As can be seen in Figure~\ref{fig-sc-sel-FovConc-SWMax},
the behaviour is different for the not scale-invariant FovConc and SWMax networks.
For the FovConc network, there is a bias in that the selected scales are
more concentrated towards the size of the training data. The contributions from
the different scale channels are also much less concentrated around the
linear trend compared to the Fov\-Avg and FovMax networks. Without access to
multi-scale training, the FovConc network does not learn scale
invariance although this would in principle be possible, {\em e.g.\/}, by learning to use equal weights for all the scales, which would implement
average pooling over scales. 

For the SWMax network, although the resulting scale selection histogram
is largely centered around a linear trend, consistent with the relative robustness to scaling transformations that this network shows, the linear trend is not as
clean as for the FovAvg and FovMax networks.
For the coarsest scale testing structures, the SWMax network largely fails to activate
corresponding scale channels beyond a certain value.
This is consistent with the previously problems of not being able generalise to larger testing scales, and is likely related to the previously discussed problem of interference from zoomed-in previously unseen
partial views that might give stronger feature responses than the zoomed-out overall shape.
Furthermore, for finer or coarser scale testing structures, there are
some scale channels for the SWMax network that contribute more to the output than others,
and thus demonstrate a lack of true scale invariance.

In the quantitative scale generalisation experiments presented earlier, it was seen that the lack of scale invariance for the SWMax network leads to lower accuracy when generalising to unseen scales and, for the FovConc network, which here shows the worst scale selection properties, no marked improvement at all over a standard CNN.  For the truly scale-invariant FovAvg and FovMax networks, on the other hand, the ability of the networks to correctly identify the scale of the object in a scale-covariant way imply 
excellent scale generalisation properties.

\begin{figure*}[hbtp]
  \begin{center}
    		\includegraphics[width=1.0\textwidth]{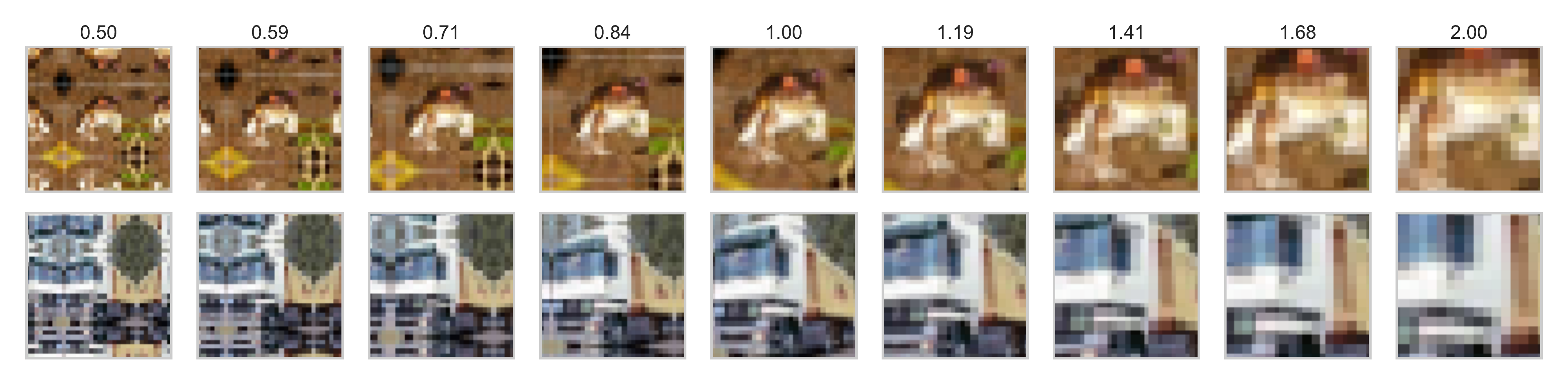} 
	\end{center}
	\caption{Sample images from the rescaled CIFAR-10 testing set (of size
          $32 \times 32$ pixels). The images in the original
          CIFAR-10 testing set are rescaled for scaling factors
          between $\frac{1}{2}$ and $2$, with mirror extension at the
          image boundaries for scaling factors $s < 1$. Top row: ``frog''. Bottom row: ``truck''.}
	\label{fig-cifar-rescaling}

        \bigskip
        
       \begin{tabular}{cc}
          \subfloat[Subfigure 1 list of figures text][Standard CNN]
          {\includegraphics[width=0.49\textwidth]{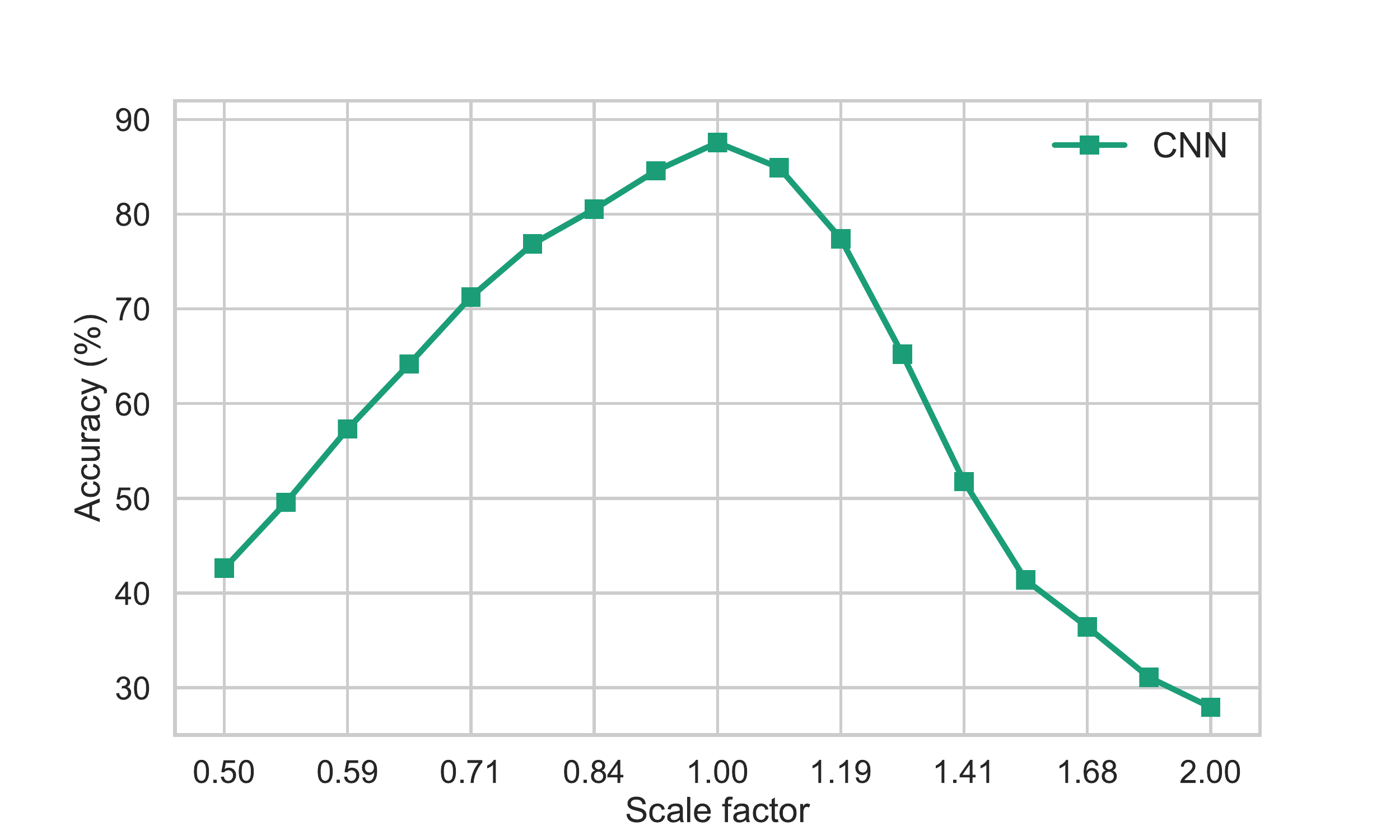}} &
	  \subfloat[Subfigure 2 list of figures text][The FovConc network]
          {\includegraphics[width=0.49\textwidth]{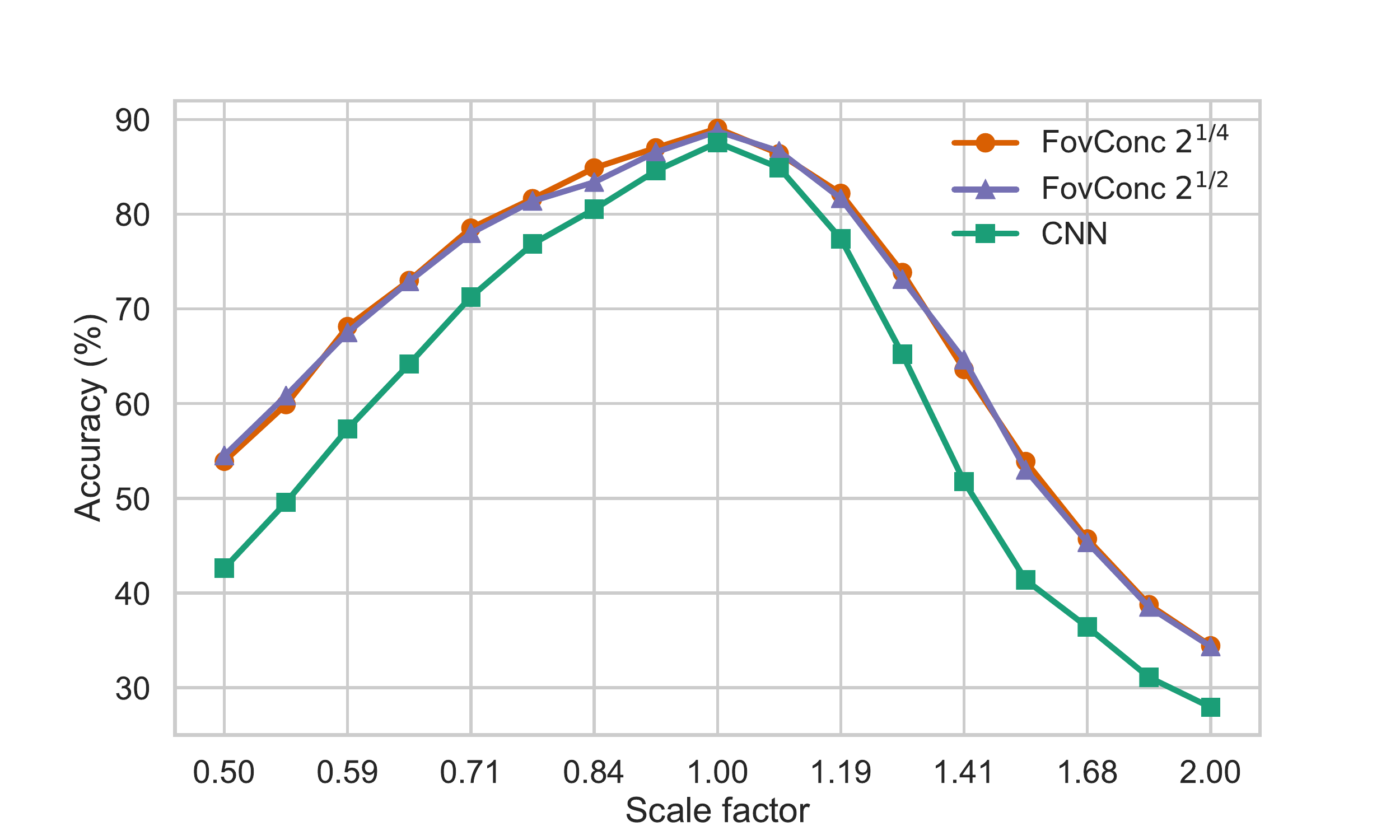}}  \\
	  \subfloat[Subfigure 3 list of figures text][The FovAvg network]
          {\includegraphics[width=0.49\textwidth]{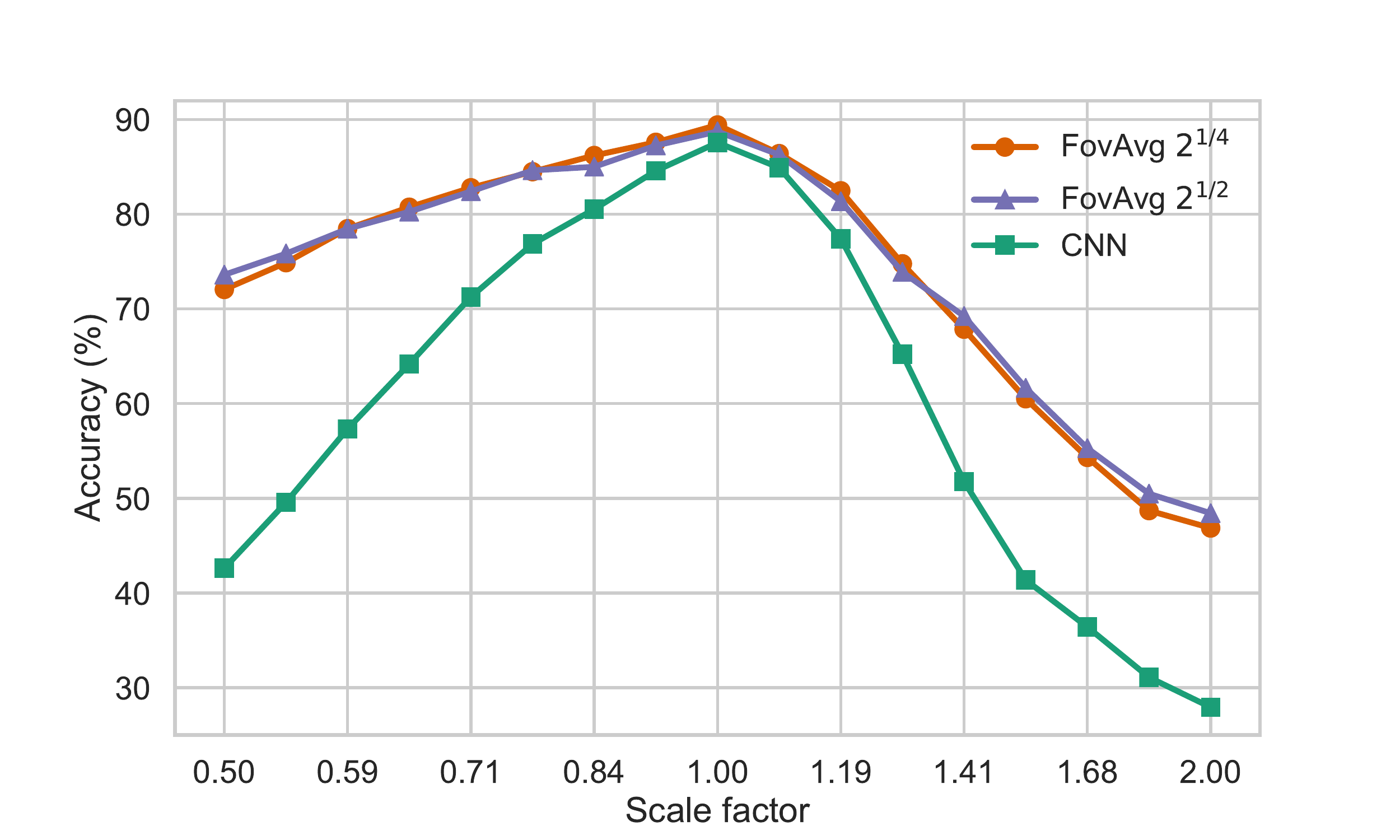}} &
	  \subfloat[Subfigure 4 list of figures text][The FovMax network]                                                                        
	  {\includegraphics[width=0.49\textwidth]{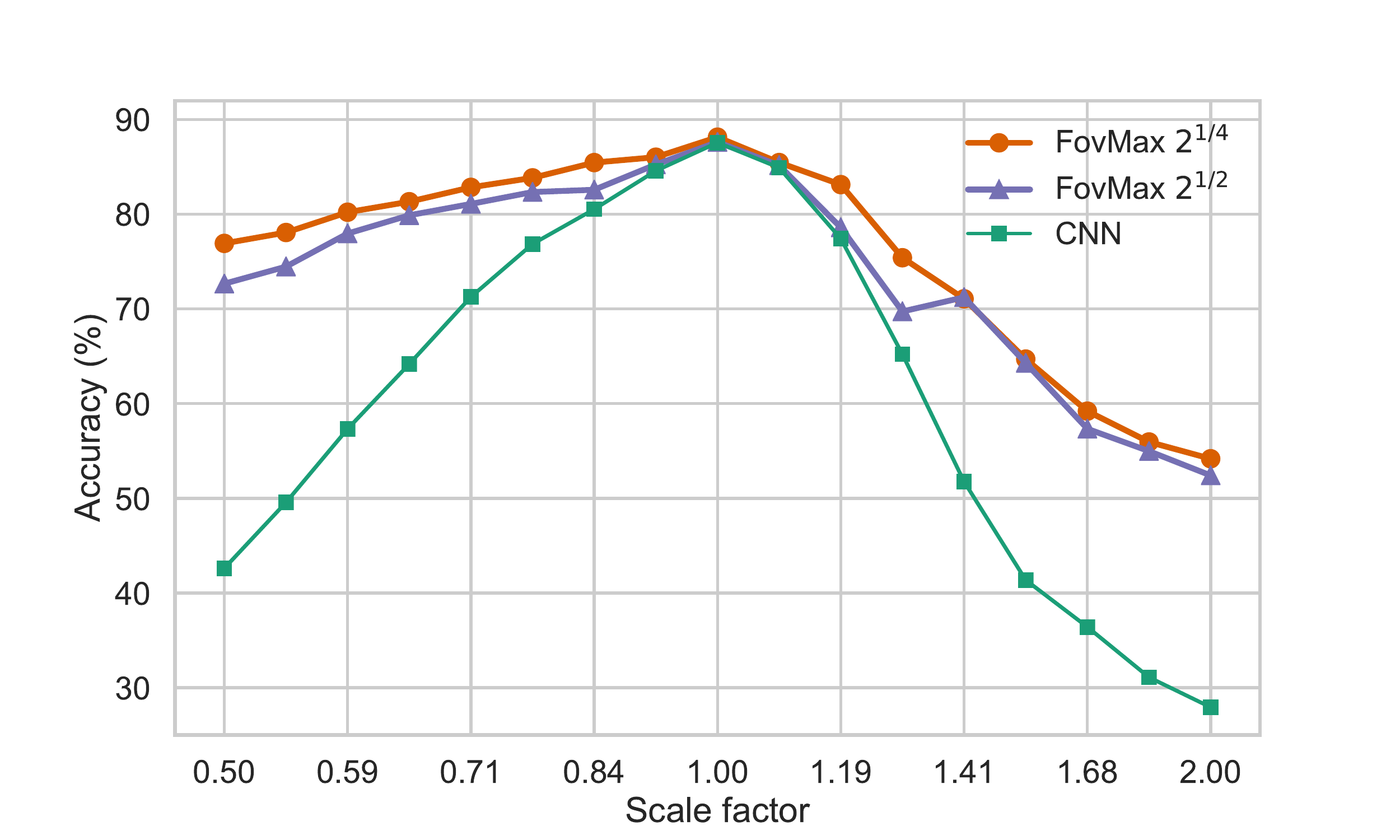}}  \\
      \end{tabular}
	\caption{{\em Generalisation ability to unseen scales for a
            standard CNN and different scale-channel network
            architectures for the rescaled CIFAR-10 dataset\/}.
          The network is trained on the
    CIFAR-10 training set (corresponding to scale factor 1.0) and
    tested on rescaled images from the testing set for relative scale factors
    between $\frac{1}{2}$ and $2$.
           The FovConc network has better scale generalisation
           compared to the standard CNN, but
           for larger deviations from the scale that the network is trained on, there is a
           clear advantage for the FovAvg and the FovMax networks.}
         \label{fig-cifar-10-scale-results}
\end{figure*}

\section{Experiments on rescalings of the CIFAR-10 dataset}
\label{sec-exp-rescaledCIFAR-10}

\subsection{Dataset}
\label{sec:cifar-scale}

To investigate if a scale-channel network can still provide a clear advantage over a standard CNN in a more challenging scenario, we use the CIFAR-10 dataset \cite{KriHin09-CIFAR}. We train on the original training set and test on synthetically rescaled copies of the test set with relative scale factors in the range $s \in [0.5, 2.0]$.
CIFAR-10 represents a dataset, where the
conditions for invariance using a scale-channel network are \emph{not
fulfilled}, in the sense that the transformations between different training and
testing sizes are not well modelled by continuous scaling
transformations, as underlie the presented theory for scale-invariant scale
channel networks, based on continuous models of both the image data
and the image filtering operations. 

Because already the original dataset is at the limit of being
undersampled, reducing the image size further for scale factors $s < 1$ 
results in additional loss of object details. 
The images are also tightly cropped, which implies that
increasing the image size for scale factors $s > 1$ implies a
loss of information towards the image boundaries, and that
sampling artefacts in the original image
data will be amplified.
Further, when reducing the image size, we extend the image by mirroring at the image boundaries, adding artefacts in the image structures, caused by the image padding operations. What we evaluate here is thus \emph{the limits} of the scale-channel networks, near or beyond the
limits of image resolution, to see if this approach can still provide a clear advantage over a standard CNN. 

Figure~\ref{fig-cifar-rescaling} shows a few images from the rescaled testing set,
with examples of two out of the 10 object classes in the dataset:
``airplanes'', ``cars'', ``birds'', ``cats'', ``deer'', ``dogs'',
``frogs'', ``horses'', ``ships'', and ``trucks''.

\subsection{Network and training details}

For the CIFAR-10 Scale dataset, we will compare the FovMax, FovAvg and
FovConc networks to a standard CNN.%
\footnote{We do not evaluate the SWMax
network on the CIFAR-10 Scale dataset, since it is not meaningful to
perform a spatial search for objects in this dataset.}
We use the same network for the CNN as for the individual scale channels, a 7-layer network with
conv+batchnorm+ReLU layers with $3 \times 3$  kernels and zero padding with width~1. We do not use any spatial max pooling, but use a stride of~2 for convolutional layers~3, 5 and~7.
After the final convolutional layer, spatial average pooling is performed over the full feature map down to $1 \times 1$ resolution, followed by a final fully connected softmax layer.
We do not use dropout, since it did not improve the results for this quite simple network with relatively few parameters. The number of feature channels is 32-32-32-64-64-128-128 for the 7 convolutional layers.

For the FovAvg and FovMax networks, max pooling and average
pooling, respectively, is performed across the logits outputs from the scale channels before the final softmax transformation and cross entropy loss.
For the FovConc network, there is a fully connected layer that combines the logits outputs from the multiple scale channels before applying a final softmax transformation and cross entropy loss.
We use bilinear interpolation and reflection padding at the image boundaries when computing the rescaled images used as input for the scale channels.

All the CIFAR-10 networks are trained for 20\,000 time steps using 50\,000 training samples from the CIFAR-10 training set over
103 epochs, using a batch size of~256 and the Adam optimiser
with default parameters in PyTorch: $\beta_1 = 0.9$ and $\beta_2 = 0.999$.
A cosine learning rate decay is used with
starting learning rate 0.001 and floor learning rate 0.00005,
where the learning rate decreases to the floor learning rate after 75~epochs.
The networks are then tested
on the 10\,000 images in the testing set, for relative scaling
factors in the interval $[\frac{1}{2}, 2]$.

We chose the learning rate and training schedule based on the CNN performance using the last
10\, 000 samples of the training set as a validation set.

\subsection{Experimental results}

The results for the {\em standard CNN\/} are shown in Figure~\ref{fig-cifar-10-scale-results}(a).
It can be seen that, already for scale factors slightly off from~1,
there is a noticeable drop in generalisation performance.

The results for the {\em FovConc network\/}, for different number of
scale channels, are presented in Figure~\ref{fig-cifar-10-scale-results}(b).
The generalisation ability to new scales is markedly better than for the
standard CNN, but the scale generalisation is not improved by
adding more scale channels. This can be compared with no improvement over a standard CNN when trained on single-scale MNIST data. We believe that the key difference is that for the CIFAR-10 dataset there are
indeed some scale variations present in the training set, and as
discussed earlier, it is possible for the FovConc network to learn to generalise by assigning appropriate weights to the layer that combines
information from the different scale channels. This illustrates that the method does have some structural advantage compared to a standard CNN,
but that multi-scale training data is required to realise this advantage. 

The results for the {\em FovMax and FovAvg networks\/}, for different numbers of
scale channels, are presented in Figure~\ref{fig-cifar-10-scale-results}(c--d), and
are significantly better than for the standard CNN and the FovConc
network. The accuracy for the smallest scale $1/2$ is improved from $\approx 40 \% $ 
for the CNN to above $70 \% $ for the FovAvg and FovMax networks, while the accuracy 
for the largest scale $2$ is improved from $\approx 30 \% $ for the CNN to $\approx 50 \% $ for the FovAvg and FovMax networks. 

For the FovMax network, there is a noticeable improvement by going to
a finer scale sampling ratio of $2^{1/4}$ compared to $2^{1/2}$.
Then, the generalisation ability for the FovMax network is also somewhat better than for the
FovAvg network. The FovAvg network does, however, have slightly better peak performance compared to the FovMax network. 

To summarise, the FovMax and FovAvg networks provide the best generalisation
ability to new scales, which is in line with theory.
This shows that, also for datasets where the conditions regarding image size and
resolution are not such that the scale-channel approach can provide
full invariance, our foveated scale-channel networks can nevertheless
provide benefits.


\section{Summary and discussion}

We have presented a methodology to handle scaling transformations in
deep networks by scale-channel networks.
Spec\-ifically, we have presented a theoretical formalism for modelling scale-channel networks
based on continuous models of the both the filters and the image data,
and shown that the continuous scale-channel networks are provably scale
covariant and translationally covariant. Combined with max pooling or
average pooling over the scale channels, our fove\-ated scale-channel networks
are additionally provably scale invariant.

Experimentally, we have demonstrated that discrete approximations
to the continuous foveated scale-channel networks FovMax and FovAvg are very robust to scaling
transformations, and allow for scale generalisation, with very good
performance for classifying image patterns at new scales not spanned by
the training data, because of the continuous invariance properties
that they approximate.
Experimentally, we have also demonstrated the very 
limited scale generalisation performance of vanilla CNNs and
scale concatenation networks 
when exposed to testing at scales not spanned by the training data,
although those approaches may work rather well when training on multi-scale training data. 
The reason why those approaches fail regarding scale generalisation,
when trained at a single scale or a over a narrow scale interval only,
is because of the lack of an explicit mechanism to enforce scale invariance.

We have further demonstrated that a foveated approach shows better generalisation performance compared to a sliding window approach, especially when moving from a smaller training scale to a large testing scale. Note that this should not be seen as an argument against any type of sliding window processing \emph{per se}. The foveated networks could, indeed, be applied in a sliding window manner to search for objects in a larger image. Instead, it illustrates that \emph{for any specific image point}, it is important to process a covariant set of image regions that correspond to different sizes in the input image. 

We have also demonstrated that our Fov\-Max and Fov\-Avg scale-channel
networks lead to improvements when training on data with significant
scale variations in the small sample regime.
We have further shown that the selected scale levels for these scale-invariant networks increase linearly with the size of the image
structures in the testing data, in a similar way as for classical
methods for scale selection. 

From the presented experimental results on the MNIST Large Scale dataset, 
it is clear that our Fov\-Max
and Fov\-Avg scale-channel networks do provide a
considerable improvement in scale generalisation ability compared to a
standard CNN as well as in relation to previous scale-channel
approaches.
Concerning the CIFAR-10 dataset, it should be noted that full invariance is not possible  because of the \emph{loss in image information} between the original and the rescaled images. 
Our experiments on this dataset show, nonetheless, that also in the presence of undersampling and serious boundary effects,
our Fov\-Max and Fov\-Avg scale-channel networks give considerably
improved generalisation ability compared to a standard CNN or alternative scale-channel networks. 

We believe that our proposed foveated scale-channel networks
could prove useful in situations where a simple approach that can
generalise to unseen scales or learn from small datasets with large
scale variations is needed.
Strong reasons for using such scale-invariant scale-channel networks
could either be because there is a limited amount of
multi-scale training data, where sharing statistical strength between
scales is valuable, or because only a single scale or a limited range of
scales is present in the training set, which implies that
generalisation outside the scales seen during training is crucial for
the performance.
Thus, we propose that this type of foveated scale-invariant
processing could be included as subparts in more complex
frameworks dealing with large scale variations.

Concerning applications towards object recognition, it should, however, be
emphasised that in this study, we have not specifically focused on
developing an integrated approach for detecting objects,
since the main focus has been to develop ways of handling the notion
of scale in a theoretically well-founded manner. Beyond the vanilla
sliding-window approach studied in this paper, which has such a built-in object detection
capability, also the foveated networks could be applied in a
sliding-window fashion, thus being able to also handle smaller objects
near the image boundaries, which is not possible if the central point
in the image is always used as the origin when resizing the image multiple
times to form the input for the different scale channels.

To avoid
explicit exhaustive search over multiple such origins for the foveated
representations, such an approach could further be naturally extended to a
two-stage approach, where
detection of points of interest is first performed using a
complementary module that detects points of interest (not
necessarily of the same kind as the current regular notion of interest points for
image-based matching and recognition),
followed by more detailed analysis of these points of interest with a
foveated representation. Such an approach would then bear
similarity to human vision, by foveating on interesting structures
to look at them in more detail. It would specifically also bear similarity to
two-stage approaches for object recognition, such as R-CNNs
\cite{GirDonDarMal14-CVPR,Gir15-ICCV,RenHeGirSun17-PAMI}, with the
difference that the initial detection step does not need to return
a full window of interest. Instead, only a single initial point is needed, where
the scale, corresponding to the size of the window, is then handled
by the built-in scale selection step in the foveated scale-channel
network. 

To conclude, the overarching aim of this study has instead been to test the limits of
CNNs to generalise to unseen scales over wide scale ranges. The key
take home message is a proof of concept that such scale generalisation is
possible, if including structural assumptions about scale in the
network design. 

\appendix
\normalsize

\section*{Appendix}

\section{The MNIST Large Scale dataset}
\label{app-mnist-large-scale}

We, here, give a more detailed description of the {\em MNIST Large Scale dataset\/}. 
The original MNIST dataset \cite{LecBotBenHaf98-ProcIEEE} contains images
of centered handwritten digits of size $28\times28$. 
The MNIST Large Scale dataset is derived from the MNIST dataset by
rescaling the original MNIST images. 
The resulting dataset contains images of size $112 \times 112$ with
scale variations of a factor of $16$. 
The scale factors $s$ relative the original MNIST images are $s \in
[\frac{1}{2}, 8]$. 
The dataset is illustrated in Figure~\ref{fig-dataset-mnist-scale}.
	
To create an image with a certain scale factor $s$, the original image
is first rescaled/resampled using bicubic interpolation. The image
range is then clipped to $[0, 256]$ to remove possible over/undershoot
resulting from the bicubic interpolation. The resulting image is
embedded into an $112 \times 112$ resolution image using zero padding
or cropping as needed. 
	
Large amounts of upsampling tend to result in discretisation
artefacts. To reduce the severity of such artefacts, the images are
post-processed with discrete Gaussian smoothing \cite{Lin90-PAMI}
followed by non-linear thresholding. The standard deviation of the
discrete Gaussian kernel varies with the scale factor as $\sigma(s) =
\frac{7}{8}s$. After smoothing, the image range is rescaled to the
range $[0, 255]$.
	
As a final step, an $\arctan$ non-linearity is applied to sharpen the
resulting image, where the final image intensity $I_{out}$ is computed
from the output of the smoothing step $I_{in}$ as:  
\begin{equation}
  I_{out} = \frac{2}{\pi}\arctan(a(I_{in} - b))
\end{equation}	
with $a=0.02$  and $b = 128$. 
Note that for scale factors $>4$, the full digit might not be visible
in the image. These scale factors are included to enable studying the
limits of generalisation when the entire object is no longer visible
(typically the digits are fully contained in the image for $s <
4\sqrt{2}$).
	
All training data sets are created from the first 50\,000 images in
the original MNIST training set, while the last 10\,000 images in the
original MNIST training set are used to create validation sets. The
testing data sets are created by rescaling the 10\,000 images in the
original MNIST testing set. For the multi-scale datasets, scale factors
for the individual images are sampled uniformly on a logarithmic scale
in the range $[s_{min}, s_{max}]$.
	
The specific MNIST Large Scale dataset used for the experiments in this paper is
available online \cite{JanLin20-MNISTLargeScale}.	
	
\bibliographystyle{splncs}
\bibliography{defs,tlmac}

\end{document}